\documentclass[10pt,twocolumn,letterpaper]{article}

\usepackage{iccv}             

\usepackage{times}
\usepackage{epsfig}
\usepackage{graphicx}
\usepackage{amsmath}
\usepackage{amssymb}
\usepackage{float}
\usepackage{mathtools}
\usepackage{booktabs}
\usepackage{subcaption}
\usepackage{comment}
\usepackage{color,colortbl}
\usepackage{multirow}
\usepackage{hhline}
\usepackage{import}
\usepackage{xargs}
\usepackage{bbm}
\usepackage{textcomp}
\usepackage[dvipsnames]{xcolor}
\usepackage[symbol]{footmisc}

\newcommand{\figref}[1]{Fig.~\ref{#1}}
\newcommand{\secref}[1]{Section~\ref{#1}}

\newcommand{\tabref}[1]{Table~\ref{#1}}

\makeatletter
\DeclareRobustCommand\onedot{\futurelet\@let@token\@onedot}
\def\@onedot{\ifx\@let@token.\else.\null\fi\xspace}
\def\eg{e.g\onedot} \def\Eg{E.g\onedot}

\makeatother

\definecolor{darkgreen}{rgb}{0,0.7,0}
\definecolor{Gray}{gray}{0.92}
\definecolor{iccvblue}{rgb}{0.21,0.49,0.74}
\usepackage[pagebackref,breaklinks,colorlinks,allcolors=iccvblue]{hyperref}

\def\paperID{14389}
\def\confName{ICCV}
\def\confYear{2025}

\title{SparseLaneSTP: Leveraging Spatio-Temporal Priors with Sparse Transformers for 3D Lane Detection}

\author{Maximilian Pittner\textsuperscript{1,2}, Joel Janai\textsuperscript{1}, Mario Faigle\footnotemark[1]\,\,\textsuperscript{1,3}, Alexandru Paul Condurache\textsuperscript{1,2} \\
\textsuperscript{1}Bosch Mobility Solutions, Robert Bosch GmbH\\
\textsuperscript{2}Institute of Neuro- and Bioinformatics, University of L\"ubeck\\
\textsuperscript{3}Institute for Signal Processing and System Theory, University of Stuttgart \\
{\tt\small \{Maximilian.Pittner, Joel.Janai, Mario.Faigle, AlexandruPaul.Condurache\}@de.bosch.com}
}

\begin{document}
\maketitle
\footnotetext[1]{Work done in context of master's thesis at Bosch.}
\begin{abstract}
3D lane detection has emerged as a critical challenge in autonomous driving, encompassing identification and localization of lane markings and the 3D road surface.
Conventional 3D methods detect lanes from dense birds-eye-viewed (BEV) features, though erroneous transformations often result in a poor feature representation misaligned with the true 3D road surface.
While recent sparse lane detectors have surpassed dense BEV approaches, they completely disregard valuable lane-specific priors. 
Furthermore, existing methods fail to utilize historic lane observations, which yield the potential to resolve ambiguities in situations of poor visibility. 
To address these challenges, we present SparseLaneSTP, a novel method that integrates both geometric properties of the lane structure and temporal information into a sparse lane transformer. 
It introduces a new lane-specific spatio-temporal attention mechanism, a continuous lane representation tailored for sparse architectures as well as temporal regularization. 

Identifying weaknesses of existing 3D lane datasets, we also introduce a precise and consistent 3D lane dataset using a simple yet effective auto-labeling strategy. 
Our experimental section proves the benefits of our contributions and demonstrates state-of-the-art performance across all detection and error metrics on existing 3D lane detection benchmarks as well as on our novel dataset.
\end{abstract}    
\section{Introduction}
\label{sec:introduction}
Accurate and robust 3D lane detection forms a pivotal task in autonomous driving addressing safe and reliable identification and localization of the road surface and lane markings. 
While significant progress has been made in 2D lane detection \cite{vpgnet,scnn,ghafoorian2018gan,lanenet,lightweightld,pizzati2019lane,zou2019robust,linecnn,laneatt,huval2015empirical,pinet,qu2021fololane,wang2022ganet}, 
the prediction output is only provided in image space lacking depth information, which is crucial for autonomous navigation in the 3D world.
In contrast, 3D lane detection jointly estimates lane markings and the road surface, directly producing 3D lanes in a vehicle-centered coordinate system. 
Conventional methods operate on dense feature maps that are transformed from the front-viewed (FV) to the birds-eye-viewed (BEV) perspective using inverse perspective mapping (IPM) \cite{3dlanenet,genlanenet,pittner20233d} or learned mappings \cite{chen2022persformer,wang2023bev,pittner2024lanecpp}. 
Although this step yields a suitable intermediate representation to capture the road, potential errors in the transformation cause misalignment between the resulting BEV and the true road surface that can hardly be compensated in the subsequent lane estimation step. 

As opposed to this two-stage approach, sparse detection methods have gained popularity in object detection \cite{misra2021detr3d} and have recently been adapted to 3D lane detection \cite{bai2023curveformer,luo2023latr}. 
These methods model 3D lane points as queries and associate them with uncorrupted FV image features - avoiding the necessity of error-prone BEV representations. 
Queries composed of learned context embedding vectors are integrated into transformer architectures \cite{carion2020detr,zhu2021defdetr} leveraging their ability to learn global context. 
However, current approaches completely ignore valuable well-known priors about lane and road geometry. 
Spatial priors have already been utilized in \cite{pittner2024lanecpp}. Based on lane continuity and other properties like line parallelism, a continuous lane representation was introduced together with an efficient regularization. Despite notable advances brought by this approach, the architecture is based on dense BEV representations and the proposed priors have not been successfully customized for sparse architectures. 
Another class of prior knowledge is grounded in the history of observations. 
Among various temporal fusion paradigms, object-centric query propagation \cite{wang2023exploring} has recently been proposed for 3D object detection in sparse architectures. 
In lane detection, temporal information has the potential to resolve ambiguities, particularly in scenarios with limited visibility or occlusions. Though, it has not been applied effectively so far.

We therefore present a new method SparseLaneSTP that combines spatial and temporal priors with a sparse 3D lane transformer architecture. 
Our model features a transformer-tailored continuous lane representation providing smooth curves directly. 
This, combined with robust regularization using additional spatial and temporal objectives, enables the model to learn regression and visibility estimation more accurately. 
Additionally, spatial and temporal knowledge is integrated into the attention mechanism in a novel way. 
Instead of standard global self-attention, we design the layers to focus on relevant relations based on lane-structure. 
Finally, lane queries are temporally propagated and incorporated into the attention to leverage valuable historic keys. 

On top of that, we provide a novel 3D lane dataset using a simple yet efficient auto-labeling pipeline.
Existing datasets often yield inaccurate labels containing noise and outliers, particularly in the far-range, since they rely on LiDAR to recover 3D information.
Our auto-labeling instead uses temporal aggregation along video-sequences resulting in consistent and accurate 3D lane labels up to 250 meters, making a valuable contribution to the research community.

\noindent Our contributions can be summarized as follows:
\begin{itemize}
	\item We present a novel 3D lane detector that integrates spatial and temporal knowledge into a sparse transformer.
	\item We propose a new attention mechanism that focuses on learning lane-structure and successfully leverages temporal keys from past lane observations.
	\item We formulate a continuous lane representation tailored for sparse architectures and a temporal regularization.	
	\item Our method achieves state-of-the-art performance on all 3D lane benchmarks.
	\item We provide a new dataset for 3D lane detection providing accurate and consistent long-range labels.
\end{itemize}
\section{Related work}
\label{sec:relatedwork}
\textbf{Dense BEV vs. sparse query-based detection}. 
A common strategy in 3D lane detection involves operating on an intermediate dense BEV representation generated from the FV. Early methods, such as 3D-LaneNet \cite{3dlanenet}, GenLaneNet \cite{genlanenet} and 3D-SpLineNet \cite{pittner20233d}, applied IPM to project FV features onto a flat ground plane. However, the assumption of a perfectly flat road is often violated, leading to degraded BEV features. 
Alternatively, methods from related fields like 3D object detection \cite{li2022bevformer,huang2021bevdet,huang2022bevdet4d} and BEV segmentation \cite{philion2020lift} proposed to learn the BEV representation. 
These concepts were applied to 3D lane detection with PersFormer \cite{chen2022persformer} leveraging deformable cross-attention (DCA) \cite{zhu2021defdetr}, whereas BEV-LaneDet \cite{wang2023bev} relies on simple multi-layer perceptrons (MLPs). 
LaneCPP \cite{pittner2024lanecpp} incorporates surface priors to learn 3D features using depth classification inspired by Lift-Splat-Shoot (LSS) \cite{philion2020lift}. Despite these advancements, all methods remain constrained by an intermediate error-prone BEV representation.

In contrast, sparse methods based on DETR \cite{carion2020detr,zhu2021defdetr} avoid BEV representation by modeling 3D objects as queries that exchange information via attention mechanisms. In 3D object detection, 3D position-aware object queries were introduced that are directly associated with the perspective view through DCA \cite{misra2021detr3d,lin2022sparse4d}. Similarly, in 3D lane detection, CurveFormer \cite{bai2023curveformer} adapts the sparse query design to lanes using polynomial-based curve queries, though these lack the flexibility to capture diverse lane structures in real-world scenarios. LATR \cite{luo2023latr} refines this approach by representing lanes as individual points but fails to incorporate crucial priors in both the attention mechanism and lane representation.

\noindent\textbf{Priors in lane detection}. 
An indisputable property about lanes lies in their smoothness and continuity. 
Several methods have proposed continuous lane representations in both 2D \cite{polylanenet,lstr,curvemodeling} and 3D \cite{liu2022learning,bai2023curveformer,pittner20233d,pittner2024lanecpp} lane detection.
In contrast to discrete approaches \cite{3dlanenet,genlanenet,chen2022persformer,huang2023anchor3dlane,wang2023bev}, continuous methods directly provide smooth curves, requiring almost no post-processing and exploit the entire available dense ground truth during training \cite{pittner2024lanecpp,pittner20233d}. 
While CLGO \cite{liu2022learning} and CurveFormer \cite{bai2023curveformer} use simple polynomials, \mbox{3D-SpLineNet} \cite{pittner20233d} and LaneCPP \cite{pittner2024lanecpp} use a B-Spline \cite{deboor197250} representation, which offers local control over curve segments enabling it to model complex shapes. 
Additionally, lane structures exhibit inherent properties, such as parallelism between lanes. 
\Eg SGNet \cite{sgld} introduces a penalty term to enforce a fixed lane width, but assumes a flat ground plane. 
GP \cite{li2022reconstruct} employs a parallelism loss to maintain a constant local distance between neighboring points, though its effectiveness is influenced by the number of anchor points used. 
In contrast, LaneCPP \cite{pittner2024lanecpp} demonstrates a more elegant way to encourage parallelism as well as other spatial properties like surface smoothness and curvature restriction using regularization losses. 
Motivated by LaneCPP, we propose a more suitable Catmull-Rom spline \cite{CRsplines} representation tailored for the transformer's attention with sparse queries.  We further present more sophisticated loss formulations and integrate prior knowledge into the transformer attention mechanism enhancing the focus on relevant relations. While a concurrent work also uses priors in attention \cite{chang2025rethinking}, our approach focuses on explicit modeling of structural lane relations and even effectively incorporates temporal knowledge in lane detection for the first time.

\begin{figure*}[tb]
	\centering
	\includegraphics[width=1.\linewidth]{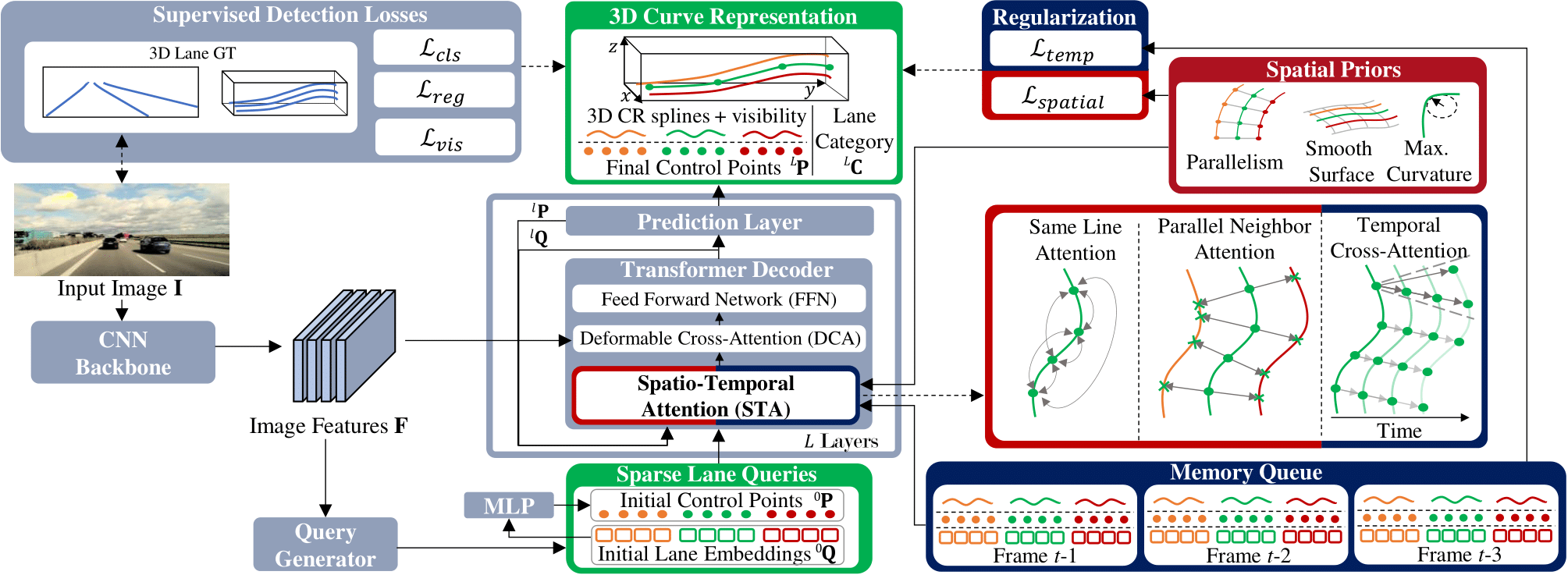}
	\caption{Overview of our method SparseLaneSTP. \textcolor{darkgreen}{Sparse lane queries} are processed by a transformer integrating \textcolor{Blue}{temporal knowledge} and \textcolor{BrickRed}{spatial lane structure priors} in a novel \textcolor{BrickRed}{spatio}-\textcolor{Blue}{temporal} attention. Based on these priors, we formulate \textcolor{BrickRed}{spatial} and \textcolor{Blue}{temporally consistent} regularization. Finally, our network predicts \textcolor{darkgreen}{control points defining our new continuous 3D lane representation}.
}
	\label{fig:overview}
\end{figure*}

\noindent\textbf{Temporal modeling}. 
Another source of prior knowledge is grounded in the history of observations. For video-based 3D object detection various approaches exist to model temporal interactions. 
BEV-based methods \cite{li2022bevformer} transform past BEV features according to the ego-motion and fuse them with the current representation. 
More advanced perspective view (PV) methods \cite{luo2022detr4d,lin2022sparse4d,liu2023petrv2} instead follow the sparse query design but require storing dense historic PV feature maps to interact with object queries. Recently, object-centric temporal modeling \cite{wang2023exploring,lin2023sparse4dv2} has been proposed, which only stores sparse historic object queries and propagates them according to ego- and object-motion in 3D space. 
In particular, Sparse4Dv2 \cite{lin2023sparse4dv2} and StreamPETR \cite{wang2023exploring} leverage the sparse query design and model object queries as hidden states of a transformer, where the relation between current and historic queries is learned via attention. 

In 3D lane detection, only few approaches exist that use temporal modeling. 
STLane3D \cite{wang2022spatio} follows a BEV-based method, PETRv2 \cite{liu2023petrv2} fuses historic PV features together with propagated 3D positional embeddings. 
On the other hand, Anchor3DLane-T \cite{huang2023anchor3dlane} leverages temporal information by associating 3D anchors from the current frame with features from previous frames. 
In contrast, our sparse query-based architecture is more suitable for the effective memory-efficient object-centric paradigm. 
Consequently, we propose a temporal attention mechanism inspired by \cite{wang2023exploring,lin2023sparse4dv2}, which exploits lane-specific properties by reinforcing most relevant query interactions. 
The static nature of lane markings also allows us to formulate regularization to encourage temporally consistent perception behavior.
\section{Methodology}
\label{sec:method}
In the following, we describe our 3D lane detection method illustrated in \figref{fig:overview}. Given an image $\mathbf{I}$, the model predicts a set of $N$ lanes as continuous curves in 3D space, parameterized by our novel tailored spline representation with $M$ control points $\mathbf{P}$ (see \secref{subsec:lanerep}) and lane categories $\mathbf{C}$. 
The core component of our architecture is the transformer decoder, where each control point, inspired by sparse query design, is paired with a learned context embedding and is iteratively processed through $L$ decoder layers.
In contrast to previous work \cite{bai2023curveformer,luo2023latr,liao2023maptr} that simply adapt the DETR \cite{carion2020detr,zhu2021defdetr} framework, our transformer architecture introduces a novel spatio-temporal attention layer to capture spatial intra- and inter-lane relations, while a temporal cross-attention mechanism integrates past observations via propagated lane queries from a memory queue. 
Finally, besides standard detection losses, our model is cautiously regularized by spatial and temporal objectives to enhance robust and consistent detection behavior.

\begin{figure}[tb]
	\begin{subfigure}[b]{0.6\linewidth}
		\includegraphics[width=0.85\linewidth]{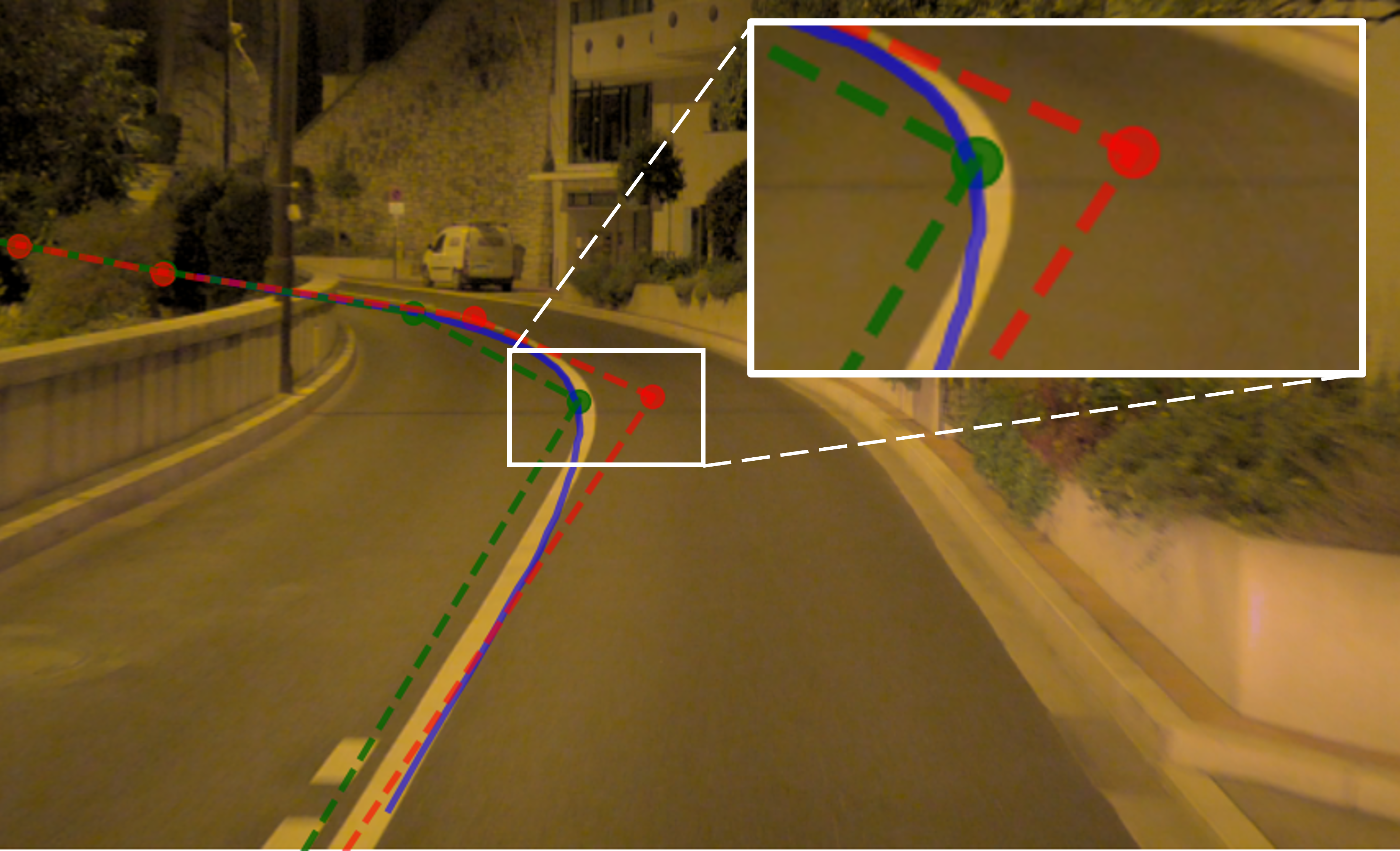}
		\caption{Front-view\label{fig:crvsbsplines1}}
	\end{subfigure}
	\begin{subfigure}[b]{0.39\linewidth}
		\includegraphics[width=0.85\linewidth]{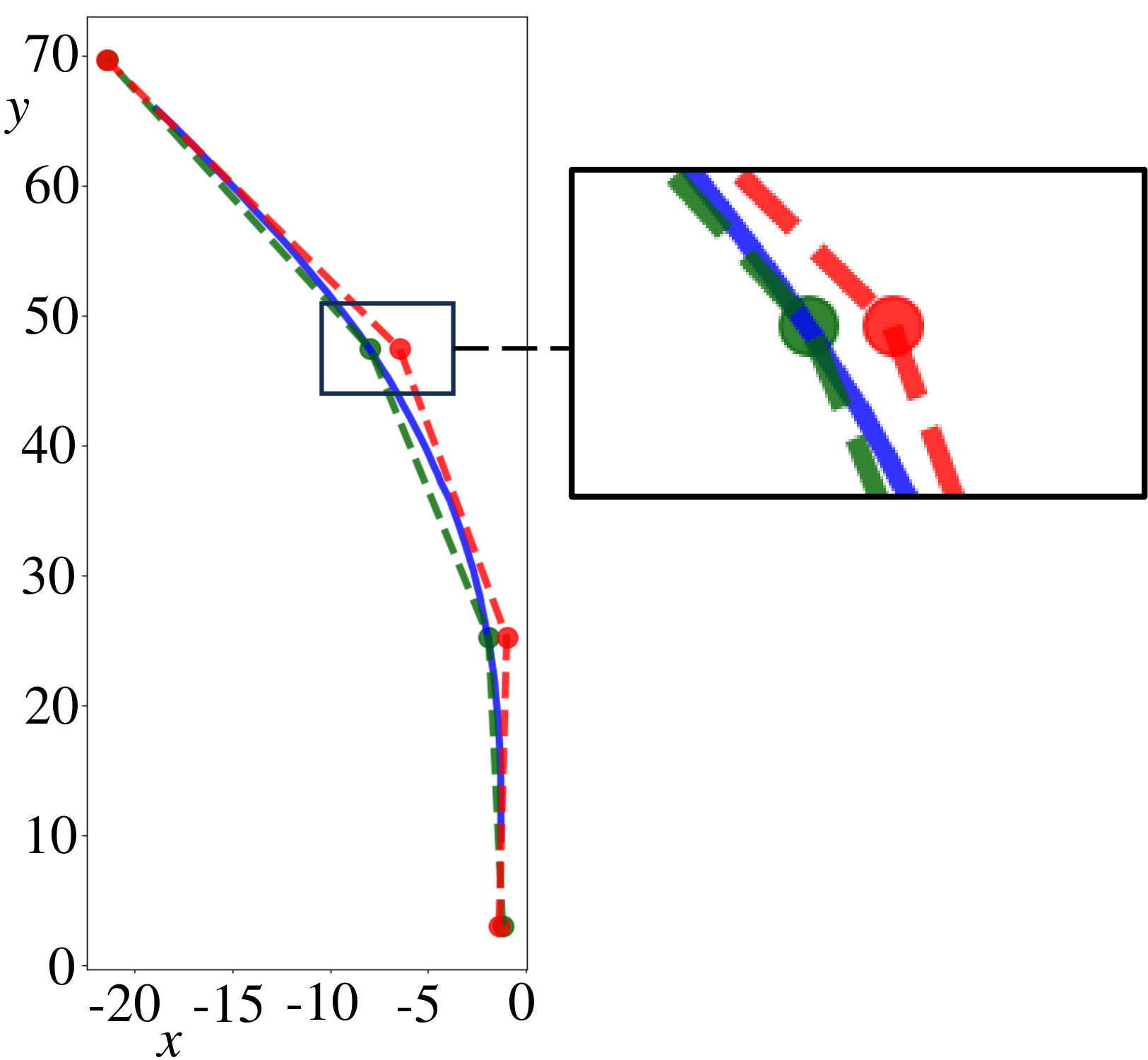}
		\caption{BEV 
		\label{fig:crvsbsplines2}}
	\end{subfigure}
	\caption{\textcolor{darkgreen}{CR spline} and \textcolor{red}{B-Spline} control points in comparison, both fit to the \textcolor{blue}{ground truth}. B-Spline control points do not align with the curve and are therefore not suiting the sparse query design, whereas CR control points exactly match curve geometry.}\label{fig:crvsbsplines}
\end{figure}

\subsection{3D lane representation}\label{subsec:lanerep}
For the 3D lane representation, we take inspiration from prior work \cite{pittner2024lanecpp,pittner20233d} that demonstrated significant benefits of continuous curves. 
In these frameworks, the model output serves as control points of B-Spline curves that do not necessarily align with the curve geometry. 
However, using the sparse query design, the 3D position of control points serve as the model's internal state and should therefore correspond to the exact 3D position of lane points. 
Thus, spline control points should directly lie on the curve, which is not the case for B-Splines (see \figref{fig:crvsbsplines}). 
Therefore, we choose Catmull-Rom (CR) splines \cite{CRsplines}, a class of piece-wise defined smooth third-order polynomial splines, for which the curve inherently passes through its control points. 

We adapt this spline-model to define a curve representation $\mathbf{f}_i(s) \in \mathbb{R}^{4} $ representing the entire $i^{\textrm{th}}$ line proposal parameterized by $M$ control points with 3 spatial ($x_{ij}$, $y_{ij}$, $z_{ij}$) and 1 visibility ($v_{ij}$) dimension as
\begin{align}
	\mathbf{f}_i(s) &= \begin{bmatrix} s^3 & s^2 & s & 1 \end{bmatrix} \cdot \mathbf{M}_{\textrm{CR}} \cdot \mathbf{P}_i \quad \textrm{with} \\
 \mathbf{P}_i &= [\mathbf{P}_{\textrm{3D},i} | \mathbf{P}_{v,i}] =
\begin{bmatrix} 
x_{i1} & y_{i1} & z_{i1} & v_{i1} \\
 \multicolumn{4}{c}{\dots}  \\
x_{ij} & y_{ij} & z_{ij} & v_{ij} \\
 \multicolumn{4}{c}{\dots}  \\
x_{iM} & y_{iM} & z_{iM} &  v_{iM} \\
\end{bmatrix} \, , 
\end{align}
with curve argument $s \in [0, \, 1]$. 
$\mathbf{P}_i \in \mathbb{R}^{M \times 4}$ denotes the control point matrix composed of $\mathbf{P}_{\textrm{3D},i} \in \mathbb{R}^{M \times 3}$ defining the curve 3D shape and visibility $\mathbf{P}_{v,i} \in \mathbb{R}^{M \times 1}$. $[\cdot|\cdot]$ is the concatenation operator and $\mathbf{M}_{\textrm{CR}}$ the CR coefficient matrix. Note that we can eventually express the curve computation as a simple matrix multiplication between the pre-computed spline arguments and the entire predicted control points $\mathbf{P}$. 

\subsection{Overall architecture}\label{subsec:architecture}
\noindent\textbf{Backbone and query initialization.} 
The RGB input image $\mathbf{I} \in \mathbb{R}^{H \times W \times 3}$ is first processed by a CNN backbone (\eg ResNet \cite{he2016deep}) to extract image features of $\mathbf{F} \in \mathbb{R}^{H_\mathbf{F} \times W_\mathbf{F} \times C}$, where $C$ is the channel size. Similar to \cite{luo2023latr}, we then feed $\mathbf{F}$ through a lane instance segmentation branch to obtain initial query embedding vectors $\prescript{0}{}{\mathbf{Q}} \in  \mathbb{R}^{N \times M \times C}$, where $N$ denotes the number of line proposals and $M$ the number of spline control points. An MLP then predicts initial control points $\prescript{0}{}{\mathbf{P}} \in  \mathbb{R}^{N \times M \times 4}$ from  $\prescript{0}{}{\mathbf{Q}}$.

\noindent\textbf{Transformer decoder.} In each transformer decoder layer $l$, queries are processed by two attention layers
\begin{align}
	\prescript{l}{}{\mathbf{Q}}_\mathrm{STA} &= \mathrm{STA}\big( \prescript{l-1}{}{\mathbf{Q}}, \, \prescript{l-1}{}{\mathbf{P}}, \, \mathbf{Q}_\mathrm{Mem}, \, \mathbf{P}_\mathrm{Mem}\big) \\
	\prescript{l}{}{\mathbf{Q}}_\mathrm{DCA} &= \mathrm{DCA}\big(\prescript{l}{}{\mathbf{Q}}_{\mathrm{STA}}, \, \mathbf{F}, \, \prescript{l-1}{}{\mathbf{P}}_\mathrm{2D}\big),
\end{align}
$\mathrm{STA}(\cdot)$ denotes our proposed spatio-temporal attention, which is explained in \secref{subsec:spatialtemporalpriors}, and $\mathrm{DCA}(\cdot)$ the standard deformable cross-attention presented in \cite{zhu2021defdetr}. 
$\mathbf{P}_\mathrm{2D}$ are the 2D spline control points in image coordinates resulting from re-projecting $\mathbf{P}_{\textrm{3D}}$. 
Finally, the query embeddings are fed through a feed-forward network $\prescript{l}{}{\mathbf{Q}} = \mathrm{FFN}(\prescript{l}{}{\mathbf{Q}}_\mathrm{DCA})$ composed of an MLP and layer normalization.

\noindent\textbf{Prediction layer.} 
After each transformer attention layer, the emerging query embeddings $\prescript{l}{}{\mathbf{Q}}$ are fed through a prediction layer that outputs the spline control points  $\prescript{l}{}{\mathbf{P}}$ as well as classification probabilities $\mathbf{C} \in \mathbb{R}^{N \times (K+1)}$ for $K$ lane categories and one background class per line. Particularly, for each layer $l$ and query $\prescript{l}{}{\mathbf{Q}}_{ij}$ three MLPs with shared weights among all layers are used to predict the $x$-, $z$- and $v$-component of the control point matrix
\begin{align}
	\prescript{l}{}{\mathbf{P}}_{x,ij} &= \sigma(\mathrm{MLP}_x(\prescript{l}{}{\mathbf{Q}}_{ij})) \cdot (x_{e}-x_{s}) + x_{s} \\
 	\prescript{l}{}{\mathbf{P}}_{z,ij} &= \sigma(\mathrm{MLP}_z(\prescript{l}{}{\mathbf{Q}}_{ij})) \cdot (z_{e}-z_{s}) + z_{s} \\ 
	\prescript{l}{}{\mathbf{P}}_{v,ij} &= \sigma(\mathrm{MLP}_v(\prescript{l}{}{\mathbf{Q}}_{ij})) \, , 
\end{align}
where the sigmoid $\sigma(\cdot)$ normalizes the MLP output to $[0, \, 1]$. The start-/end- of $x$- and $z$-range $x_{s/e}$, $z_{s/e}$ scales up the normalized output to the desired range. The longitudinal $y$-component $\prescript{l}{}{\mathbf{P}}_{y,ij}$ is pre-defined and uniformly distributed between $y_{s}$ and $y_{e}$ to avoid over-parameterization and potential overfitting as argued in \cite{pittner2024lanecpp}. 
For classification, all queries belonging to the same line proposal are averaged and fed into an MLP followed by a softmax layer. 

\noindent Finally, the formulation of our supervised losses is inspired by \cite{pittner2024lanecpp}, with L1 loss for regression of $x$ and $z$, binary cross-entropy for visibility $v$ and focal loss \cite{lin2017focal} for classification.

\begin{figure}[tb]
       \centering
       \includegraphics[width=1.\linewidth]{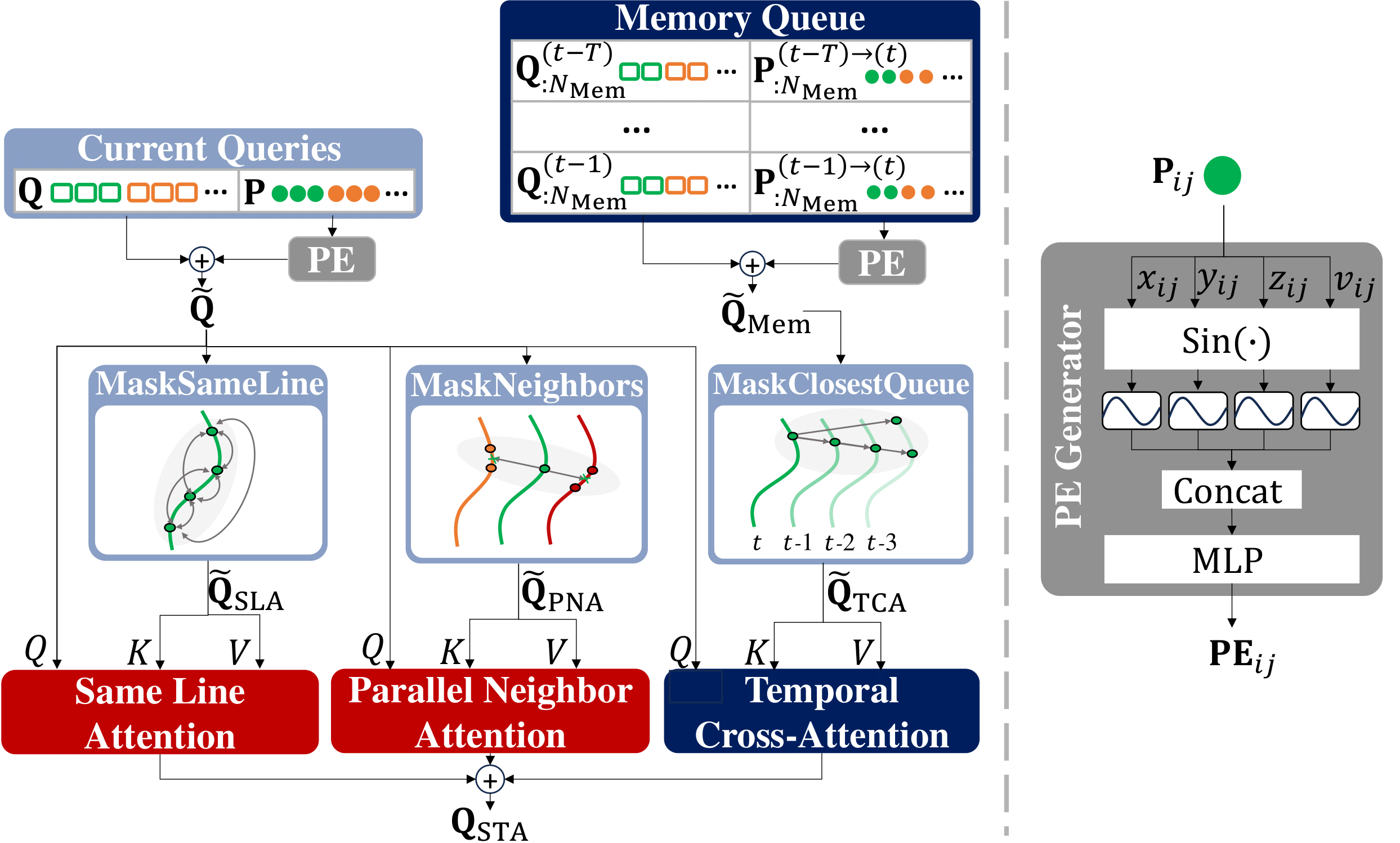}
       \caption{Overview of our spatio-temporal attention module. 
}
       \label{fig:sta}
\end{figure}

\subsection{Integrating spatial and temporal knowledge}\label{subsec:spatialtemporalpriors} 
\noindent \textbf{Memory queue.} 
To facilitate attention to past observations, queries are stored within a memory queue. Following a similar approach to \cite{wang2023exploring}, this queue is recursively updated after each prediction step and retains queries from the past $T$ frames. Therefore, embedding vectors and their associated control points are recursively pushed into the queue using a first-in, first-out (FIFO) strategy, which yields
\begin{align}
	\mathbf{{Q}}_{\textrm{Mem}} &= \big[ \mathbf{Q}_{:N_\textrm{Mem}}^{(t-1)} \, \big| \, \mathbf{Q}_{:N_\textrm{Mem}}^{(t-2)} \, \big| \, \dots \, \big| \, \mathbf{Q}_{:N_\textrm{Mem}}^{(t-T)} \big]  \quad \textrm{and} \\
	\mathbf{{P}}_{\textrm{Mem}} &= \big[ \mathbf{P}_{:N_\textrm{Mem}}^{(t-1)} \, \big| \, \mathbf{P}_{:N_\textrm{Mem}}^{(t-2)} \, \big| \, \dots \, \big| \, \mathbf{P}_{:N_\textrm{Mem}}^{(t-T)} \big] \, ,
\end{align}
where $\mathbf{Q}_{:N_\textrm{Mem}}^{(t)} \in \mathbb{R}^{(N_\textrm{Mem} \times M \times C)}$ denotes the embedding vectors from the $N_{\mathrm{Mem}}$ most confident predictions of frame $t$ and $\mathbf{Q}_{\textrm{Mem}} \in \mathbb{R}^{((T \cdot N_\textrm{Mem}) \times M \times C)}$ the full memory queue. 

\noindent \textbf{Temporal propagation.} 
Since the model is supposed to learn spatio-temporal context, it must incorporate the positional information of 3D control points from past frames. However, due to the vehicle's ego-motion, lane positions shift relative to the vehicle. Therefore, the 3D coordinates of past lane control points stored in the memory queue must be transformed according to the ego-pose $\mathbf{E}^{(t)}$ of frame $t$, which is represented as a homogeneous transformation matrix. 
Consequently, a control point $\mathbf{{P}}_{ij}^{(t-k)\rightarrow(t)}$ propagated from frame $t-k$ to present frame $t$ is given as
\begin{align}
	\mathbf{{P}}_{ij}^{(t-k)\rightarrow(t)} = \big[ \, \mathbf{E}_{inv}^{(t)} \cdot \mathbf{E}^{(t-k)} \cdot \mathbf{{P}}_{\textrm{3D},ij}^{(t-k)} \, \big| \, \mathbf{{P}}_{v,ij}^{(t-k)} \,\big] \, , 
\end{align}
where $\mathbf{E}_{inv}$ denotes the inverted matrix $\mathbf{E}$. Obviously, only the geometry component is affected by the propagation, whereas the visibility component remains unchanged. 
To incorporate spatio-temporal information from propagated control points into queries, we apply a positional encoding ($\mathrm{PE}$) based on 3D geometry and visibility (see \figref{fig:sta}). This process yields position-informed embedding vectors 
\begin{align}
	\mathbf{\tilde{Q}}_{ij}^{(t-k)\rightarrow(t)} = \mathbf{{Q}}_{ij}^{(t-k)\rightarrow(t)} + \mathrm{PE}\big( \mathbf{{P}}_{ij}^{(t-k)\rightarrow(t)} \big) \, , 
\end{align}
which are processed by the spatio-temporal attention layer.

\noindent \textbf{Spatio-temporal attention.} 
Most methods \cite{luo2023latr,bai2023curveformer} rely solely on global self-attention, which contains a majority of redundant lane query interactions that can distract the model from learning relevant context. To address this, we leverage spatial and temporal priors, distinguishing three key relation types: intra-lane (between adjacent points on the same lane), inter-lane (between parallel lanes), and historic (between current and past queries in the memory queue). 

We therefore introduce same line attention ($\mathrm{SLA}$), parallel neighbor attention ($\mathrm{PNA}$) and temporal cross-attention ($\mathrm{TCA}$). 
Applying specific masking, $\mathrm{SLA}$ restricts interactions to queries $\mathbf{\tilde{Q}}_\mathrm{SLA}$ within the same line, $\mathrm{PNA}$ facilitates interactions with queries $\mathbf{\tilde{Q}}_\mathrm{PNA}$ from neighboring lines, and $\mathrm{TCA}$ enables interactions with closest past queries $\mathbf{\tilde{Q}}_\mathrm{TCA}$ in the memory queue (see \figref{fig:sta}). 
Such restriction of query interactions reduces redundancy, encourages the model to focus on critical lane-specific spatial context and integrates valuable temporal information from past observations. 

\noindent \textbf{Regularization.} 
Based on spatial and temporal priors we can formulate regularization techniques that promote robust and consistent detection behavior. 
For the spatial regularization $\mathcal{L}_{spatial}$ we adopt the scheme proposed in \cite{pittner2024lanecpp} encouraging lane parallelism, surface smoothness and suppressing excessive curvature. 
Besides, temporal consistency can be achieved by optimizing a smoothness loss 
\begin{align}
	\mathcal{L}_{temp} = \frac{1}{N} \sum_i^N \int_s \mathbf{\bar{f}}_{v,i}^{(t)}(s) \cdot \big|\big| \mathbf{f}_{\textrm{3D},i}(s) - \mathbf{\bar{f}}_{\textrm{3D},i}^{(t)}(s) \big|\big|_1 \, ,
\end{align}
where $\mathbf{\bar{f}}_{\textrm{3D},i}^{(t)}(s)$ or $\mathbf{\bar{f}}_{v,i}^{(t)}(s)$ is an exponentially moving average of the past predictions 
\begin{align}
	\mathbf{\bar{f}}_{i}^{(t)} = \alpha \cdot \mathbf{\bar{f}}_{i}^{(t)} + (1-\alpha) \cdot \mathbf{\bar{f}}_{i}^{(t-1)}
\end{align}
with smoothing factor $\alpha \in [0, 1]$.
\begin{figure}[tb]
	\centering
	\includegraphics[width=0.99\linewidth]{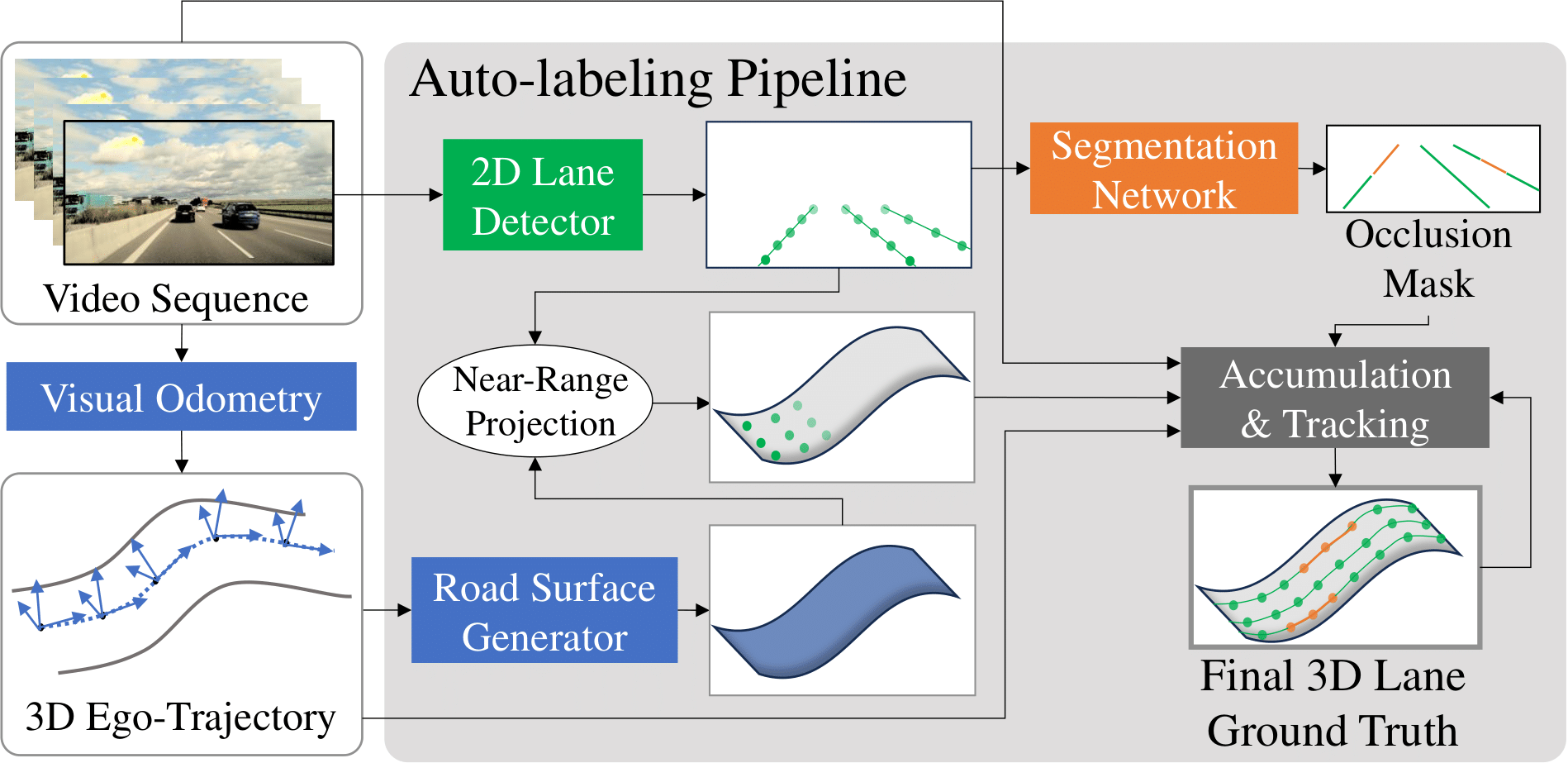}
	\caption{Our auto-labeling strategy. }	
	\label{fig:autolabeling}
\end{figure}
\begin{figure}
	\centering
	\begin{subfigure}[b]{.49\linewidth}
		\centering
		\includegraphics[width=1.\linewidth]{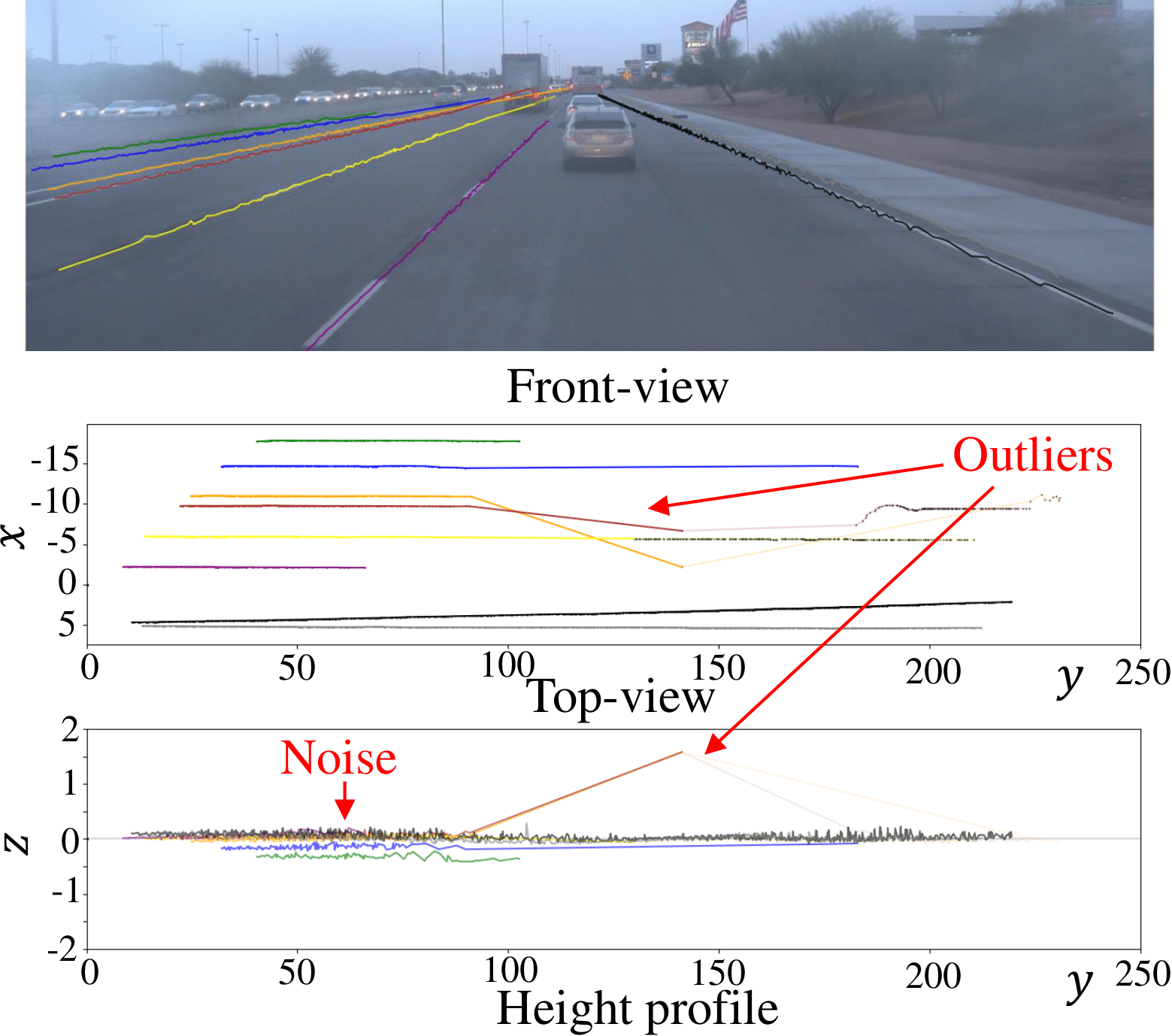}
		\caption{OpenLane\label{fig:dataexamples1}}
	\end{subfigure}
	\begin{subfigure}[b]{.49\linewidth}
		\centering
		\includegraphics[width=1.\linewidth]{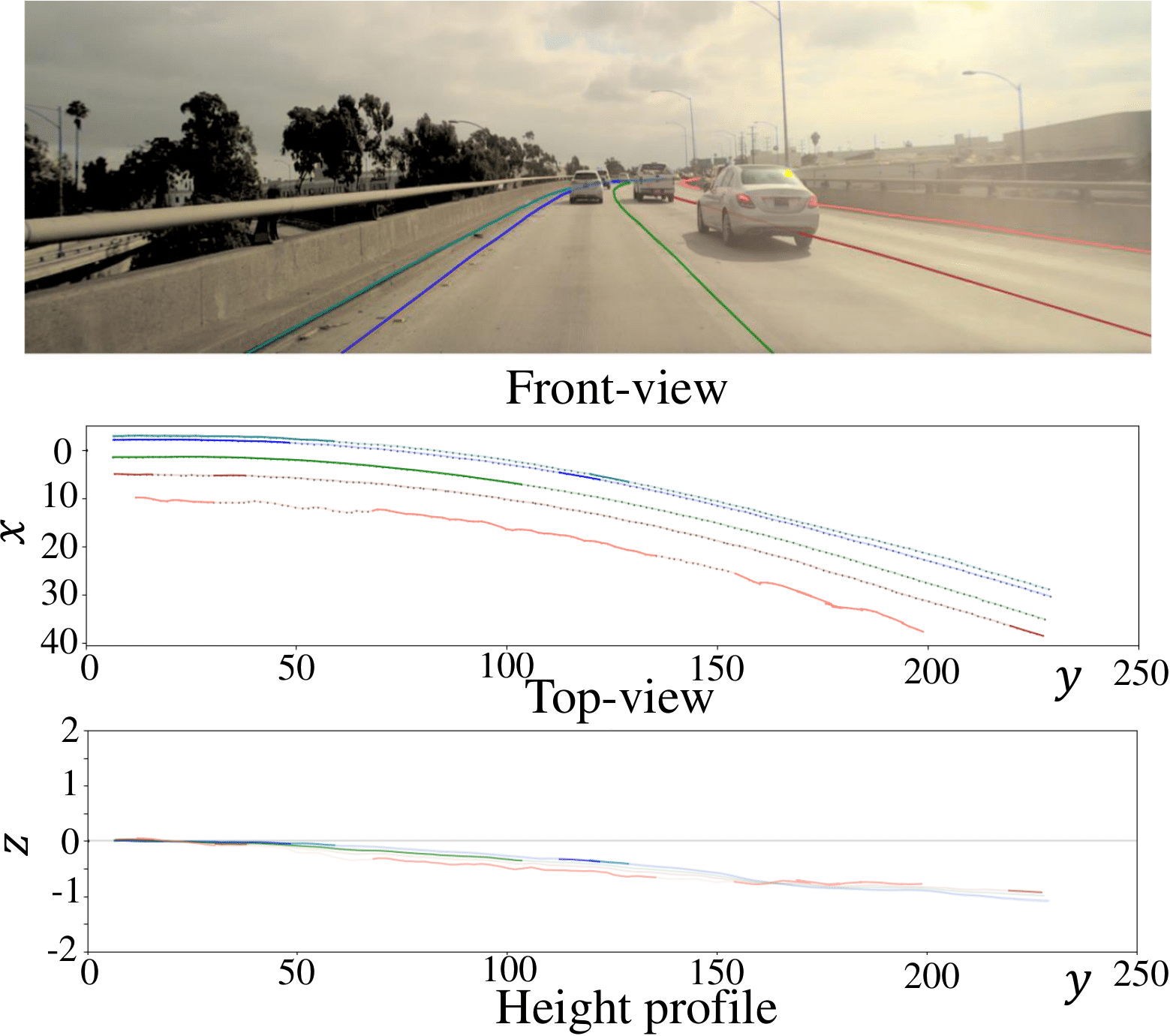}
		\caption{Our 3D lane dataset\label{fig:dataexamples2}}
	\end{subfigure}
	\caption{Examples from OpenLane compared to ours.}\label{fig:data_examples}
\end{figure}

\section{Auto-labeling and our 3D lane dataset}
\label{sec:data}
In this section, we describe our auto-labeling pipeline and provide details about our new 3D lane dataset. 

\subsection{Auto-labeling strategy}
\label{subsec:autlabeling}
While 3D lane datasets like OpenLane \cite{chen2022persformer} and ONCE-3DLanes \cite{once3dlanes} have contributed significantly to the field, they possess specific shortcomings that our auto-labeling strategy addresses. 
As illustrated in \figref{fig:dataexamples1} the LiDAR-based annotation approach, which recovers 3D information of 2D lane labels, introduces noise and outliers, particularly in the far range due to sparse scan points, while also posing calibration and synchronization challenges. To overcome these limitations, we propose an automated labeling approach (see \figref{fig:autolabeling}) that leverages confident near-range lane point estimates from a 2D detector, which are temporally accumulated in 3D space along an accurate ego-trajectory.
More precisely, common visual odometry methods \cite{5152255,steinbrucker2011real,agostinho2022practical} are applied to recover the vehicle ego-trajectory from the input video sequence. Our road surface generator then fits a spline surface to the vehicle's trajectory and orientations. On the other hand, lane pseudo-labels in image space are obtained from a state-of-the-art 2D lane detector \cite{linecnn,laneatt}, providing particularly reliable detection results in in the near-range. Lane and surface information are then combined in 3D by projecting near-range detection results to the road surface and accumulating line points frame by frame along the ego-trajectory of the sequence. Finally, we apply an accurate semantic segmentation network to automatically assign occlusion labels to each lane point.

\subsection{Dataset properties}
\label{subsec:dataproperties}
Using the approach described in \ref{subsec:autlabeling}, we are able to automatically generate temporally consistent, high-quality 3D lane labels in a range up to $250\,\mathrm{m}$ (see \figref{fig:dataexamples2}). Besides 3D lane geometry, we provide camera intrinsics and extrinsics, lane categories, lane visibility information and consistent track IDs. We further supply relational information about lanes (like line parallelism), local 3D ego-trajectory and global pose information for each frame. Important to mention is also that in contrast to former datasets, our visibility information contains occlusions caused by other traffic participants, enabling models to consistently learn lane structure independent of interruptions and lane occlusions. 
We refer to the supplementary material for additional information. 

\begin{figure}
	\centering
	\begin{subfigure}[b]{0.32\linewidth}
		\centering
		\includegraphics[width=1.\linewidth]{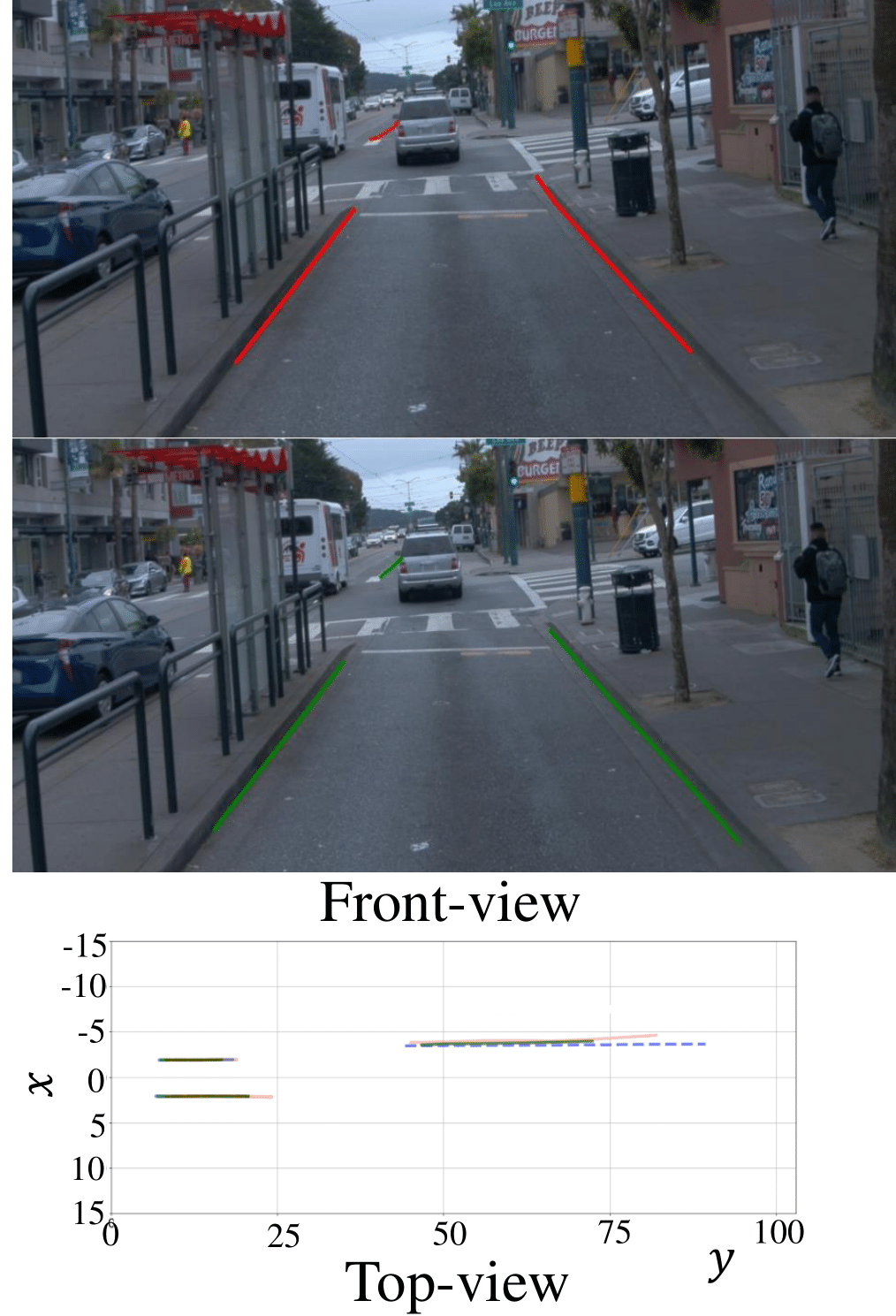}
		\caption{$t=1$ \label{fig:temp1}}
	\end{subfigure}
	\begin{subfigure}[b]{0.32\linewidth}
		\centering
		\includegraphics[width=0.98\linewidth]{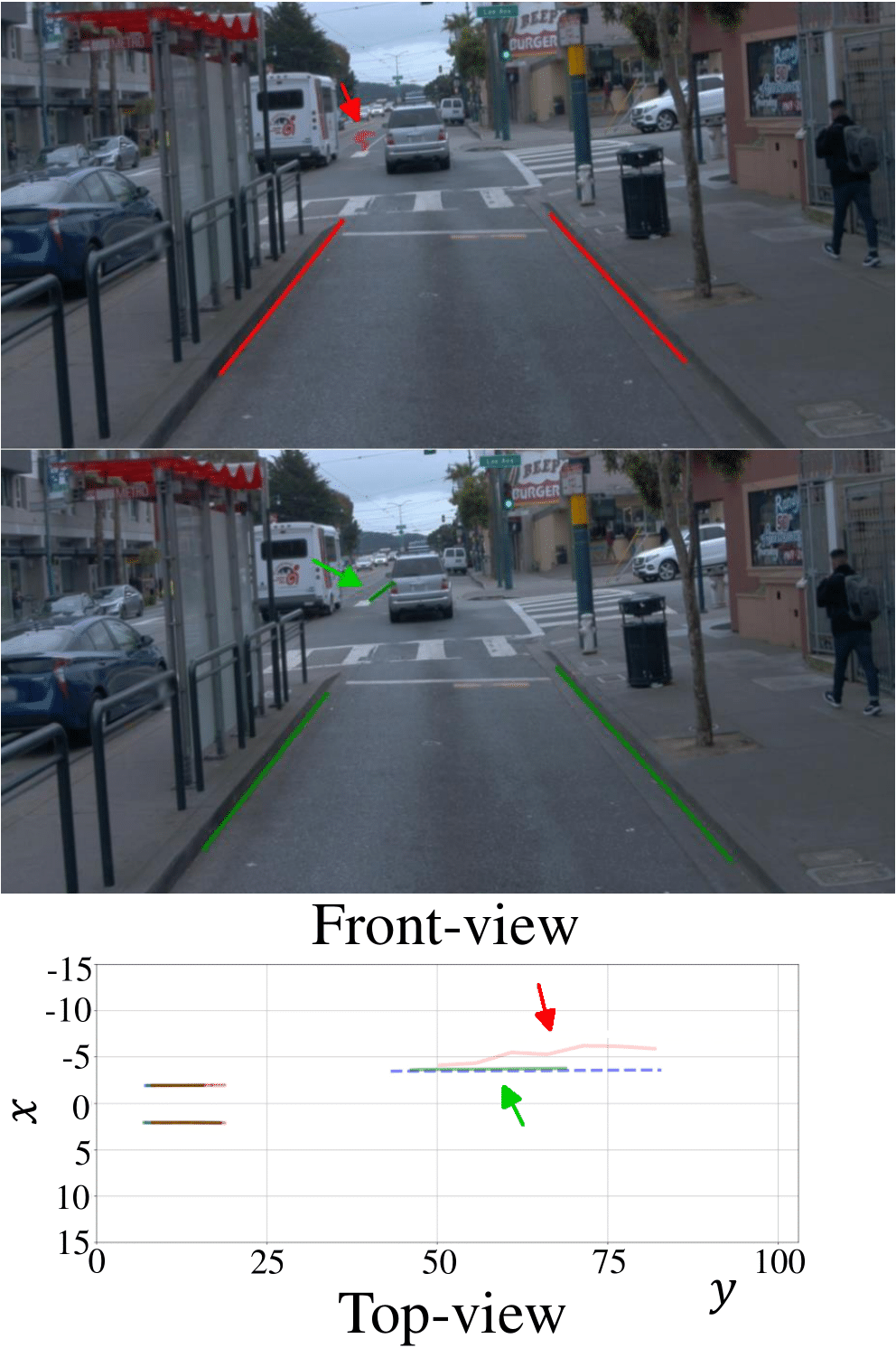}
		\caption{$t=2$ \label{fig:temp2}}
	\end{subfigure}
	\begin{subfigure}[b]{0.32\linewidth}
		\centering
		\includegraphics[width=0.981\linewidth]{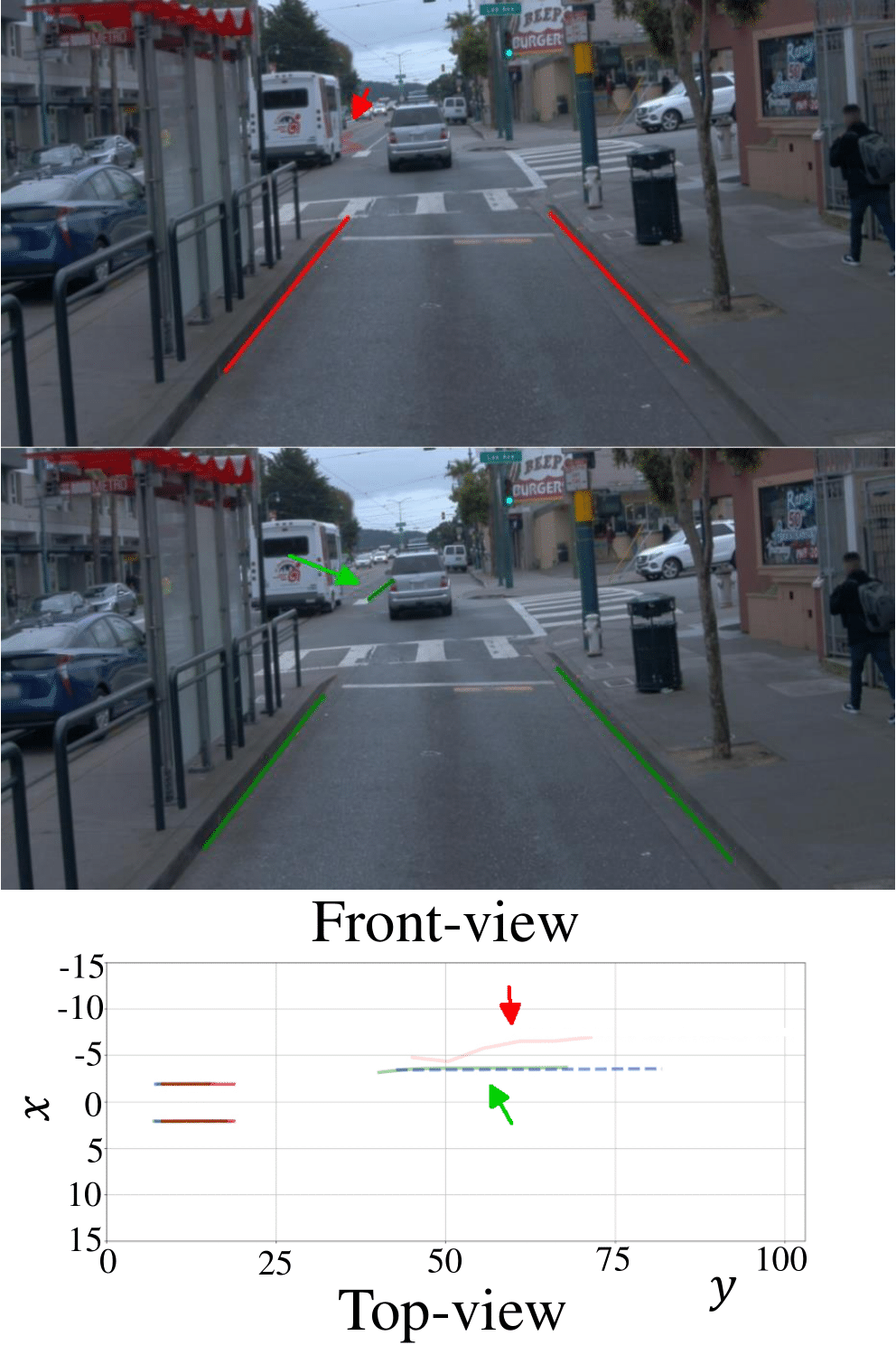}
		\caption{$t=3$ \label{fig:temp3}}
	\end{subfigure}
	\caption{Qualitative comparison of two models, one \textcolor{red}{without temporal} and one \textcolor{darkgreen}{with temporal} for three consecutive frames. \label{fig:temp} 
}
\end{figure}

\section{Experiments}
\label{sec:experiments}
In this section, we first outline our experimental setup, followed by a comprehensive analysis of our approach on two public and our own novel 3D lane dataset.

\subsection{Datasets and evaluation metrics}\label{subsec:datasets}
\noindent \textbf{OpenLane} \cite{chen2022persformer} based on the Waymo Open dataset \cite{sun2020scalability} remains the most prominent real-world 3D lane dataset to date. 
It is comprised of 200K images from 1000 sequences recorded in several cities in the USA. 

\noindent \textbf{ONCE-3DLanes} \cite{once3dlanes} is another real-world dataset derived from the ONCE dataset \cite{mao2021once}. It consists of 211K images collected from multiple cities in China. 

\noindent \textbf{Our 3D lane dataset}, which was introduced in \secref{sec:data}, was gathered from multiple regions and countries across the globe. 
It consists of 511K images with a  $90\,\%$ / $10\,\%$ train/test split. 

\noindent \textbf{Evaluation metrics.}  The OpenLane evaluation scheme \cite{genlanenet,chen2022persformer} uses the euclidean distance at uniformly distributed points and a range-IoU to compute the F1-Score. Moreover, x- and z-errors in near- ($0$-$40\,\mathrm{m}$) and far-range ($40$-$100\,\mathrm{m}$) evaluate the geometric accuracy. 
ONCE-3DLanes instead uses the uni-lateral chamfer distance (CD) and computes Precision (P), Recall (R) and F1-Score based on it. 

For our dataset, we extend the well-established OpenLane evaluation framework \cite{chen2022persformer,genlanenet}. Due to the long-range labels, we add x-/z-error intervals for the ranges $100$-$150\,\mathrm{m}$ and $150$-$200\,\mathrm{m}$. Since our dataset offers visibility labels for occlusions, we also add a visibility Intersection-over-Union metric (Vis-IoU). This score helps to gain understanding about the model's capability of distinguishing visible lane segments from parts occluded by other traffic participants and provides a measure for start-end range detection.

\subsection{Implementation details}\label{subsec:implementation}
We use an image input size $720 \times 960$ and ResNet-50 \cite{he2016deep} as our standard backbone. The query generator is trained on the auxiliary task of lane instance segmentation similar to \cite{luo2023latr}. The transformer decoder applied in the large model used for the evaluation in \secref{subsec:mainresults} consists of $L=6$ layers, whereas the lighter model used for ablation studies in Sec.~\ref{subsec:ablationstudies} has $2$ layers. The lane representation uses $M=20$ control points on OpenLane and ONCE-3DLanes and $30$ control points on our dataset due to the larger range. The number of lane proposals is set to $N=40$ for OpenLane and ours and $10$ for ONCE-3DLanes, which is sufficient for the small number of occurring lanes per frame. Regarding temporal fusion, our best model uses propagated queries from $T=3$ historic frames. 

\begin{table}
	\centering
	\resizebox{\columnwidth}{!}{
		\begin{tabular}{ccc||cc}
			\toprule
			\bf{Continuous Rep.} 	& \bf{STA} 				& \bf{Regularization}		& \bf{F1(\%)}$\uparrow$ 	& \bf{Gain(\%)} \\ 
			\hhline{=====}
			& & &  												$61.8$ & 	 	(baseline)				\\
			\checkmark & & &  									$62.9$ & 			$+1.1$ 				\\
			\checkmark & \checkmark & &  						$65.0$ & 			$+2.1$				\\
			\checkmark & \checkmark & \checkmark &  			$65.3$ & 			$+0.3$ 				\\
			\bottomrule
	\end{tabular}}
	\caption{Performance gain for different contributions on OpenLane using our novel \textbf{Continuous Lane Representation}, \textbf{Spatio-Temporal Attention (STA)} and \textbf{Regularization}.} \label{tab:benefits}
\end{table}
\begin{table}
	\centering
	\resizebox{\columnwidth}{!}{
		\begin{tabular}{l||c|c|c|c|c}
			\toprule
				{\bf{Attention}}		& {Global}			& {SLA}			& {PNA}			& {SLA + PNA}		& \bf{SLA + PNA + TCA}		\\
			\hline
			\bf{F1-Score} 							& $62.9$ 							& $62.8$							& $63.5$ 							& $63.8$						& $\bf{65.0}$		\\
			\bottomrule
	\end{tabular}}
	\caption{Analysis of spatio-temporal attention. Same line attention (SLA), parallel neighbor attention (PNA) and temporal cross-attention (TCA) and combinations compared to global attention. \label{tab:sta}}
\end{table}
\begin{table}
	\centering
	\resizebox{\columnwidth}{!}{
		\begin{tabular}{c||cccccc}
			\toprule
			\bf{Num. Frames ($T$)} 			& {0}		& {1} 		& {2} 		& \bf{3} 		& {4} 		& {5} 	\\ 
			\hline
			 \bf{F1(\%)}$\uparrow$		& $63.8	$	& $64.1$	& $64.6$ 	& $\bf{65.0}$	& $64.5$	& $63.9$	\\
			\bottomrule
	\end{tabular}}
	\caption{Analysis of number of frames in the memory for STA.} \label{tab:temporal}
\end{table}

\begin{table*}[tb]
		\centering
		\resizebox{\linewidth}{!}{
			\begin{tabular}{lll||ccccc|cccccc}
				\toprule
				\multirow{2}{*}{\bf{Method}} & \multirow{2}{*}{\bf{Backbone}}	& \multirow{2}{*}{\bf{Resolution}} & \multirow{2}{*}{\bf{F1-Score(\%)}$\uparrow$} & \bf{X-error} & \bf{X-error} & \bf{Z-error} & \bf{Z-error} & \multicolumn{6}{c}{\bf{F1-Score(\%) per Scenario} $\uparrow$} \\ 
				& & &  & \bf{near(m)}$\downarrow$ & \bf{far(m)}$\downarrow$ & \bf{near(m)}$\downarrow$ & \bf{far(m)}$\downarrow$ & \bf{U\&D} & \bf{C} & \bf{EW} & \bf{N} & \bf{I} & \bf{M\&S} \\
				\hhline{===||=====|======}
				3D-LaneNet \cite{3dlanenet} 								& VGG-16			& 360 x 480			& $44.1$ 			& $0.479$ 			& $0.572$				& $0.367$ 			& $0.443$ 			
																													& $40.8$ 			& $46.5$ 			& $47.5$				& $41.5$ 			& $32.1$ 			& $41.7$ 			\\
				Gen-LaneNet \cite{genlanenet} 								& ERFNet 			& 360 x 480			& $32.3$ 			& $0.591$ 			& $0.684$ 				& $0.411$ 			& $0.521$ 
																													& $25.4$  			& $33.5$ 			& $28.1$ 				& $18.7$ 			& $21.4$ 			& $31.0$ 			\\
				PersFormer \cite{chen2022persformer} 						& EfficientNet-B7 	& 360 x 480	 		& $50.5$ 			& $0.485$ 			& $0.553$ 				& $0.364$ 			& $0.431$ 
																													& $42.4$			& $55.6$ 			& $48.6$ 				& $46.6$ 			& $40.0$ 			& $50.7$ 			\\
				CurveFormer \cite{bai2023curveformer} 						& EfficientNet-B7 	& 360 x 480	 		& $50.5$ 			& $0.340$ 			& $0.772$ 				& $0.207$ 			& $0.651$ 			
																													& $45.2$ 			& $56.6$ 			& $49.7$ 				& $49.1$ 			& $42.9$ 			& $45.4$ 			\\
				BEV-LaneDet \cite{wang2023bev}  							& ResNet-34 		& 360 x 480	  		& $58.4$ 			& $0.309$ 			& $0.659$ 				& $0.244$ 			& $0.631$ 
																													& $48.7$ 			& $63.1$ 			& $53.4$ 				& $53.4$ 			& $50.3$ 			& $53.7$ 			\\
				Anchor3DLane-T \cite{huang2023anchor3dlane} 				& ResNet-18 		& 360 x 480			& $54.3$ 			& $0.275$ 			& $0.310$				& $0.105$ 			& $0.135$ 
																													& $47.2$			& $58.0$ 			& $52.7$ 				& $48.7$ 			& $45.8$ 			& $51.7$ 			\\
				PETRv2 \cite{liu2023petrv2}									& VoV-99			& 360 x 480			& $61.2$ 			& $0.400$ 			& $0.573$ 				& $0.265$ 			& $0.413$ 
																													& $-$				& $-$ 				& $-$					& $-$				& $-$				& $-$				\\
				LATR \cite{luo2023latr} 									& ResNet-50 		& 720 x 960	 		& $61.9$ 			& $\underline{0.219}$& $0.259$ 				& $\underline{0.075}$& $\underline{0.104}$ 			
																													& $55.2$ 			& $\underline{68.2}$& $57.1$ 				& $55.4$ 			& $52.3$ 			& $\underline{61.5}$ 			\\
				LaneCPP \cite{pittner2024lanecpp}							& EfficientNet-B7 	& 360 x 480 		& $60.3$ 			& $0.264$ 			& $0.310$ 				& $0.077$			& $0.117$ 			
																													& $53.6$ 			& $64.4$			& $56.7$	 			& $54.9$			& $52.0$ 			& $58.7$			\\ 
				PVALane \cite{zheng2024pvalane} 							& Swin-B 			& 720 x 960  		& $63.4$ 			& $0.226$ 			& $\underline{0.257}$	& $0.093$ 			& $0.119$ 			
																													& $\underline{56.1}$& $67.7$ 			& $\mathbf{64.0}$ 		& $\mathbf{58.6}$ 	& $\underline{53.6}$& $60.8$ 			\\
				GroupLane \cite{li2024grouplane} 							& ConvNext-Base 	& not specified 	& $\underline{64.1}$& $0.320$  			& $0.441$ 				& $0.233$ 			& $0.402$ 		
																													& $-$ 				& $-$ 				& $-$ 					& $-$ 				& $-$ 				& $-$ 				\\
				\rowcolor{Gray} SparseLaneSTP (Ours) 						& ResNet-50 		& 720 x 960 		& $\mathbf{66.1}$ 	& $\mathbf{0.203}$	& $\mathbf{0.240}$ 		& $\mathbf{0.066}$	& $\mathbf{0.092}$ 	
																													& $\mathbf{57.3}$	& $\mathbf{73.0}$	& $\underline{60.1}$	& $\underline{58.3}$& $\mathbf{58.2}$	& $\mathbf{66.5}$	\\
				\bottomrule
		\end{tabular}}
		\caption{Quantitative comparison on OpenLane \cite{chen2022persformer}. \textbf{Best performance} and \underline{second best} are highlighted. 
}
		\label{tab:comparison-quant-ol}
\end{table*}
\begin{figure}[tb]
       \centering
       \includegraphics[width=.7\linewidth]{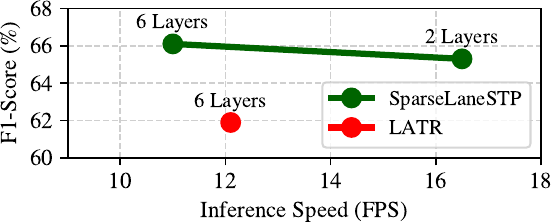}
       \caption{Efficiency analysis of SparseLaneSTP.}
       \label{fig:runtime}
\end{figure}

\begin{figure*}
	\centering
	\begin{subfigure}[b]{0.245\linewidth}
		\centering
		\includegraphics[width=1.05\linewidth]{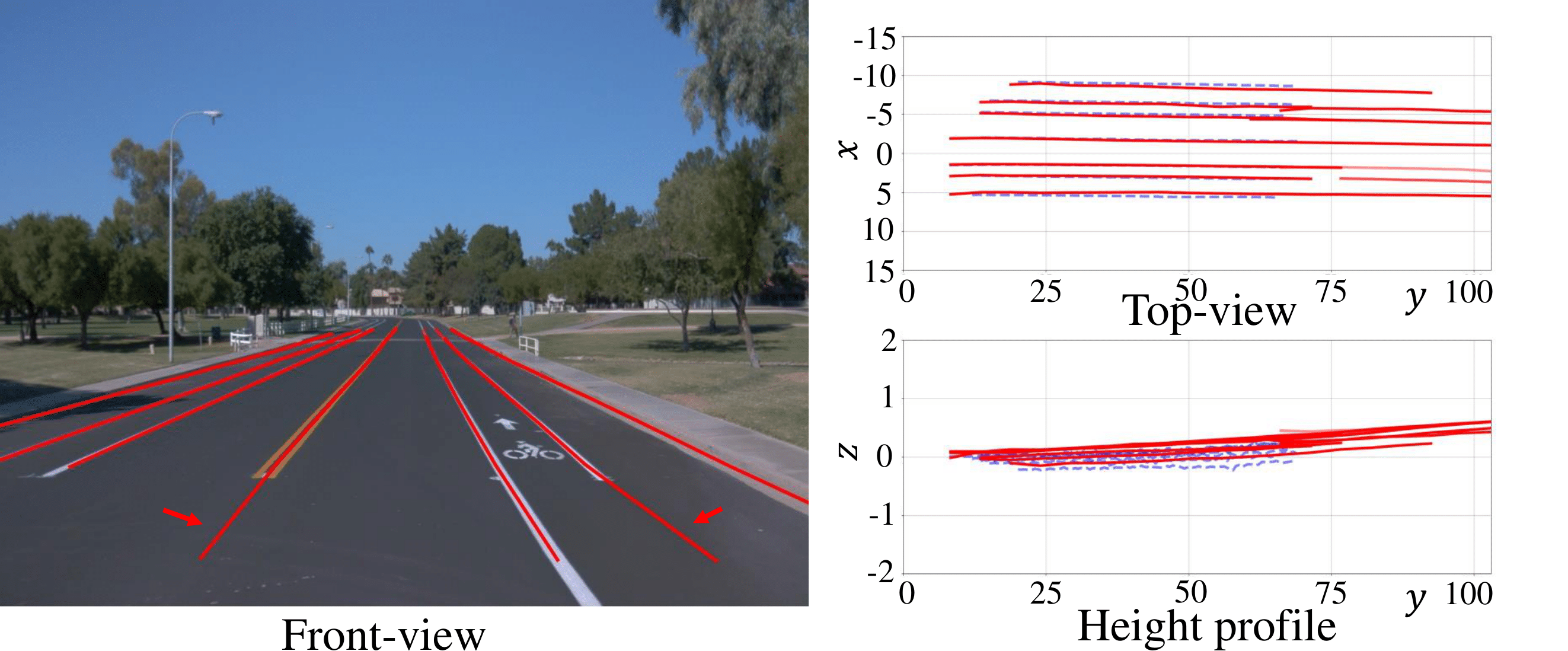}
		\includegraphics[width=1.05\linewidth]{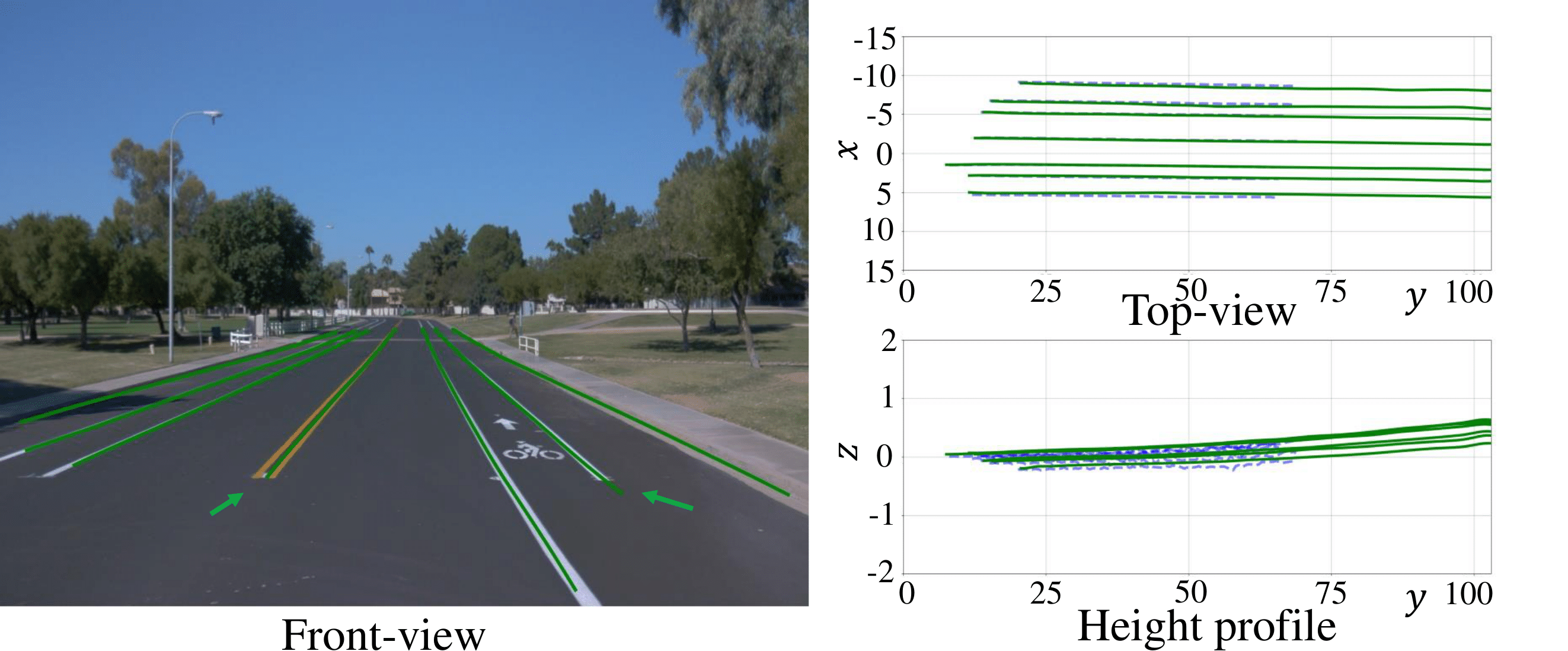}
		\caption{\label{fig:comparison1}}
	\end{subfigure}
	\begin{subfigure}[b]{0.245\linewidth}
		\centering
		\includegraphics[width=1.05\linewidth]{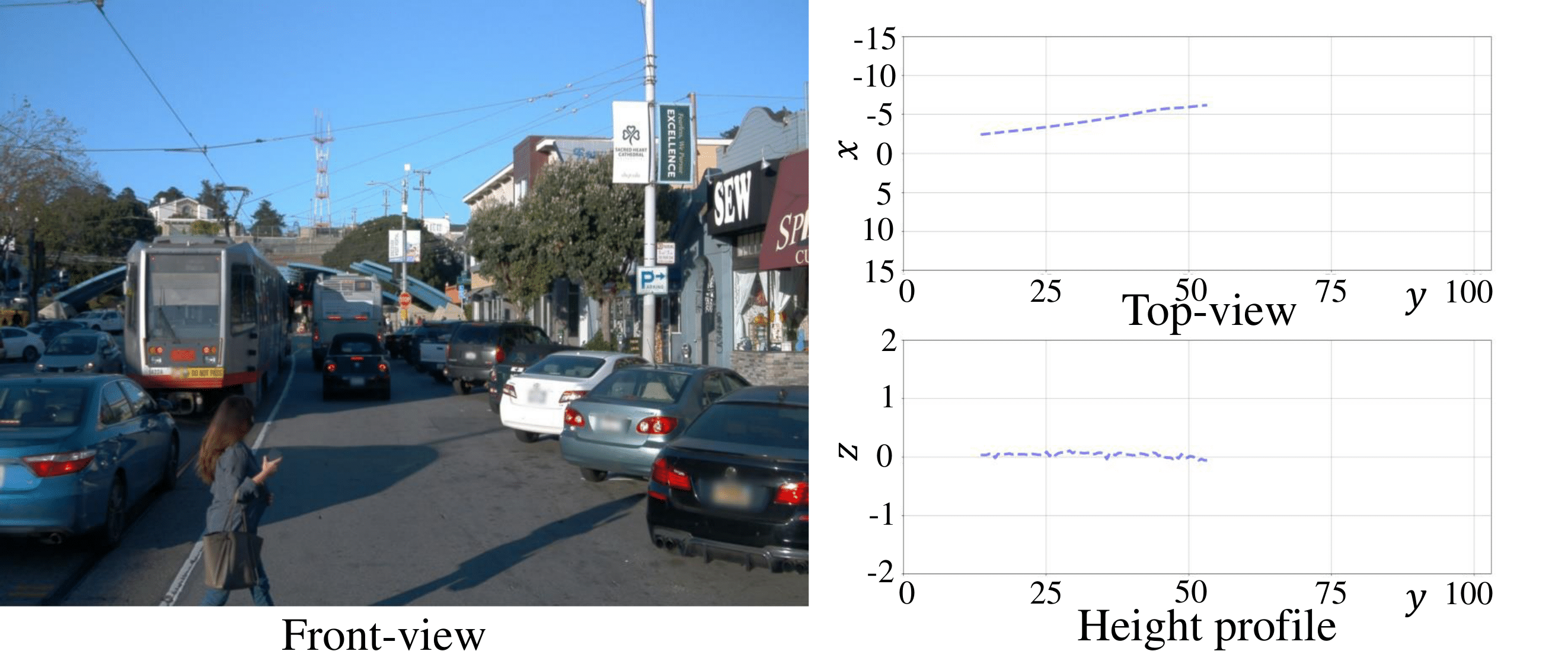}
		\includegraphics[width=1.05\linewidth]{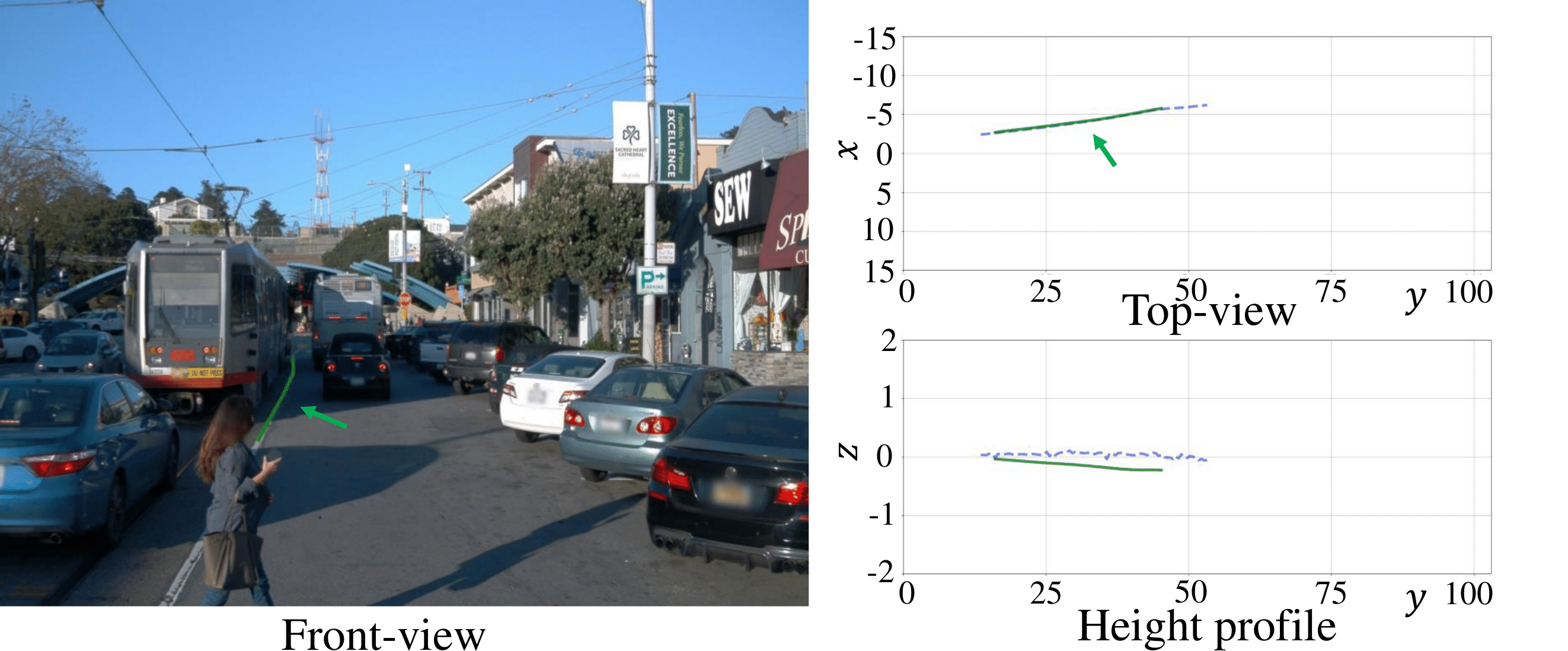}
		\caption{\label{fig:comparison2}}
	\end{subfigure}
	\begin{subfigure}[b]{0.245\linewidth}
		\centering
		\includegraphics[width=1.05\linewidth]{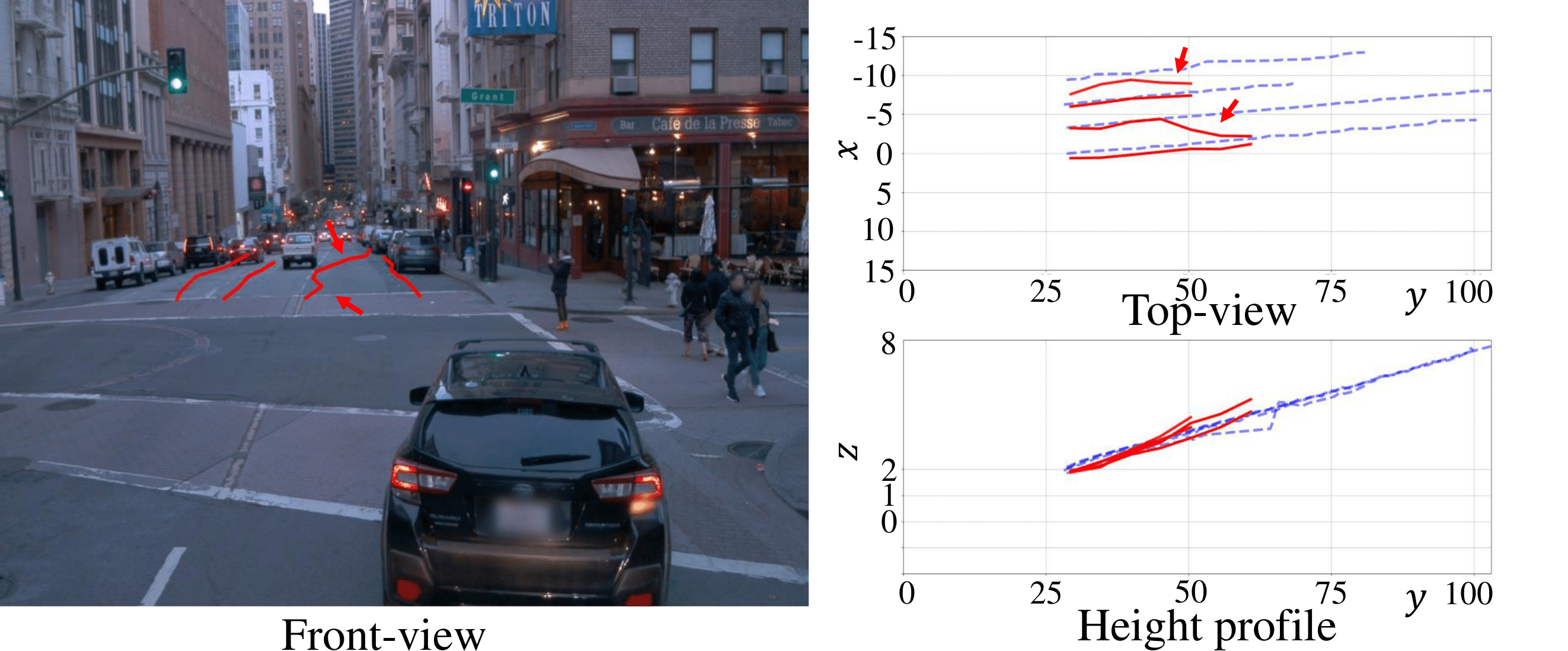}
		\includegraphics[width=1.05\linewidth]{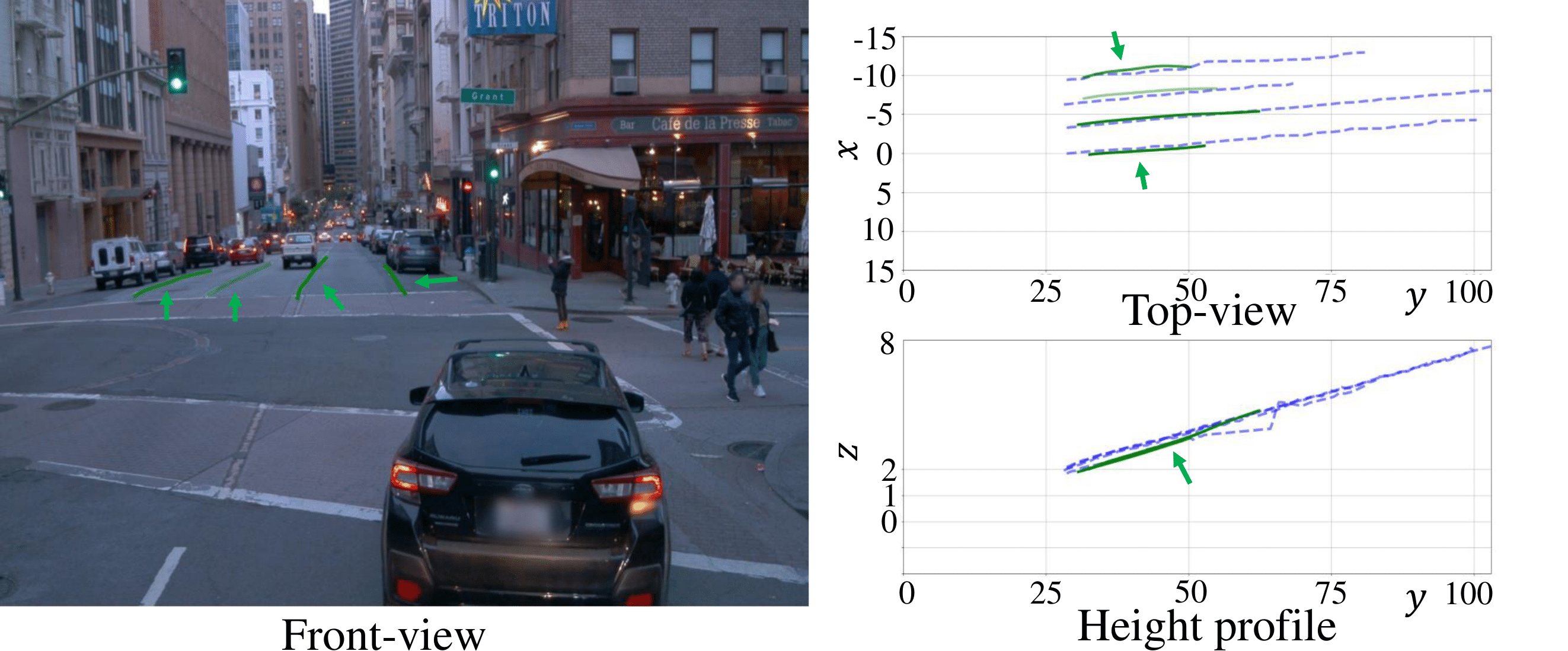}
		\caption{\label{fig:comparison3}}
	\end{subfigure}
	\begin{subfigure}[b]{0.245\linewidth}
		\centering
		\includegraphics[width=1.05\linewidth]{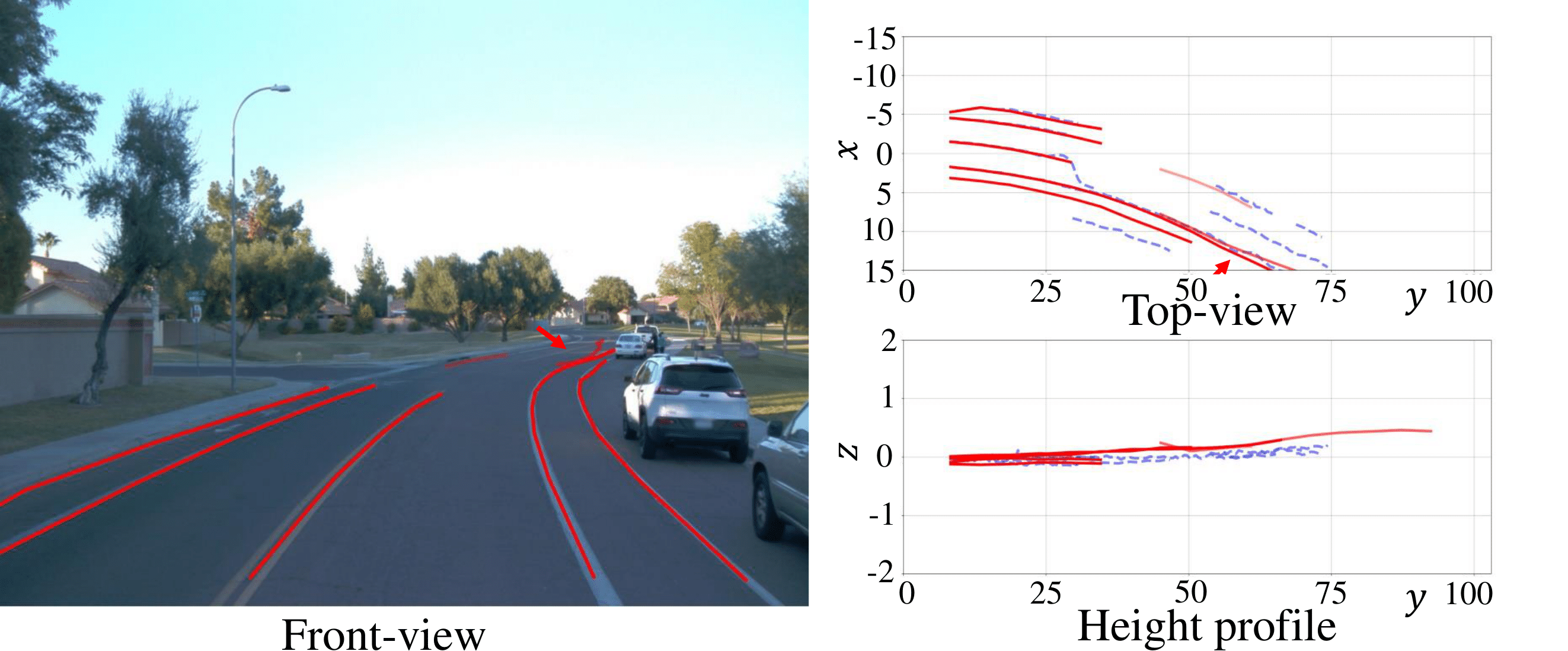}
		\includegraphics[width=1.05\linewidth]{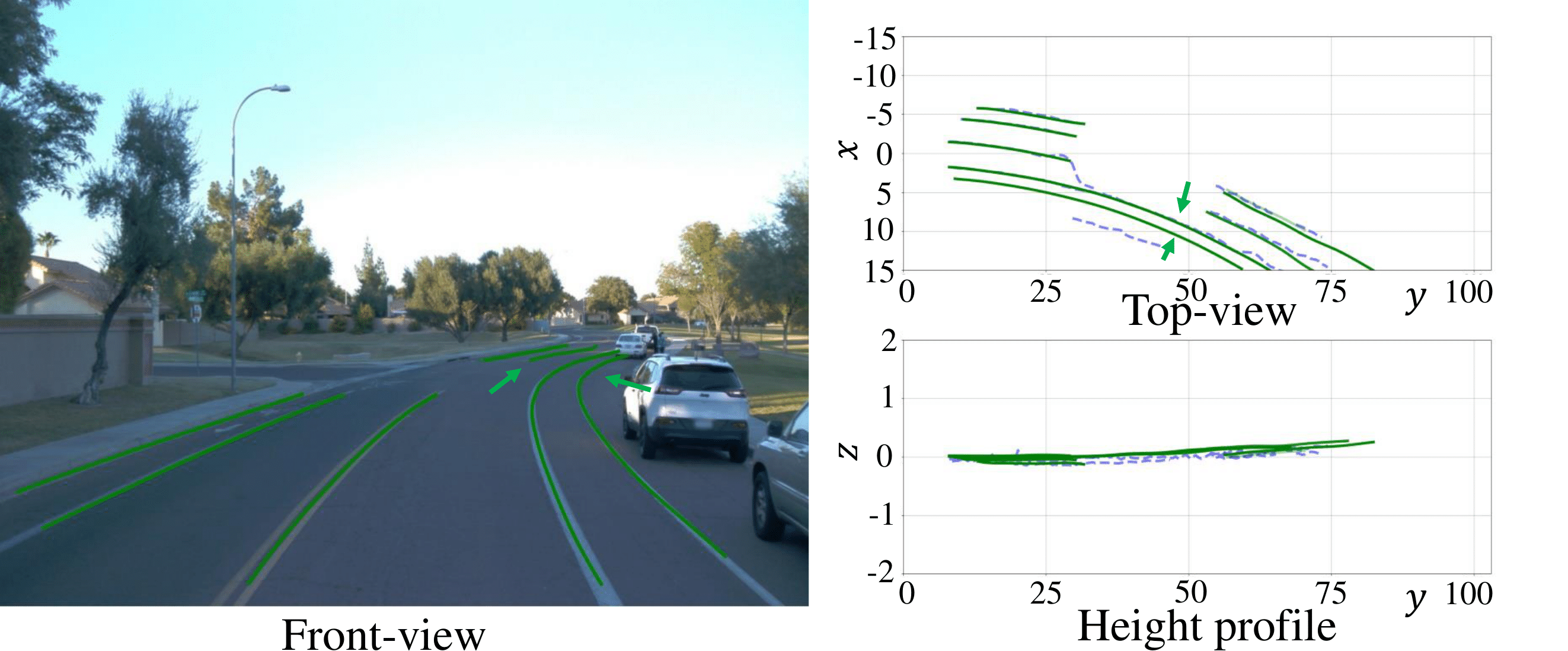}
		\caption{\label{fig:comparison4}}
	\end{subfigure}
	\caption{Qualitative comparison of \textcolor{red}{LATR} and \textcolor{darkgreen}{SparseLaneSTP} on OpenLane with \textcolor{blue}{ground truth}. Main differences are highlighted. 
\label{fig:comparison}}
\end{figure*}
\begin{table*}[tb]
	\centering
	\resizebox{\linewidth}{!}{
		\begin{tabular}{l||c|cccc|c}
			\toprule
			\multirow{2}{*}{\bf{Method}}  	& \multirow{2}{*}{\bf{F1-Score(\%)}$\uparrow$} 	& \multicolumn{4}{c|}{\bf{X-error / Z-error (m)}$\downarrow$} & \multirow{2}{*}{\bf{Vis-IoU(\%)$\uparrow$}}  \\ 
			&  																		& \bf{0~m - 40~m} 	& \bf{40~m - 100~m} 	& \bf{100~m - 150~m} 	& \bf{150~m - 200~m} &  \\
			\hhline{=||=|====|=}
			PersFormer \cite{chen2022persformer}	 		& $59.2$ 				& $0.191$ / $0.072$								& $0.344$ / $0.145$						& $0.671$ / $0.220$ 							& $0.845$ / $0.379$ 						& $70.1$				\\
			LaneCPP \cite{pittner2024lanecpp}				& $63.1$ 				& $0.144$ / $0.040$								& $0.298$ / $\underline{0.093}$			& $0.494$ / $0.188$ 							& $0.761$ / $0.293$  						& $77.5$				\\			
			Baseline (LATR) \cite{luo2023latr} 				& $\underline{65.1}$ 	& $\underline{0.122}$ / $\underline{0.038}$		& $\underline{0.263}$ / $0.098$			& $\underline{0.468}$ / $\underline{0.185}$ 	& $\underline{0.700}$ / $\underline{0.264}$ & $\underline{78.3}$	\\
			\rowcolor{Gray} \textbf{SparseLaneSTP (ours)} 	& $\mathbf{68.2}$ 		& $\mathbf{0.109}$ / $\mathbf{0.033}$ 			& $\mathbf{0.233}$ / $\mathbf{0.092}$ 	& $\mathbf{0.443}$ / $\mathbf{0.182}$ 			&  $\mathbf{0.646}$ / $\mathbf{0.263}$  	& $\mathbf{81.4}$		\\
			\bottomrule
	\end{tabular}}
	\caption{Quantitative comparison on our 3D lane dataset. \textbf{Best performance} and \underline{second best} are highlighted. 
}
\label{tab:ourdataset}
\end{table*}

\begin{table}[tb]
	\centering
	\resizebox{\columnwidth}{!}{
		\begin{tabular}{l||cccc}
			\toprule
			\bf{Method} 										& \bf{F1(\%)}$\uparrow$ & \bf{P(\%)}$\uparrow$ 	& \bf{R(\%)}$\uparrow$ 	& \bf{CD(m)}$\downarrow$	\\ 
			\hhline{=||====}
			3D-LaneNet \cite{3dlanenet} 						& $44.73$ 				& $61.46$ 				& $35.16$				& $0.127$ 					\\
			Gen-LaneNet \cite{genlanenet} 						& $45.59$ 				& $63.95$ 				& $35.42$ 				& $0.121$  					\\
			SALAD \cite{once3dlanes} 							& $64.07$ 				& $75.90$				& $55.42$ 				& $0.098$ 					\\
			PersFormer \cite{chen2022persformer} 				& $74.33$ 				& $80.30$				& $69.18$ 				& $0.074$ 					\\
			Anchor3DLane \cite{huang2023anchor3dlane} 			& $74.87$ 				& $80.85$ 				& $69.71$ 				& $0.060$ 					\\
			LATR \cite{luo2023latr} 							& $80.59$ 				& $\underline{86.12}$ 	& $75.73$ 				& $\underline{0.052}$ 		\\
			PVALane \cite{zheng2024pvalane} 					& $76.35$ 				& $82.81$ 				& $70.83$ 				& $0.059$					\\
			GroupLane \cite{li2024grouplane} 					& $\underline{80.73}$ 	& $82.56$  				& $\underline{78.90}$ 	& $0.053$ 					\\ 
			\rowcolor{Gray} \textbf{SparseLaneSTP (ours)} 		& $\mathbf{82.75}$    	& $\mathbf{86.47}$      & $\mathbf{79.33}$ 		& $\mathbf{0.048}$	        \\
			\bottomrule
	\end{tabular}}
	\caption{Quantitative comparison on ONCE-3DLanes \cite{once3dlanes}. 
} 
\label{tab:once}
\end{table}

\subsection{Ablation studies}\label{subsec:ablationstudies}
We perform a comprehensive analysis validating the effectiveness of our design choices in SparseLaneSTP. For these experiments, we used a smaller model consisting of two decoder layers evaluated on OpenLane. 

\tabref{tab:benefits} summarizes the impact of our contributions. Our CR spline based lane representation improves the F1-Score by over $1$ percentage point compared to the discrete one (baseline). 
The most significant gain in F1-Score is achieved by the novel STA component, which encourages focus on relevant relations based on lane structure and, more importantly, on related past observations by integrating temporal cross-attention. 
Eventually, regularization enhances robustness and consistency, leading to improved generalization, as evidenced by the additional F1-Score gain. 

As shown in \tabref{tab:sta}, global self-attention apparently fails to capture important relations of lane structures. The combination of our proposed SLA and PNA demonstrates the benefits of enhancing the focus on intra- and inter-lane relations by an F1-Score increase $+0.9\,\%$. 
Integrating knowledge about past observations via our suggested TCA yields the greatest improvement of additional $+1.2\,\%$. As indicated by \tabref{tab:temporal}, $T=3$ frames in the memory produces the best results. Using too few frames provides insufficient temporal context, whereas too many frames apparently introduce redundancy and distract the model from relevant information. 
We further qualitatively demonstrate the beneficial impact of our primary contribution on temporal fusion in \figref{fig:temp}. 
The non-temporal model fails to retain the detection from frame $t=1$ in subsequent frames as lane visibility deteriorates. In contrast, the temporally-aware model maintains the detection despite partial occlusions, demonstrating improved robustness. Regularization further preserves temporal consistency mitigating the gradual disappearing and drifting in the non-temporal model.

Finally, \figref{fig:runtime} reveals efficiency benefits of our approach. SparseLaneSTP with six decoder layers achieves $11.0$ FPS, comparable to LATR at $12.1$ FPS. While our proposed temporal integration (memory queue and STA) introduces a modest $9\,\%$ overhead, it also enables a smaller model with only two layers to reach $65.3\,\%$ F1-Score ($+3.4\,\%$ over six-layer LATR) at a significantly higher speed of $16.5$ FPS.

\subsection{Main results}\label{subsec:mainresults}
In this section, we compare our method to other 3D lane detectors on public benchmarks as well as on our own dataset. 

\noindent \textbf{Results on OpenLane.} \tabref{tab:comparison-quant-ol} presents a quantitative comparison demonstrating the superiority of our approach. SparseLaneSTP reaches the highest F1-Score, exceeding $66\,\%$, significantly surpassing the second-best model despite its larger backbone and even achieves the lowest geometric errors. 

\figref{fig:comparison} illustrates the advantages of our approach qualitatively across multiple aspects. We compare our SparseLaneSTP to LATR \cite{luo2023latr}, which is most related to our baseline. As shown in \figref{fig:comparison1}, our continuous lane representation enables precise visibility estimation with accurate detection of start/end of markings, whereas LATR - constrained by fixed discrete anchor points - produces erroneous estimates. In \figref{fig:comparison2}, our temporal fusion preserves the line despite partial occlusions, whereas non-temporal LATR fails. It also struggles to estimate faint lane markings on the far side of the junction shown in \figref{fig:comparison3}, while with STA our model captures lane structures more effectively. Eventually, in \figref{fig:comparison4} our model exhibits a stronger reliance on priors resulting in robust and consistent behavior, most likely due to the incorporated regularization.

\noindent \textbf{Results on ONCE-3DLanes.} 
As shown in \tabref{tab:once}, our method achieves the highest detection scores and demonstrates accurate regression with a significantly lower CD, despite the inaccurate camera parameters provided by the dataset. 
Notably, even though ONCE-3DLanes has a lower frame rate, our temporal approach performs effectively. 

\begin{figure}
	\centering
		\includegraphics[width=0.9\linewidth]{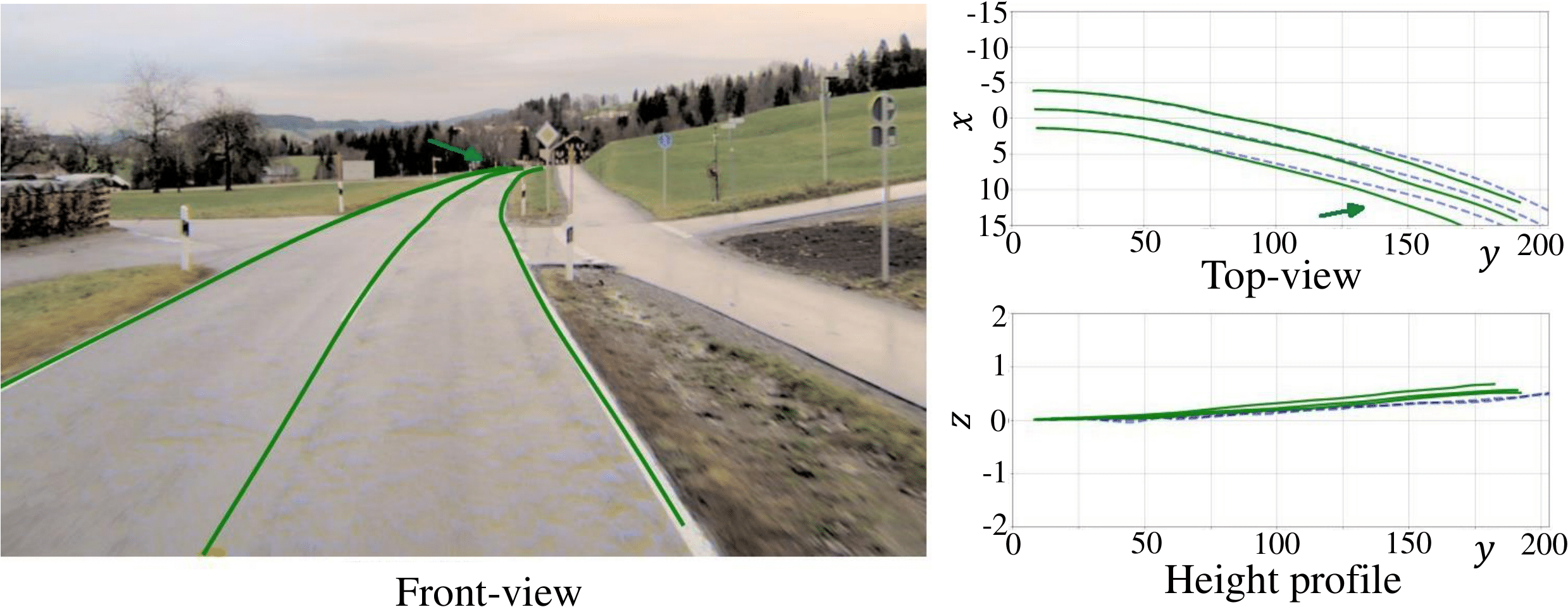}
	\caption{Qualitative example of \textcolor{darkgreen}{SparseLaneSTP} on our data.  
} \label{fig:qualitativeours}
\end{figure}

\noindent \textbf{Results on our 3D lane dataset.} 
Initial results on our novel dataset are presented in \tabref{tab:ourdataset}. Our proposed extension of the evaluation metrics provides a more comprehensive assessment of overall performance, particularly in the far-range and for occlusion detection. To facilitate a fair comparison, we train the baseline and other methods with specific modifications to range and capacity. Our method excels in F1-Score, error metrics and notably in the Vis-IoU, highlighting the superior performance of our temporal model in detecting occlusions. \figref{fig:qualitativeours} shows a qualitative result on our dataset, showcasing accurate long-range lane estimation ($200\,\mathrm{m}$). For more qualitative results with comparisons to the baseline we refer to the supplementary.

\section{Conclusion}
\label{sec:conclusion}
In this work, we presented a novel 3D lane detection approach that combines sparse lane transformers with spatial and temporal priors. 
We proposed a continuous lane representation, tailored for the sparse query design, thereby overcoming limitations of discrete representations. 
Additionally, we developed a novel attention layer that effectively captures intra- and inter-lane relations while integrating temporal information through propagated lane queries. Furthermore, we introduced a dedicated regularization scheme enhancing temporal consistency.
Our experiments validate the effective integration of temporal information and spatial priors, as well as the benefits of our lane representation, 
leading to state-of-the-art performance across all evaluated 3D lane datasets. 
Moreover, we presented a new 3D lane dataset, constructed using a simple yet effective auto-labeling pipeline, providing geometrically accurate labels that span long distances and explicitly annotate occlusions, fostering further advancements in the field. 
In future, we will explore an extension to 3D lane tracking to leverage the full potential of temporal information. 
{
	\small
	\bibliographystyle{ieee_fullname}
	\bibliography{survey_bib_wo_url}

\begin{thebibliography}{10}\itemsep=-1pt

\bibitem{agostinho2022practical}
Lucas~R Agostinho, Nuno~M Ricardo, Maria~I Pereira, Antoine Hiolle, and Andry~M
  Pinto.
\newblock A practical survey on visual odometry for autonomous driving in
  challenging scenarios and conditions.
\newblock {\em IEEE Access}, 2022.

\bibitem{bai2023curveformer}
Yifeng Bai, Zhirong Chen, Zhangjie Fu, Lang Peng, Pengpeng Liang, and Erkang
  Cheng.
\newblock Curveformer: 3d lane detection by curve propagation with curve
  queries and attention.
\newblock In {\em Proc. IEEE International Conf. on Robotics and Automation
  (ICRA)}, 2023.

\bibitem{carion2020detr}
Nicolas Carion, Francisco Massa, Gabriel Synnaeve, Nicolas Usunier, Alexander
  Kirillov, and Sergey Zagoruyko.
\newblock End-to-end object detection with transformers.
\newblock In {\em Proc. of the European Conf. on Computer Vision (ECCV)}, 2020.

\bibitem{CRsplines}
Edwin Catmull and Raphael Rom.
\newblock A class of local interpolating splines.
\newblock {\em Computer aided geometric design}, 1972.

\bibitem{chang2025rethinking}
Yifan Chang, Junjie Huang, Xiaofeng Wang, Yun Ye, Zhujin Liang, Yi Shan, Dalong
  Du, and Xingang Wang.
\newblock Rethinking lanes and points in complex scenarios for monocular 3d
  lane detection.
\newblock In {\em Proceedings of the Computer Vision and Pattern Recognition
  Conference}, 2025.

\bibitem{chen2022persformer}
Li Chen, Chonghao Sima, Yang Li, Zehan Zheng, Jiajie Xu, Xiangwei Geng,
  Hongyang Li, Conghui He, Jianping Shi, Yu Qiao, et~al.
\newblock Persformer: 3d lane detection via perspective transformer and the
  openlane benchmark.
\newblock In {\em Proc. of the European Conf. on Computer Vision (ECCV)}, 2022.

\bibitem{dao2023flashattention}
Tri Dao.
\newblock Flashattention-2: Faster attention with better parallelism and work
  partitioning.
\newblock {\em arXiv:2307.08691}, 2023.

\bibitem{dao2022flashattention}
Tri Dao, Dan Fu, Stefano Ermon, Atri Rudra, and Christopher R{\'e}.
\newblock Flashattention: Fast and memory-efficient exact attention with
  io-awareness.
\newblock {\em Advances in Neural Information Processing Systems (NeuIPS)},
  2022.

\bibitem{deboor197250}
Carl {de Boor}.
\newblock On calculating with b-splines.
\newblock {\em Journal of Approximation Theory}, 1972.

\bibitem{curvemodeling}
Zhengyang Feng, Shaohua Guo, Xin Tan, Ke Xu, Min Wang, and Lizhuang Ma.
\newblock Rethinking efficient lane detection via curve modeling.
\newblock In {\em Proc. IEEE Conf. on Computer Vision and Pattern Recognition
  (CVPR)}, 2022.

\bibitem{3dlanenet}
Noa Garnett, Rafi Cohen, Tomer Pe'er, Roee Lahav, and Dan Levi.
\newblock 3d-lanenet: End-to-end 3d multiple lane detection.
\newblock In {\em Proc. of the IEEE International Conf. on Computer Vision
  (ICCV)}, 2019.

\bibitem{ghafoorian2018gan}
Mohsen Ghafoorian, Cedric Nugteren, N{\'{o}}ra Baka, Olaf Booij, and Michael
  Hofmann.
\newblock {EL-GAN:} embedding loss driven generative adversarial networks for
  lane detection.
\newblock In {\em Proc. of the European Conf. on Computer Vision (ECCV)}, 2018.

\bibitem{genlanenet}
Yuliang Guo, Guang Chen, Peitao Zhao, Weide Zhang, Jinghao Miao, Jingao Wang,
  and Tae~Eun Choe.
\newblock Gen-lanenet: {A} generalized and scalable approach for 3d lane
  detection.
\newblock In {\em Proc. of the European Conf. on Computer Vision (ECCV)}, 2020.

\bibitem{he2016deep}
Kaiming He, Xiangyu Zhang, Shaoqing Ren, and Jian Sun.
\newblock Deep residual learning for image recognition.
\newblock In {\em Proc. IEEE Conf. on Computer Vision and Pattern Recognition
  (CVPR)}, 2016.

\bibitem{lightweightld}
Yuenan Hou, Zheng Ma, Chunxiao Liu, and Chen~Change Loy.
\newblock Learning lightweight lane detection cnns by self attention
  distillation.
\newblock In {\em Proc. of the IEEE International Conf. on Computer Vision
  (ICCV)}, 2019.

\bibitem{huang2022bevdet4d}
Junjie Huang and Guan Huang.
\newblock Bevdet4d: Exploit temporal cues in multi-camera 3d object detection.
\newblock {\em CoRR}, abs/2203.17054, 2022.

\bibitem{huang2021bevdet}
Junjie Huang, Guan Huang, Zheng Zhu, and Dalong Du.
\newblock Bevdet: High-performance multi-camera 3d object detection in
  bird-eye-view.
\newblock {\em CoRR}, abs/2112.11790, 2021.

\bibitem{huang2023anchor3dlane}
Shaofei Huang, Zhenwei Shen, Zehao Huang, Zi han Ding, Jiao Dai, Jizhong Han,
  Naiyan Wang, and Si Liu.
\newblock Anchor3dlane: Learning to regress 3d anchors for monocular 3d lane
  detection.
\newblock In {\em Proc. IEEE Conf. on Computer Vision and Pattern Recognition
  (CVPR)}, 2023.

\bibitem{huval2015empirical}
Brody Huval, Tao Wang, Sameep Tandon, Jeff Kiske, Will Song, Joel
  Pazhayampallil, Mykhaylo Andriluka, Pranav Rajpurkar, Toki Migimatsu, Royce
  Cheng{-}Yue, Fernando~A. Mujica, Adam Coates, and Andrew~Y. Ng.
\newblock An empirical evaluation of deep learning on highway driving.
\newblock {\em arXiv/1504.01716}, 2015.

\bibitem{pinet}
YeongMin Ko, Jiwon Jun, Donghwuy Ko, and Moongu Jeon.
\newblock Key points estimation and point instance segmentation approach for
  lane detection.
\newblock {\em arXiv/2002.06604}, 2020.

\bibitem{vpgnet}
Seokju Lee, Junsik Kim, Jae~Shin Yoon, Seunghak Shin, Oleksandr Bailo, Namil
  Kim, Tae{-}Hee Lee, Hyun~Seok Hong, Seung{-}Hoon Han, and In~So Kweon.
\newblock Vpgnet: Vanishing point guided network for lane and road marking
  detection and recognition.
\newblock In {\em Proc. of the IEEE International Conf. on Computer Vision
  (ICCV)}, 2017.

\bibitem{li2022reconstruct}
Chenguang Li, Jia Shi, Ya Wang, and Guangliang Cheng.
\newblock Reconstruct from top view: A 3d lane detection approach based on
  geometry structure prior.
\newblock In {\em Proc. IEEE Conf. on Computer Vision and Pattern Recognition
  (CVPR)}, 2022.

\bibitem{linecnn}
Xiang Li, Jun Li, Xiaolin Hu, and Jian Yang.
\newblock Line-cnn: End-to-end traffic line detection with line proposal unit.
\newblock {\em IEEE Trans. on Intelligent Transportation Systems (T-ITS)},
  2020.

\bibitem{li2024grouplane}
Zhuoling Li, Chunrui Han, Zheng Ge, Jinrong Yang, En Yu, Haoqian Wang, Xiangyu
  Zhang, and Hengshuang Zhao.
\newblock Grouplane: End-to-end 3d lane detection with channel-wise grouping.
\newblock {\em IEEE Robotics and Automation Letters (RA-L)}, 2024.

\bibitem{li2022bevformer}
Zhiqi Li, Wenhai Wang, Hongyang Li, Enze Xie, Chonghao Sima, Tong Lu, Yu Qiao,
  and Jifeng Dai.
\newblock Bevformer: Learning bird's-eye-view representation from multi-camera
  images via spatiotemporal transformers.
\newblock In {\em Proc. of the European Conf. on Computer Vision (ECCV)}, 2022.

\bibitem{liao2023maptr}
Bencheng Liao, Shaoyu Chen, Xinggang Wang, Tianheng Cheng, Qian Zhang, Wenyu
  Liu, and Chang Huang.
\newblock Maptr: Structured modeling and learning for online vectorized {HD}
  map construction.
\newblock In {\em Proc. of the International Conf. on Learning Representations
  (ICLR)}, 2023.

\bibitem{lin2017focal}
Tsung{-}Yi Lin, Priya Goyal, Ross~B. Girshick, Kaiming He, and Piotr
  Doll{\'{a}}r.
\newblock Focal loss for dense object detection.
\newblock In {\em Proc. of the IEEE International Conf. on Computer Vision
  (ICCV)}, 2017.

\bibitem{lin2022sparse4d}
Xuewu Lin, Tianwei Lin, Zixiang Pei, Lichao Huang, and Zhizhong Su.
\newblock Sparse4d: Multi-view 3d object detection with sparse spatial-temporal
  fusion.
\newblock {\em CoRR}, abs/2211.10581, 2022.

\bibitem{lin2023sparse4dv2}
Xuewu Lin, Tianwei Lin, Zixiang Pei, Lichao Huang, and Zhizhong Su.
\newblock Sparse4d v2: Recurrent temporal fusion with sparse model.
\newblock {\em CoRR}, abs/2305.14018, 2023.

\bibitem{liu2022learning}
Ruijin Liu, Dapeng Chen, Tie Liu, Zhiliang Xiong, and Zejian Yuan.
\newblock Learning to predict 3d lane shape and camera pose from a single image
  via geometry constraints.
\newblock In {\em Proc. of the Conf. on Artificial Intelligence (AAAI)}, 2022.

\bibitem{lstr}
Ruijin Liu, Zejian Yuan, Tie Liu, and Zhiliang Xiong.
\newblock End-to-end lane shape prediction with transformers.
\newblock In {\em Proc. of the IEEE Winter Conference on Applications of
  Computer Vision (WACV)}, 2021.

\bibitem{liu2023petrv2}
Yingfei Liu, Junjie Yan, Fan Jia, Shuailin Li, Aqi Gao, Tiancai Wang, and
  Xiangyu Zhang.
\newblock Petrv2: A unified framework for 3d perception from multi-camera
  images.
\newblock In {\em Proc. of the IEEE International Conf. on Computer Vision
  (ICCV)}, 2023.

\bibitem{luo2023latr}
Yueru Luo, Chaoda Zheng, Xu Yan, Tang Kun, Chao Zheng, Shuguang Cui, and Zhen
  Li.
\newblock Latr: 3d lane detection from monocular images with transformer.
\newblock In {\em Proc. of the IEEE International Conf. on Computer Vision
  (ICCV)}, 2023.

\bibitem{luo2022detr4d}
Zhipeng Luo, Changqing Zhou, Gongjie Zhang, and Shijian Lu.
\newblock Detr4d: Direct multi-view 3d object detection with sparse attention.
\newblock {\em arXiv preprint arXiv:2212.07849}, 2022.

\bibitem{mao2021once}
Jiageng Mao, Minzhe Niu, Chenhan Jiang, Hanxue Liang, Jingheng Chen, Xiaodan
  Liang, Yamin Li, Chaoqiang Ye, Wei Zhang, Zhenguo Li, Jie Yu, Chunjing Xu,
  and Hang Xu.
\newblock One million scenes for autonomous driving: {ONCE} dataset.
\newblock In {\em NeurIPS Datasets and Benchmarks}, 2021.

\bibitem{misra2021detr3d}
Ishan Misra, Rohit Girdhar, and Armand Joulin.
\newblock An end-to-end transformer model for 3d object detection.
\newblock In {\em Proc. of the IEEE International Conf. on Computer Vision
  (ICCV)}, 2021.

\bibitem{lanenet}
Davy Neven, Bert~De Brabandere, Stamatios Georgoulis, Marc Proesmans, and
  Luc~Van Gool.
\newblock Towards end-to-end lane detection: an instance segmentation approach.
\newblock In {\em Proc. IEEE Intelligent Vehicles Symposium (IV)}, 2018.

\bibitem{scnn}
Xingang Pan, Jianping Shi, Ping Luo, Xiaogang Wang, and Xiaoou Tang.
\newblock Spatial as deep: Spatial {CNN} for traffic scene understanding.
\newblock In {\em Proc. of the Conf. on Artificial Intelligence (AAAI)}, 2018.

\bibitem{philion2020lift}
Jonah Philion and Sanja Fidler.
\newblock Lift, splat, shoot: Encoding images from arbitrary camera rigs by
  implicitly unprojecting to 3d.
\newblock In {\em Proc. of the European Conf. on Computer Vision (ECCV)}, 2020.

\bibitem{pittner20233d}
Maximilian Pittner, Alexandru Condurache, and Joel Janai.
\newblock 3d-splinenet: 3d traffic line detection using parametric spline
  representations.
\newblock In {\em Proc. of the IEEE Winter Conference on Applications of
  Computer Vision (WACV)}, 2023.

\bibitem{pittner2024lanecpp}
Maximilian Pittner, Joel Janai, and Alexandru~P Condurache.
\newblock Lanecpp: Continuous 3d lane detection using physical priors.
\newblock In {\em Proc. IEEE Conf. on Computer Vision and Pattern Recognition
  (CVPR)}, 2024.

\bibitem{pizzati2019lane}
Fabio Pizzati, Marco Allodi, Alejandro Barrera, and Fernando Garc{\'{\i}}a.
\newblock Lane detection and classification using cascaded cnns.
\newblock In {\em Proc. of the International Conf. on Computer Aided Systems
  Theory (EUROCAST)}, 2019.

\bibitem{qu2021fololane}
Zhan Qu, Huan Jin, Yang Zhou, Zhen Yang, and Wei Zhang.
\newblock Focus on local: Detecting lane marker from bottom up via key point.
\newblock In {\em Proc. IEEE Conf. on Computer Vision and Pattern Recognition
  (CVPR)}, 2021.

\bibitem{5152255}
Davide Scaramuzza, Friedrich Fraundorfer, and Roland Siegwart.
\newblock Real-time monocular visual odometry for on-road vehicles with 1-point
  ransac.
\newblock In {\em Proc. IEEE International Conf. on Robotics and Automation
  (ICRA)}, 2009.

\bibitem{steinbrucker2011real}
Frank Steinbr{\"u}cker, J{\"u}rgen Sturm, and Daniel Cremers.
\newblock Real-time visual odometry from dense rgb-d images.
\newblock In {\em Proc. of the IEEE International Conf. on Computer Vision
  (ICCV) Workshops}, 2011.

\bibitem{sgld}
Jinming Su, Chao Chen, Ke Zhang, Junfeng Luo, Xiaoming Wei, and Xiaolin Wei.
\newblock Structure guided lane detection.
\newblock In {\em Proc. of the International Joint Conf. on Artificial
  Intelligence (IJCAI)}, 2021.

\bibitem{sun2020scalability}
Pei Sun, Henrik Kretzschmar, Xerxes Dotiwalla, Aurelien Chouard, Vijaysai
  Patnaik, Paul Tsui, James Guo, Yin Zhou, Yuning Chai, Benjamin Caine, et~al.
\newblock Scalability in perception for autonomous driving: Waymo open dataset.
\newblock In {\em Proc. IEEE Conf. on Computer Vision and Pattern Recognition
  (CVPR)}, 2020.

\bibitem{laneatt}
Lucas Tabelini, Rodrigo Berriel, Thiago~M Paixao, Claudine Badue, Alberto~F
  De~Souza, and Thiago Oliveira-Santos.
\newblock Keep your eyes on the lane: Real-time attention-guided lane
  detection.
\newblock In {\em Proc. IEEE Conf. on Computer Vision and Pattern Recognition
  (CVPR)}, 2021.

\bibitem{polylanenet}
Lucas~Tabelini Torres, Rodrigo~Ferreira Berriel, Thiago~M. Paix{\~{a}}o,
  Claudine Badue, Alberto F.~De Souza, and Thiago Oliveira{-}Santos.
\newblock Polylanenet: Lane estimation via deep polynomial regression.
\newblock In {\em Proc. of the International Conf. on Pattern Recognition
  (ICPR)}, 2020.

\bibitem{wang2022ganet}
Jinsheng Wang, Yinchao Ma, Shaofei Huang, Tianrui Hui, Fei Wang, Chen Qian, and
  Tianzhu Zhang.
\newblock A keypoint-based global association network for lane detection.
\newblock In {\em Proc. IEEE Conf. on Computer Vision and Pattern Recognition
  (CVPR)}, 2022.

\bibitem{wang2023bev}
Ruihao Wang, Jian Qin, Kaiying Li, Yaochen Li, Dong Cao, and Jintao Xu.
\newblock Bev-lanedet: An efficient 3d lane detection based on virtual camera
  via key-points.
\newblock In {\em Proc. IEEE Conf. on Computer Vision and Pattern Recognition
  (CVPR)}, 2023.

\bibitem{wang2023exploring}
Shihao Wang, Yingfei Liu, Tiancai Wang, Ying Li, and Xiangyu Zhang.
\newblock Exploring object-centric temporal modeling for efficient multi-view
  3d object detection.
\newblock In {\em Proc. of the IEEE International Conf. on Computer Vision
  (ICCV)}, 2023.

\bibitem{wang2022spatio}
Yin Wang, Qiuyi Guo, Peiwen Lin, Guangliang Cheng, and Jian Wu.
\newblock Spatio-temporal fusion-based monocular 3d lane detection.
\newblock In {\em Proc. of the British Machine Vision Conf. (BMVC)}, 2022.

\bibitem{once3dlanes}
Fan Yan, Ming Nie, Xinyue Cai, Jianhua Han, Hang Xu, Zhen Yang, Chaoqiang Ye,
  Yanwei Fu, Michael~Bi Mi, and Li Zhang.
\newblock Once-3dlanes: Building monocular 3d lane detection.
\newblock In {\em Proc. IEEE Conf. on Computer Vision and Pattern Recognition
  (CVPR)}, 2022.

\bibitem{zheng2024pvalane}
Zewen Zheng, Xuemin Zhang, Yongqiang Mou, Xiang Gao, Chengxin Li, Guoheng
  Huang, Chi-Man Pun, and Xiaochen Yuan.
\newblock Pvalane: Prior-guided 3d lane detection with view-agnostic feature
  alignment.
\newblock In {\em Proc. of the Conf. on Artificial Intelligence (AAAI)}, 2024.

\bibitem{zhu2021defdetr}
Xizhou Zhu, Weijie Su, Lewei Lu, Bin Li, Xiaogang Wang, and Jifeng Dai.
\newblock Deformable {DETR:} deformable transformers for end-to-end object
  detection.
\newblock In {\em Proc. of the International Conf. on Learning Representations
  (ICLR)}, 2021.

\bibitem{zou2019robust}
Qin Zou, Hanwen Jiang, Qiyu Dai, Yuanhao Yue, Long Chen, and Qian Wang.
\newblock Robust lane detection from continuous driving scenes using deep
  neural networks.
\newblock {\em IEEE Trans. on Vehicular Technology (VTC)}, 2020.

\end{thebibliography}
}

\appendix
\setcounter{page}{1}
\maketitlesupplementary
\section{More details on the 3D lane representation}
\label{sec:supp_rep}
In this section, we provide additional information about CR splines and the curve computation.
\subsection{CR splines}\label{subsec:crsplines}
\begin{figure}[tb]
	\centering
	\begin{subfigure}[b]{0.49\linewidth}
		\centering
		\includegraphics[width=0.99\linewidth]{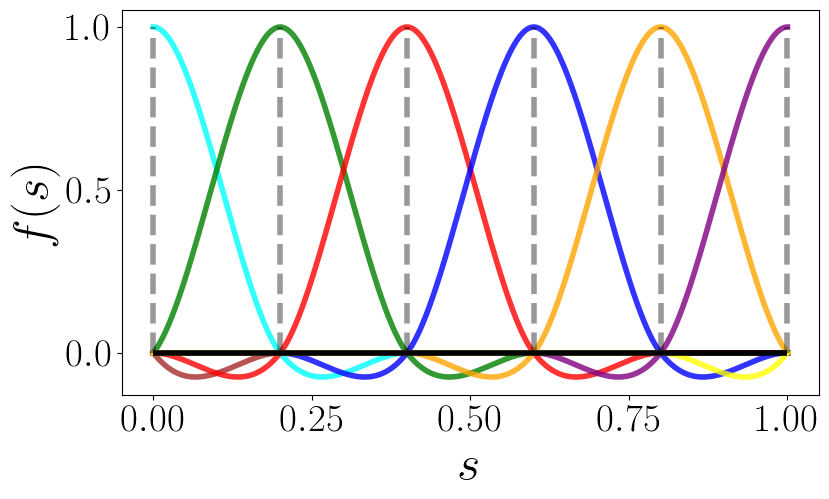}
		\caption{CR basis polynomials\label{fig:crsplines1}}
	\end{subfigure}
	\begin{subfigure}[b]{0.49\linewidth}	
		\centering
		\includegraphics[width=0.99\linewidth]{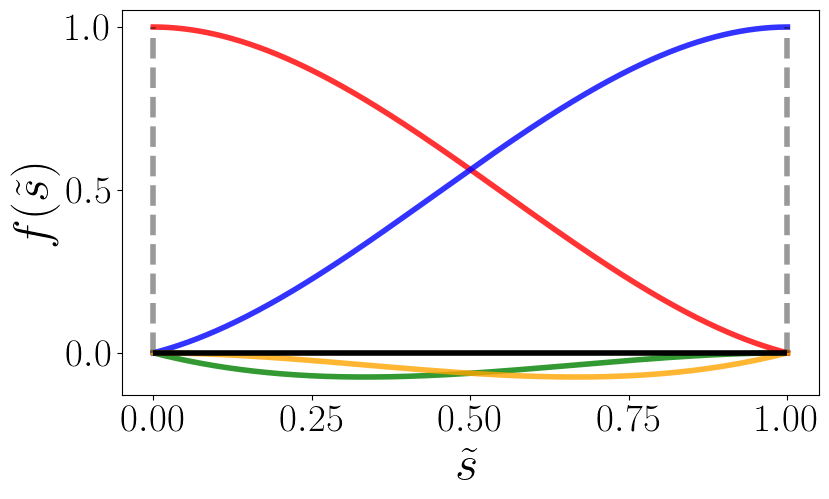}
		\caption{Local segment 
		\label{fig:crsplines2}}
	\end{subfigure}
	\caption{Global and local CR spline basis functions.}	\label{fig:crsplines}
\end{figure}

We provide details on our selected curve representation, Catmull-Rom (CR) splines \cite{CRsplines}.
CR splines are a class of splines comprised of piece-wise defined smooth third-order polynomial splines. A special property that we exploit for our transformer-tailored representation is that the curve inherently passes through its control points. In \figref{fig:crsplines1} we illustrate the basis functions for an example CR spline using 6 control points. Each value of a CR spline function is affected by the 4 nearest control points. Thus, a local segment between two consecutive control points $p_{j}$ and $p_{j+1}$ of a 1D spline $f(\tilde{s})$ with curve argument $\tilde{s} \in [0, \, 1]$  is defined as
\begin{align}
	f(\tilde{s}) = \begin{bmatrix} \tilde{s}^3 & \tilde{s}^2 & \tilde{s} & 1 \end{bmatrix} \cdot \tilde{\mathbf{M}}_{\textrm{CR}} \cdot \begin{bmatrix} p_{j-1} \\ p_{j} \\ p_{j+1} \\ p_{j+2} \end{bmatrix} 
\end{align}
and illustrated in \figref{fig:crsplines2}. 
$\tilde{\mathbf{M}}_{\textrm{CR}} \in \mathbb{R}^{4 \times 4}$ is the coefficient matrix for this segment defined as 
\begin{align}
	\tilde{\mathbf{M}}_{\textrm{CR}} = \frac{1}{2} \, \cdot \, \begin{bmatrix}
	-1 		&  3 	& -3 	&  1 	\\
	 2 		& -5 	&  4 	& -1 	\\
	-1		&  0	&  1	&  0	\\
	 0		&  2	&  0 	&  0
\end{bmatrix} 
\end{align}
and $\tilde{\mathbf{s}} = \begin{bmatrix} \tilde{s}^3 & \tilde{s}^2 & \tilde{s} & 1 \end{bmatrix}$ the spline argument vector. 
The product of $\tilde{\mathbf{s}}$ and $\tilde{\mathbf{M}}_{\textrm{CR}}$ represents the four involved CR basis polynomials. The multiplication with the control point vector yields a weighted sum, which defines the shape of the curve segment. 

To extend this formulation from a local segment to a global curve with an arbitrary number of control points $M$, we have to apply two changes. First, the spline argument vector needs to be normalized based on the knot positions $s_k$, which are visualized in \figref{fig:crsplines1} by the dashed vertical lines, such that we obtain 
\begin{align}
\mathbf{s} = \begin{bmatrix} \big({({s - s_k}) / (s_{k+1} - s_k)}\big)^3 \\ \big({({s - s_k}) / (s_{k+1} - s_k)}\big)^2 \\ ({s - s_k}) / (s_{k+1} - s_k)  \\ 1 \end{bmatrix}^T
\end{align}
for $s \in [s_k, \, s_{k+1}]$. Second, the coefficient matrix needs to be augmented to ${\mathbf{M}}_{\textrm{CR}} \in \mathbb{R}^{4 \times M}$, where most values remain zero and only the entries affecting the four involved control points equal $\tilde{\mathbf{M}}_{\textrm{CR}}$. Thus, ${\mathbf{M}}_{\textrm{CR}}(s)$ now varies with the curve parameter $s$ and is determined by the interval between the corresponding knots $s_k$ and $s_{k+1}$.
Consequently, we obtain 
\begin{align}
	f({s}) &= \mathbf{s} \cdot {\mathbf{M}}_{\textrm{CR}}(s) \cdot \mathbf{p} \quad \textrm{with} \\
	{\mathbf{M}}_{\textrm{CR}}(s) &= \begin{cases}
    \tilde{\mathbf{M}}_{\textrm{CR}} & \text{if } s \in [s_k, s_{k+1}]\\
        0                                  & \text{else} \, ,
    \end{cases}                      
\end{align}
with control point vector $\mathbf{p} \in \mathbb{R}^{M}$. Note, that although $\mathbf{M}_{\textrm{CR}}(s)$ depends on $s$, which we omitted in the main paper for simplicity, the matrix $\mathbf{s} \cdot {\mathbf{M}}_{\textrm{CR}}(s)$ can be pre-computed before training or inference. Thus, in each training iteration or inference step, the post-processing required to generate full splines reduces to a single matrix multiplication. 

Finally, extending the 1D function to a full curve model $\mathbf{f}(s) \in \mathbb{R}^{4} $ with 3 spatial ($x_{j}$, $y_{j}$, $z_{i}$) and 1 visibility ($v_{j}$) results in
\begin{align}
	\mathbf{f}(s) &= \begin{bmatrix} s^3 & s^2 & s & 1 \end{bmatrix} \cdot \mathbf{M}_{\textrm{CR}}(s) \cdot \mathbf{P} \quad \textrm{with} \\
 \mathbf{P} &= 
\begin{bmatrix} 
x_{1} & y_{1} & z_{1} & v_{1} \\
 \multicolumn{4}{c}{\dots}  \\
x_{j} & y_{j} & z_{j} & v_{j} \\
 \multicolumn{4}{c}{\dots}  \\
x_{M} & y_{M} & z_{M} &  v_{M} \\
\end{bmatrix} \, , 
\end{align}
with control point matrix $\mathbf{P} \in \mathbb{R}^{M \times 4}$ and $\mathbf{M}_{\textrm{CR}}(s)$ the full coefficient matrix. 

\subsection{Curve computation and parameterization}\label{subsec:parametrization}
In practice, the curve is computed by sampling arguments $s$ in the range $[0, \, 1]$, pre-computing the matrix $\mathbf{s} \cdot {\mathbf{M}}_{\textrm{CR}}(s)$ and generating the curve via multiplication with the control point matrix (as described in \secref{subsec:crsplines}). 

During training, we need to determine the curve argument  $\hat{s}$ for a given ground truth sample $(\hat{x}, \hat{y}, \hat{z}, \hat{v})$ in order to compute the loss at the corresponding position in the predicted curve. Similar to \cite{pittner20233d}, we keep the longitudinal $y$-component of the spline control points fixed and distribute them uniformly along the range $[y_s, y_e]$. This simplifies the determination of curve arguments $\hat{s}$ to
\begin{align}
\hat{s} = (\hat{y} - y_s) / (y_e - y_s) \, .
\end{align}
\section{Detection losses}
\label{sec:losses}
In this section, more details about our utilized losses are provided.
\subsection{Classification loss}
We employ focal loss \cite{lin2017focal} for category classification, formulated as follows
\begin{align}
\mathcal{L}_{cls} =& - \frac{1}{N} \sum_{i=1}^{N} \sum_{k=1}^{K+1} \Big( \hat{\mathbf{C}}_{ik} \cdot  \big(1 - \mathbf{C}_{ik} \big)^{\gamma} \cdot \log\big({\mathbf{C}}_{ik}\big) \Big) \, ,
\end{align}
with predicted category probability distribution of the $i^{\textrm{th}}$ line proposal $\mathbf{C}_{i}$, ground truth one-hot vector $\hat{\mathbf{C}}_{i}$ and focusing parameter $\gamma \geq 0$.
Note, that the probabilities for predicted categories and the background class sum up to $\sum_{k=1}^{K+1} {\mathbf{C}}_{ik} = 1$.

\subsection{Regression loss}
Inspired by \cite{pittner2024lanecpp}, we apply L1 loss along the visible points  
\begin{align}
\mathcal{L}_{reg} = \frac{1}{N} \sum_i^N \sum_{j}^{M_\textrm{GT}} \hat{v}_{ij} \cdot 
\Big|\Big| 
\begin{pmatrix}
\mathbf{f}_{x,i}(\hat{s}_j) \\
\mathbf{f}_{z,i}(\hat{s}_j)
\end{pmatrix}
-  
\begin{pmatrix}
\hat{x}_{ij}   \\
\hat{z}_{ij}	 
\end{pmatrix}
\Big|\Big|_1 \, , 
\label{eq:regloss}
\end{align}
with ground truth points ($\hat{x}_{ij}$, $\hat{y}_{ij}$, $\hat{z}_{ij}$, $\hat{v}_{ij}$) of the $i^\textrm{th}$ line proposal and $M_\textrm{GT}$ the number of samples in the ground truth. 
The corresponding curve argument $\hat{s}_j$ is obtained as described in \secref{subsec:parametrization}.

\subsection{Visibility loss}
For the visibility, we use binary cross-entropy
\begin{align}
	\mathcal{L}_{vis} =& \frac{1}{N} \sum_i^N \sum_{j}^{M_\textrm{GT}}  
\hat{v}_{ij} \cdot \log\big( \mathbf{f}_{v, ij}(s_j) \big) + \\ 
& (1 - \hat{v}_{ij}) \cdot \log\big(1 - \mathbf{f}_{v, ij}(s_j) \big) \, .
\end{align}

\section{Spatio-temporal attention}
As discussed in the main paper, the three components of the spatio-temporal attention mechanism interact with distinct subsets of queries, which serve as keys and values. This is implemented through masked attention, where a dedicated mask is computed for each component $-$ $\mathrm{SLA}$, $\mathrm{PNA}$ and $\mathrm{TCA}$ $-$ based on either the structure of the query matrix or the positions of the associated control points.

\subsection{Masking}
In $\mathrm{SLA}$, for a given input query $\mathbf{\tilde{Q}}_{ij}$, the corresponding set of keys and values, $\mathbf{\tilde{Q}}_\mathrm{SLA}$, consists of queries from the same line proposal $i$. Thus, each query in $\mathrm{SLA}$ interacts with the $M$ queries on the same line.

In $\mathrm{PNA}$, points on neighboring lines in the orthogonal direction to the curve are selected. To achieve this, we employ the method described in \cite{pittner2024lanecpp} to identify the two nearest neighbors on each adjacent line in the orthogonal direction. Thus, each query in $\mathrm{PNA}$ interacts with the $2 \cdot (N-1)$ parallel neighbors.

In $\mathrm{TCA}$, queries in the memory associated with the $M_\textrm{TCA}$ nearest temporally propagated control points are selected. For the distance metric we use euclidean distance. Thus, each query in $\mathrm{TCA}$ interacts with the $M_\textrm{TCA}$ nearest queries in the memory queue. Note, that we used $M_\textrm{TCA}=10$ in our experiments.

\subsection{Potential for efficiency benefits}
We also note that our approach has potential for further efficiency improvements, which could be explored as future work:
In its core, the proposed spatio-temporal attention module masks out redundant relations of tokens. The resulting attention matrix has only $\sim10\,\%$ of active token relations and thus $\sim90\,\%$ sparsity. While our implementation is currently based on a simplistic standard attention-layer that uses masking without leveraging sparsity, optimized approaches like FlashAttention \cite{dao2022flashattention,dao2023flashattention} could be used in future in combination with deployment on more modern GPUs. Such an implementation could exploit the large extent of sparsity in the attention-layers and potentially lead to severe savings in runtime and memory.
\begin{figure}[tb]
  \centering	
  	\begin{subfigure}[b]{0.9\linewidth}
		\centering
		\includegraphics[width=0.75\linewidth]{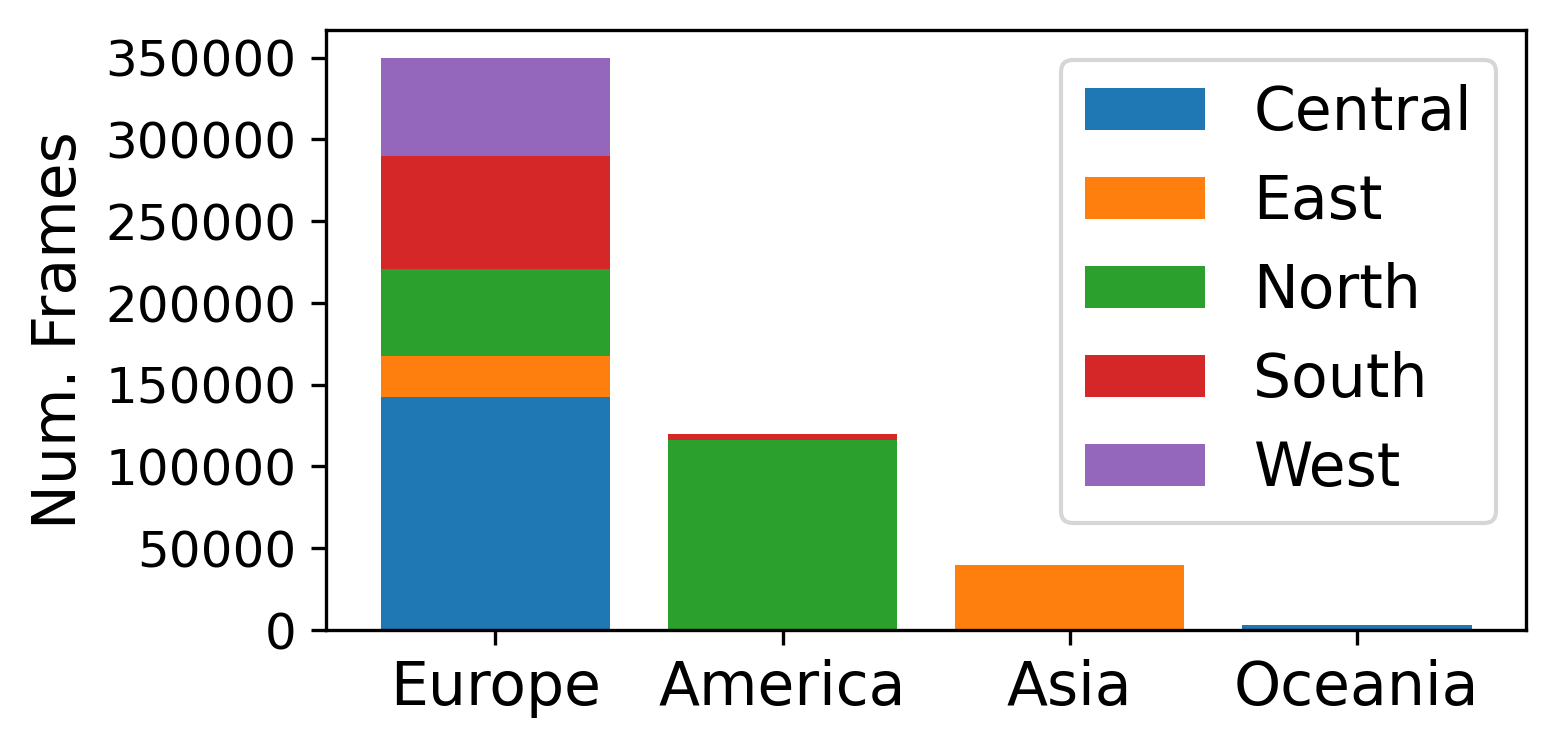}
		\centering
		\caption{Regions and countries \label{fig:stats-a}}
	\end{subfigure}
	\begin{subfigure}[b]{0.9\linewidth}
		\centering
		\includegraphics[width=0.75\linewidth]{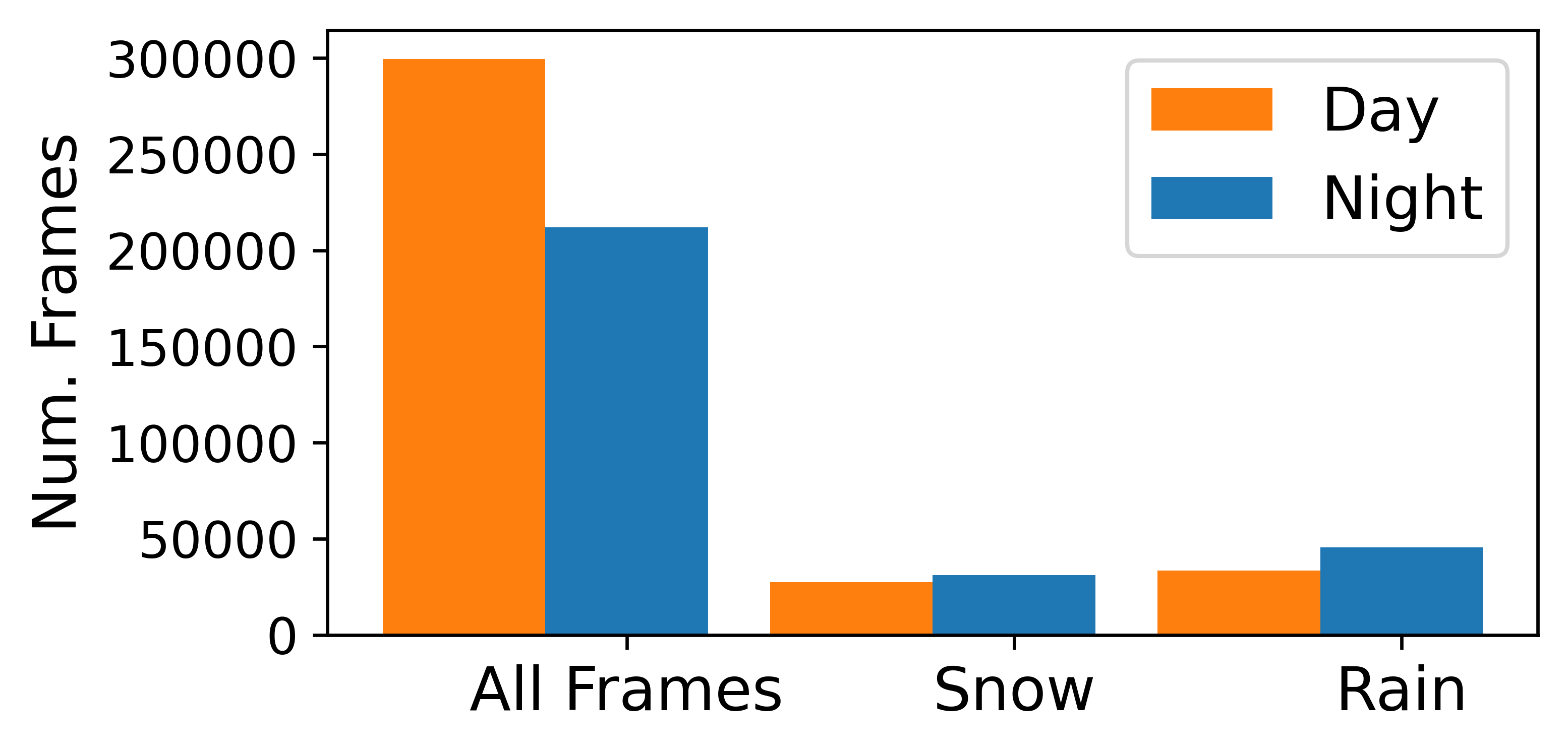}
		\centering
		\caption{Weather and daytime\label{fig:stats-b}}
	\end{subfigure}
	\begin{subfigure}[b]{0.9\linewidth}
		\centering
		\includegraphics[width=0.45\linewidth]{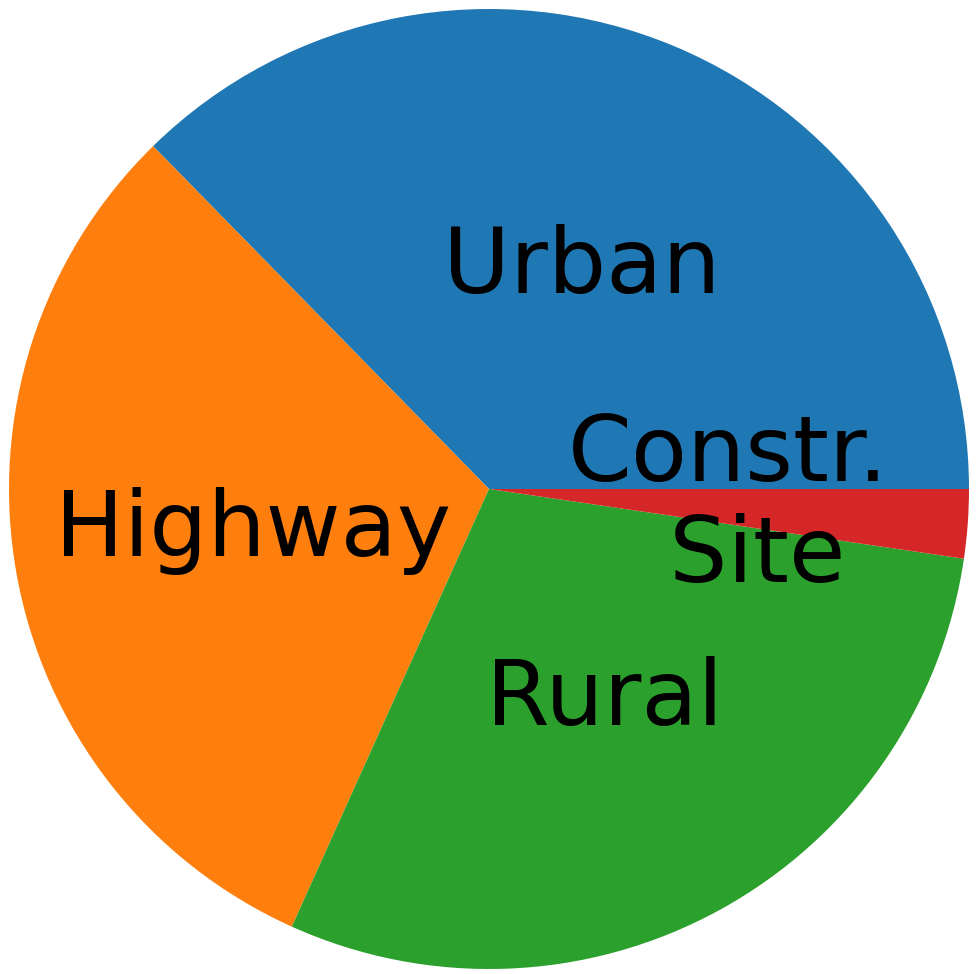}
		\centering
		\caption{Environments \label{fig:stats-c}}
	\end{subfigure}   
	\begin{subfigure}[b]{0.9\linewidth}
		\centering
		\includegraphics[width=0.75\linewidth]{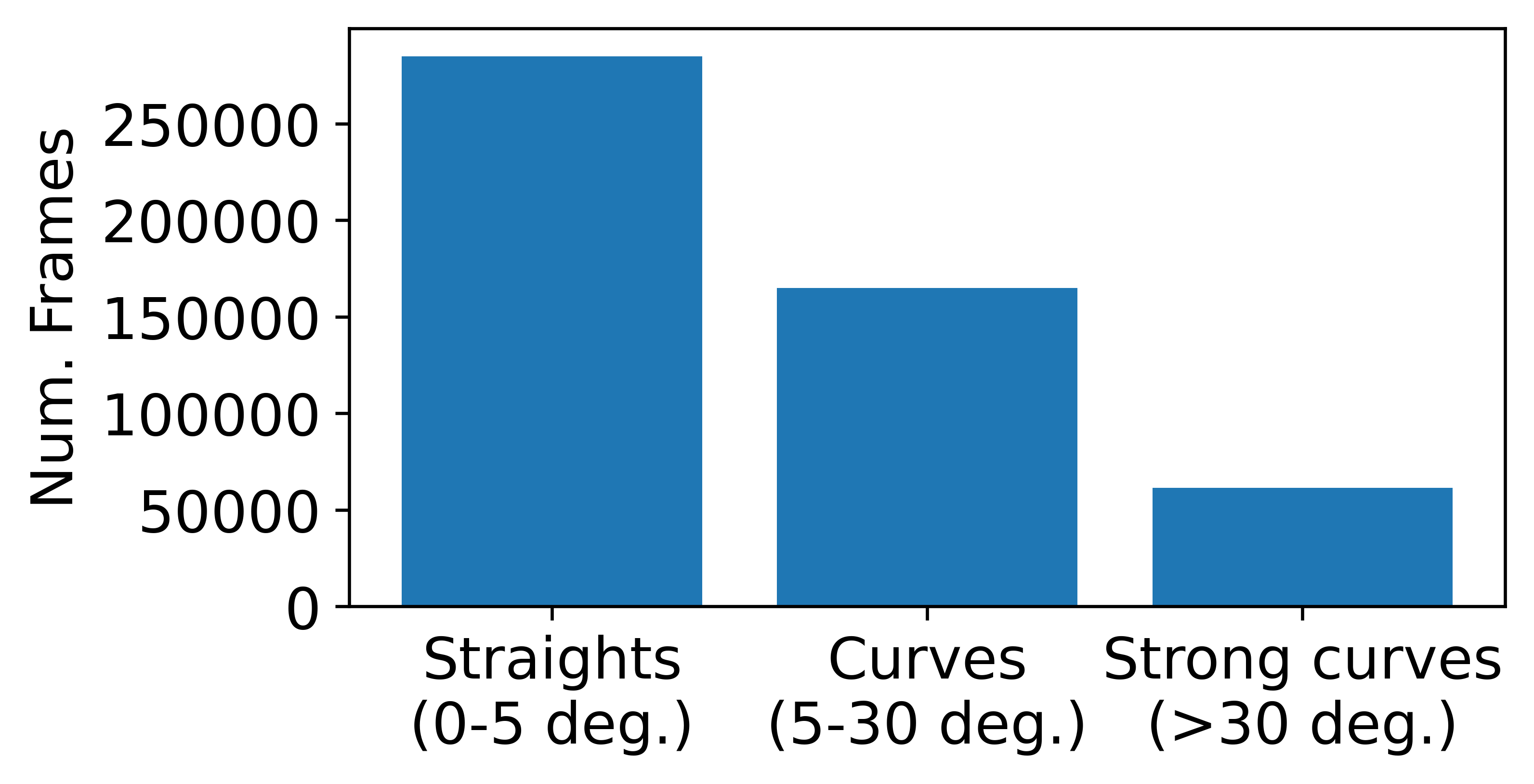}
		\centering
		\caption{Lane curvature (according to accumulated degree) \label{fig:stats-d}}
	\end{subfigure}   
\caption{Scenario-based dataset statistics.}
  \label{fig:datasetstats}
\end{figure}
\begin{figure}[tb]
  \centering	
  	\begin{subfigure}[b]{.9\linewidth}
		\centering
		\includegraphics[width=0.8\linewidth]{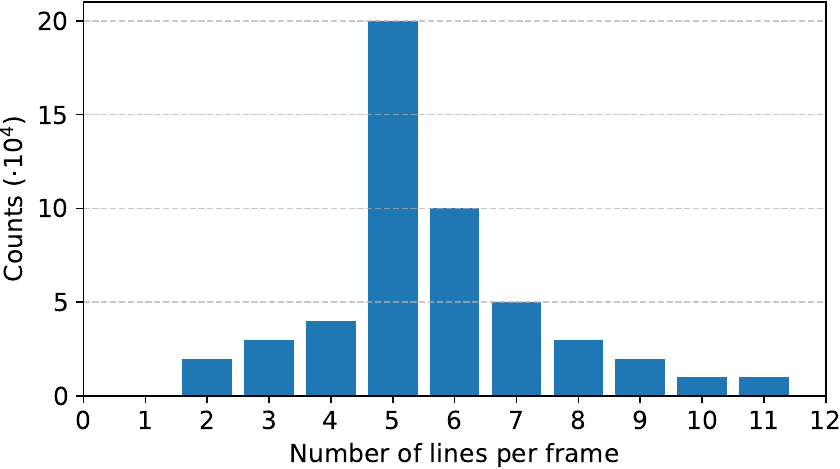}
		\centering
		\caption{Distribution of number of lines per frame \label{fig:stats-linenumbers}}
	\end{subfigure}
	\begin{subfigure}[b]{.9\linewidth}
		\centering
		\includegraphics[width=0.7\linewidth]{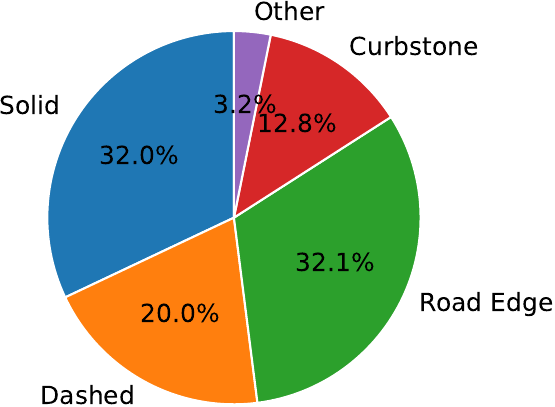}
		\centering
		\caption{Line category distribution\label{fig:stats-categories}}
	\end{subfigure}
\caption{Lane marking specific dataset statistics.}
  \label{fig:linestats}
\end{figure}

\section{Dataset statistics}
\label{sec:suppdataset}
\figref{fig:datasetstats} shows several statistics of our dataset with respect to the driving scenario and environment. 
The sequences were recorded in various regions and countries across the globe with the distribution given in \figref{fig:stats-a}. 
We made sure to collect the data under various weather conditions and daytimes (\figref{fig:stats-b}). Moreover, the sequences were chosen such that the proportions of highway, urban and rural situations are approximately equal and made sure that even corner case situations like construction zones are included (\figref{fig:stats-c}). Finally, \figref{fig:stats-d} demonstrates that the dataset also covers lanes from high curvature roads to enable the trained model to learn from challenging driving scenarios.

In \figref{fig:linestats} lane marking specific statistics are illustrated. From \figref{fig:stats-linenumbers} is is obvious that the number of occurring lane markings per frame varies across our dataset. The highest count of frames is shown for a number of 5 markings per frame and the maximum number of occurring lines is 11. In \figref{fig:stats-categories} we show the distribution of line categories. Besides the frequently occurring dashed and solid markings, the labels also contain a large number of road edges and even curbstones. ``Other'' contains special kinds of line-like objects like botts' dots and road furniture.

\section{Auto-labeling details}
In the main paper we already gave an overview of the components and functionality of our auto-labeling pipeline. More details regarding the most important components are provided in the following.

\textbf{2D lane detector.} 
A profound 2D lane detector based on LaneATT \cite{laneatt} is utilized to generate 2D pseudo labels, which are elevated to 3D space in a later stage. Certain modifications were applied to the model, which lead to slight improvements and more robust detection behavior particularly in the near-range detection. The network was further trained using an additional set of 2D labeled images obtained from recordings captured with the same vehicles and cameras. Subsequently, the model is deployed to perform the 2D pseudo-labeling task on the unlabeled set of sequences in our dataset.
For a more detailed assessment of the performance of the 2D lane detector, we refer to the qualitative and quantitative results of the experiment section from \cite{laneatt}.

\textbf{3D surface model.} 
We use a simple yet sufficient road surface model based on the precise ego-trajectory of the vehicle. The assumption of the road model is that the orientation of the vehicle, which is given in the ego-trajectory, spans a local plane-based surface segment around each of its locations. Each plane segment is defined between the current and next position in the ego-trajectory, resulting in a first degree spline surface in 3D space. Since the ego-trajectory, which defines this surface model, was precisely optimized using visual odometry approaches, it provides an accurate estimation of the real road surface in the vicinity of the vehicle. Since we propose to project only the first segments of 2D pseudo-labels to this surface, the resulting points in 3D space will usually not have a large distance to the vehicle. The 3D position of the projected lane points can therefore be considered sufficiently reliable.

\textbf{Projection to 3D.} Since the surface model consists merely of stacked planes, the intersection of a visual ray (originating in the camera center traveling through a 2D lane point) and the road surface can be analytically computed. Thus, the projection of 2D lane points onto the road surface is a simple and low-cost operation.

\textbf{Accumulation and Tracking.} Connecting line points in 3D to obtain global line instances requires tracking. For the tracker a solution based on the prominent Kalman Filter is utilized with certain modifications to get a tailored solution for the line tracking task (instead of standard object tracking). Applying the line tracker to the 3D line points, we receive full line instances with consistent track IDs.

\textbf{3D ground truth per frame.} After accumulation and tracking, the resulting global line instances, which are defined for an entire sequence, are simply transformed to the local vehicle coordinate systems for each frame in the sequence utilizing the ego-pose information for the respective frame. As a result we obtain 3D lines for each frame in its local 3D coordinate system.
\begin{table*}[tb]
		\centering
		\resizebox{\linewidth}{!}{
			\begin{tabular}{l||ccc|ccccc|ccc}
				\toprule
				\multirow{2}{*}{\bf{Method}}	& \multicolumn{3}{c|}{\bf{Environment}} & \multicolumn{5}{c|}{\bf{Daytime / Weather}} & \multicolumn{3}{c}{\bf{Curvature}} \\ 
				& \bf{Highway} & \bf{Urban} & \bf{Rural} & \bf{Sun} & \bf{Night} & \bf{Rain} & \bf{Snow} & \bf{Fog} & \bf{Straights} & \bf{Curves} & \bf{Strong Curves} \\
				\hhline{=||===|=====|===}
				PersFormer \cite{chen2022persformer} 							& $64.2~\%$ & $48.5~\%$ & $60.7~\%$ & $61.2~\%$ & $55.1~\%$ & $52.6~\%$ & $52.9~\%$ & $54.3~\%$ & $64.4~\%$ & $46.8~\%$ & $22.1~\%$	\\
				LaneCPP \cite{pittner2024lanecpp} 								& $67.9~\%$ & $51.4~\%$ & $65.1~\%$ & $64.0~\%$ & $58.9~\%$ & $55.4~\%$ & $56.3~\%$ & $58.1~\%$ & $70.2~\%$ & $50.8~\%$ & $26.3~\%$	\\
				Baseline (LATR) \cite{luo2023latr} 								& $\underline{70.4~\%}$ & $\underline{53.8~\%}$ & $\underline{66.3~\%}$ & $\underline{66.5~\%}$ & $\underline{60.2~\%}$ & $\underline{57.7~\%}$ & $\underline{58.7~\%}$ & $\underline{59.3~\%}$ & $\underline{73.0~\%}$ & $\underline{51.6~\%}$ & $\underline{29.8~\%}$	\\
				\rowcolor{Gray} \bf{SparseLaneSTP (ours)} 										& $\mathbf{74.6~\%}$ & $\mathbf{58.0~\%}$ & $\mathbf{72.2~\%}$ & $\mathbf{70.6~\%}$ & $\mathbf{64.7~\%}$ & $\mathbf{64.9~\%}$ & $\mathbf{61.5~\%}$ & $\mathbf{66.6~\%}$ & $\mathbf{75.8~\%}$ & $\mathbf{59.5~\%}$ & $\mathbf{35.3~\%}$			\\
			\bottomrule
		\end{tabular}}
		\caption{F1-Score comparison for different scenarios on our dataset. \textbf{Best performance} and \underline{second best} are highlighted.} 
		\label{tab:comparison-quant-scenarios}
\end{table*}

\section{Additional results}
\label{sec:suppresults}
We provide additional qualitative comparisons of our model SparseLaneSTP and LATR \cite{luo2023latr}.

\subsection{Qualitative results on OpenLane}
\figref{fig:supp-comparison-ol} shows additional qualitative results from our model compared to LATR. These examples emphasize the advantages of our proposed contributions discussed in the main paper while also exploring new scenarios in greater detail.

\figref{fig:supp-comparison-ol-1} demonstrates the advantage of our continuous representation in precise start-point estimation. Additionally, \figref{fig:supp-comparison-ol-2} highlights another key benefit. The network inherently produces smooth curves, whereas the discrete lane model struggles to accurately represent the sharp continuous curve on the right. 

\figref{fig:supp-comparison-ol-3} illustrates a case where our model accurately detects the two splitting lines, whereas LATR fails. This suggests that our novel STA component effectively focuses on neighboring points of both lines, whereas LATR relies on global attention, incorporating a majority of redundant relations. Additionally, the applied spatial regularization enforces smoothness, penalizing predictions similar to those produced by LATR. 

\figref{fig:supp-comparison-ol-4} and \figref{fig:supp-comparison-ol-5} illustrate the advantages of temporal modeling over the non-temporal LATR approach. In \figref{fig:supp-comparison-ol-4}, the crossing vehicle further obstructs the already faint lane markings. Nonetheless, our model maintains consistent detections. A similar behavior is observed in \figref{fig:supp-comparison-ol-5}, where partial occlusion does not impact detection performance. 

An interesting case is shown in \figref{fig:supp-comparison-ol-6}, depicting a scenario with severely limited visibility due to extreme weather conditions. With most of the image blurred, LATR fails and produces random predictions. In contrast, our model leverages queries stored in memory, enabling stable and sufficient detection.

\subsection{Qualitative results on our 3D lane dataset}
\begin{figure*}
	\centering
	\begin{subfigure}[b]{0.325\linewidth}
		\centering
		\includegraphics[width=1.\linewidth]{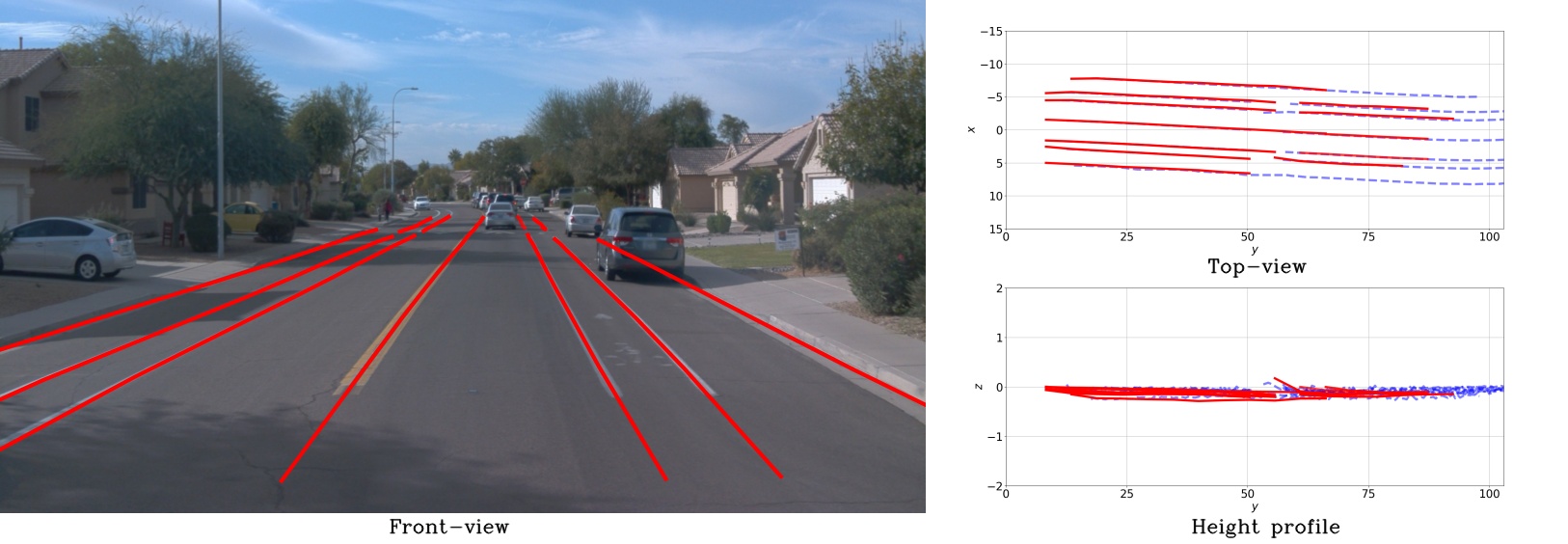}
		\includegraphics[width=1.\linewidth]{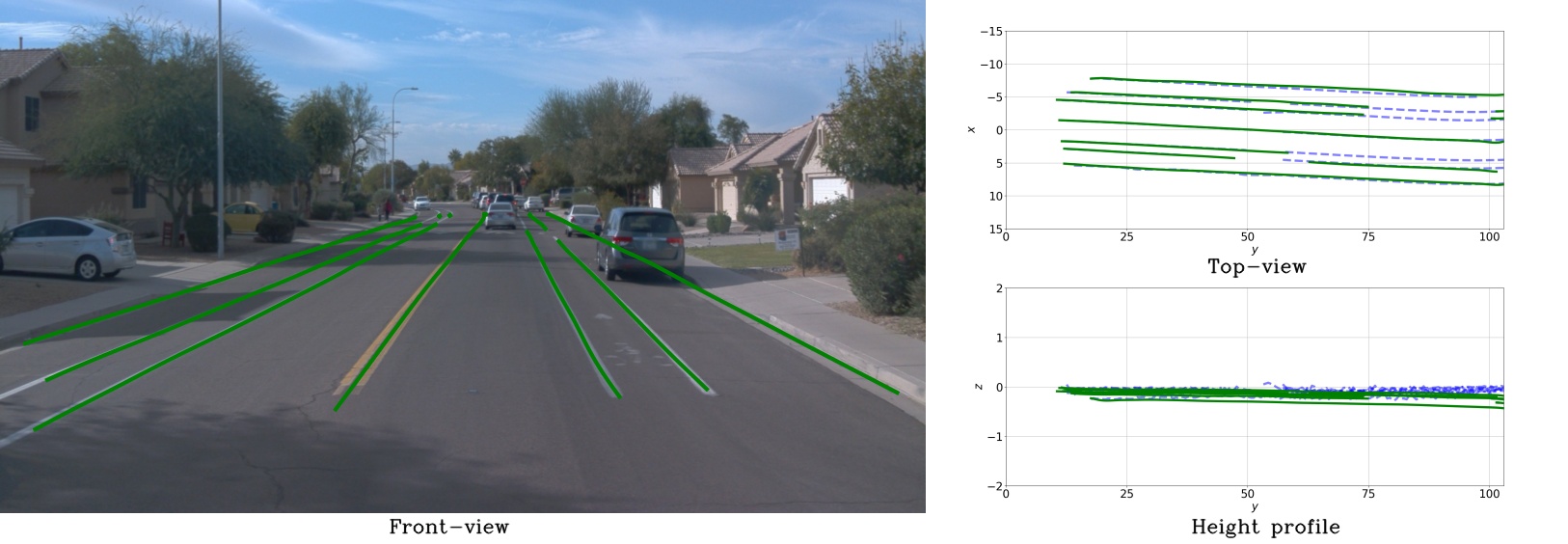}
		\caption{\label{fig:supp-comparison-ol-1}}
	\end{subfigure}
	\begin{subfigure}[b]{0.325\linewidth}
		\centering
		\includegraphics[width=1.\linewidth]{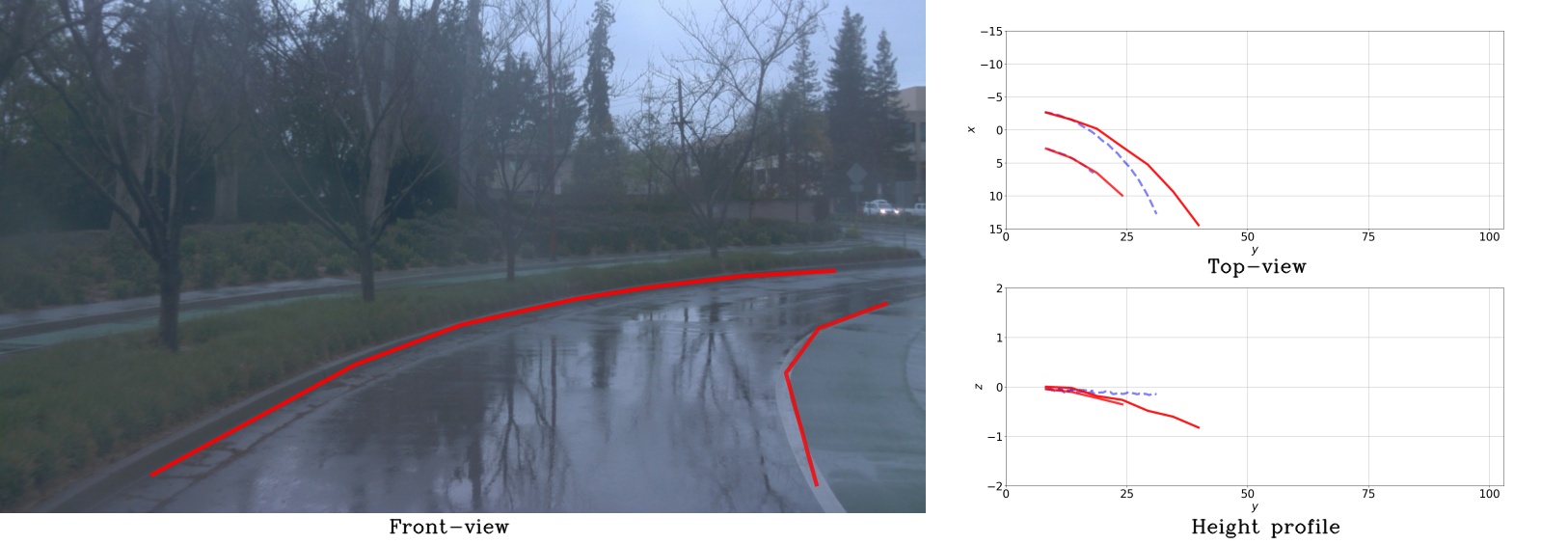}
		\includegraphics[width=1.\linewidth]{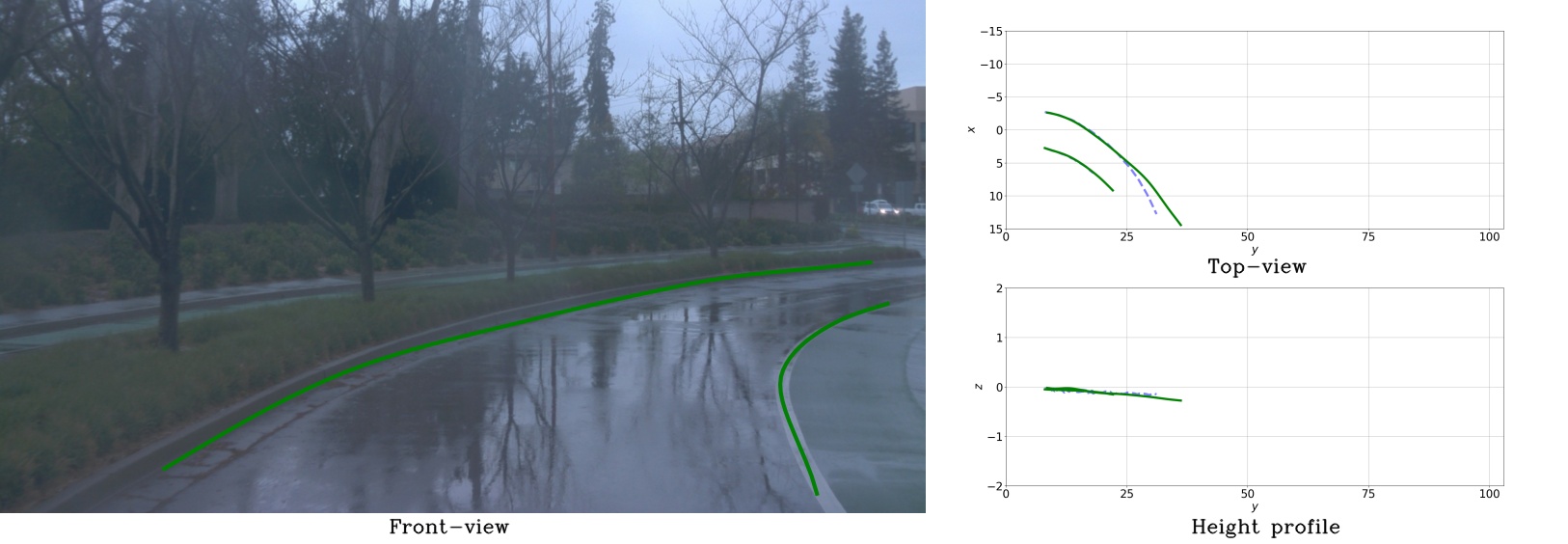}
		\caption{\label{fig:supp-comparison-ol-2}}
	\end{subfigure}
	\begin{subfigure}[b]{0.325\linewidth}
		\centering
		\includegraphics[width=1.\linewidth]{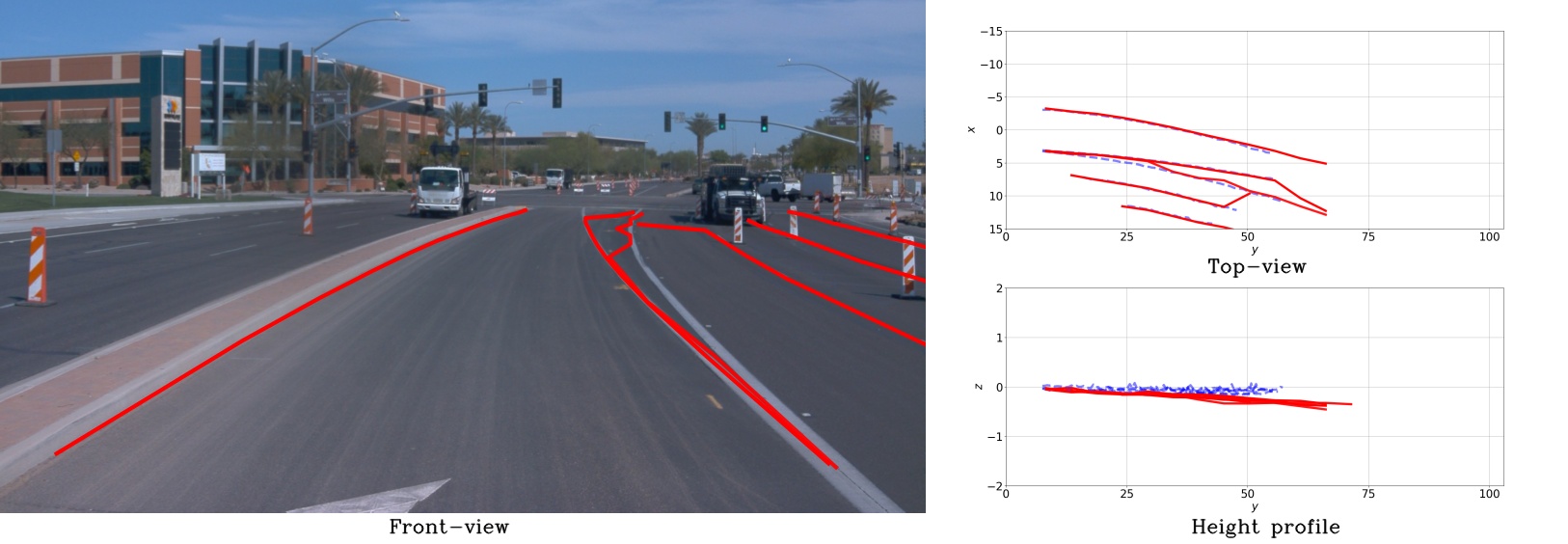}
		\includegraphics[width=1.\linewidth]{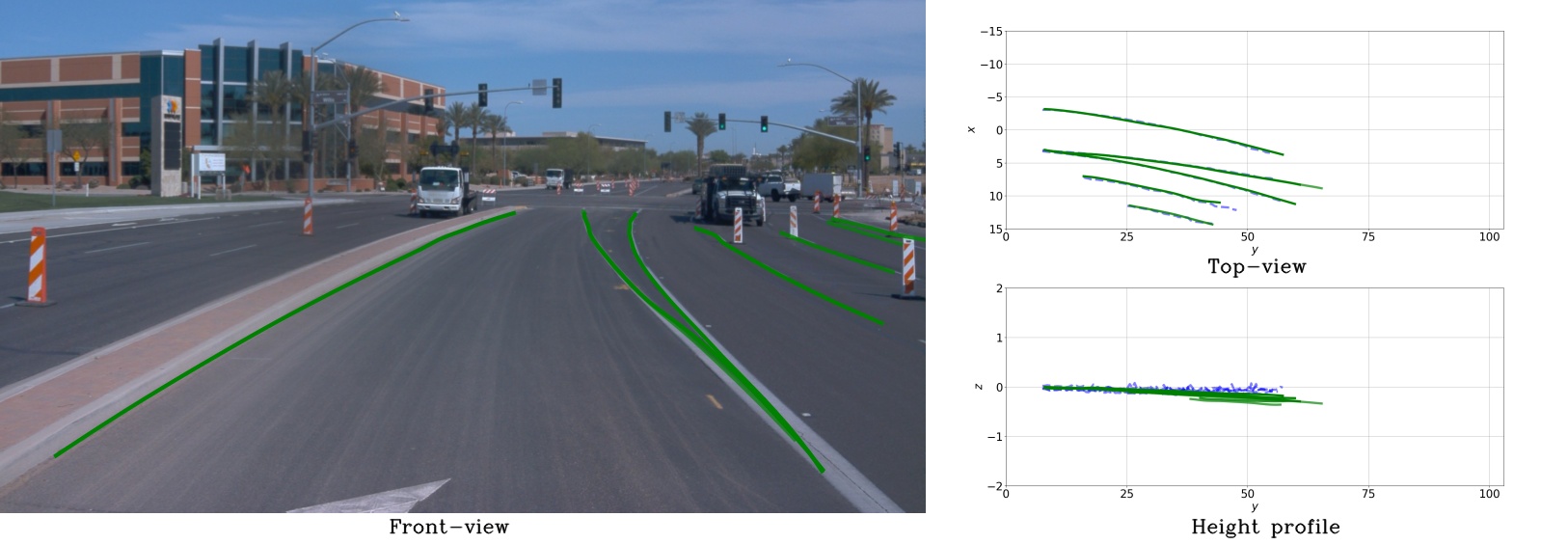}
		\caption{\label{fig:supp-comparison-ol-3}}
	\end{subfigure}

\begin{subfigure}[b]{0.325\linewidth}
		\centering
		\includegraphics[width=1.\linewidth]{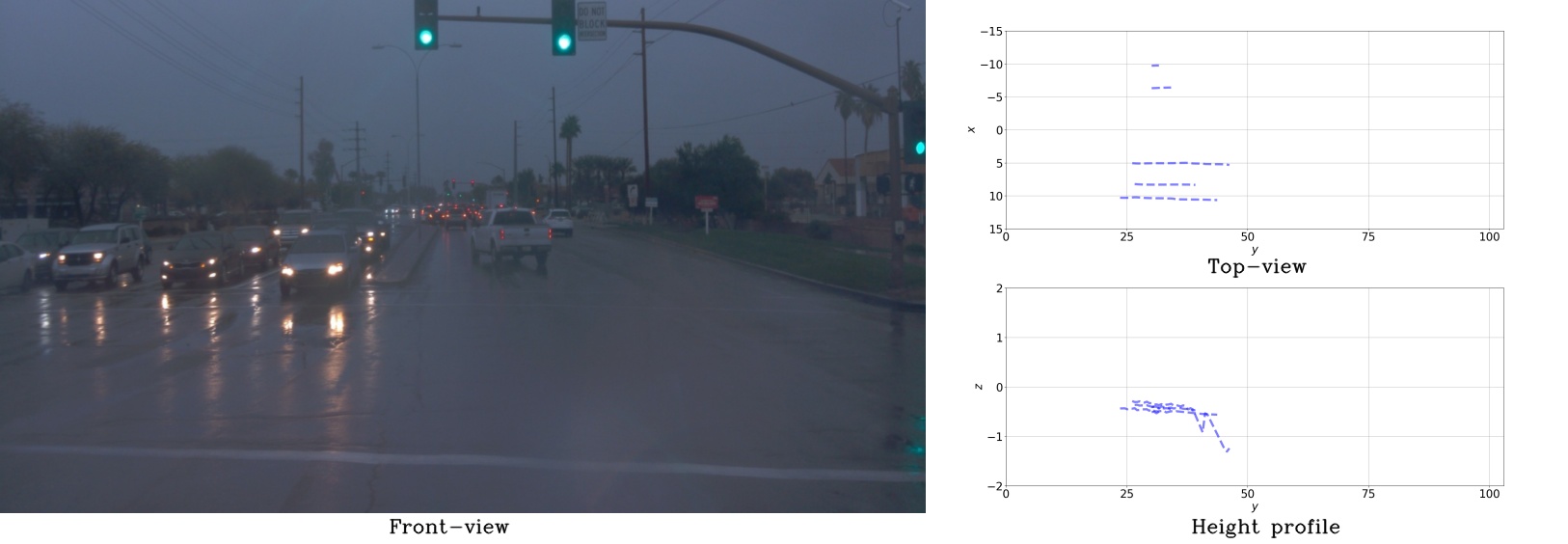}
		\includegraphics[width=1.\linewidth]{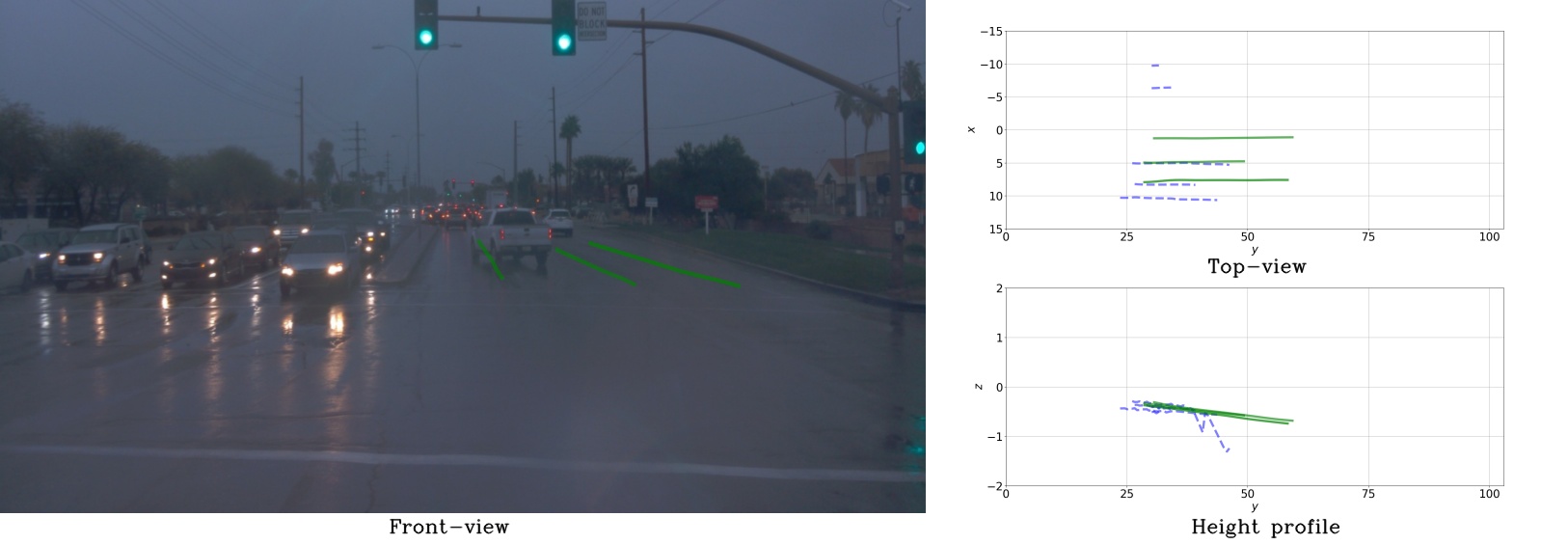}
		\caption{\label{fig:supp-comparison-ol-4}}
	\end{subfigure}
	\begin{subfigure}[b]{0.325\linewidth}
		\centering
		\includegraphics[width=1.\linewidth]{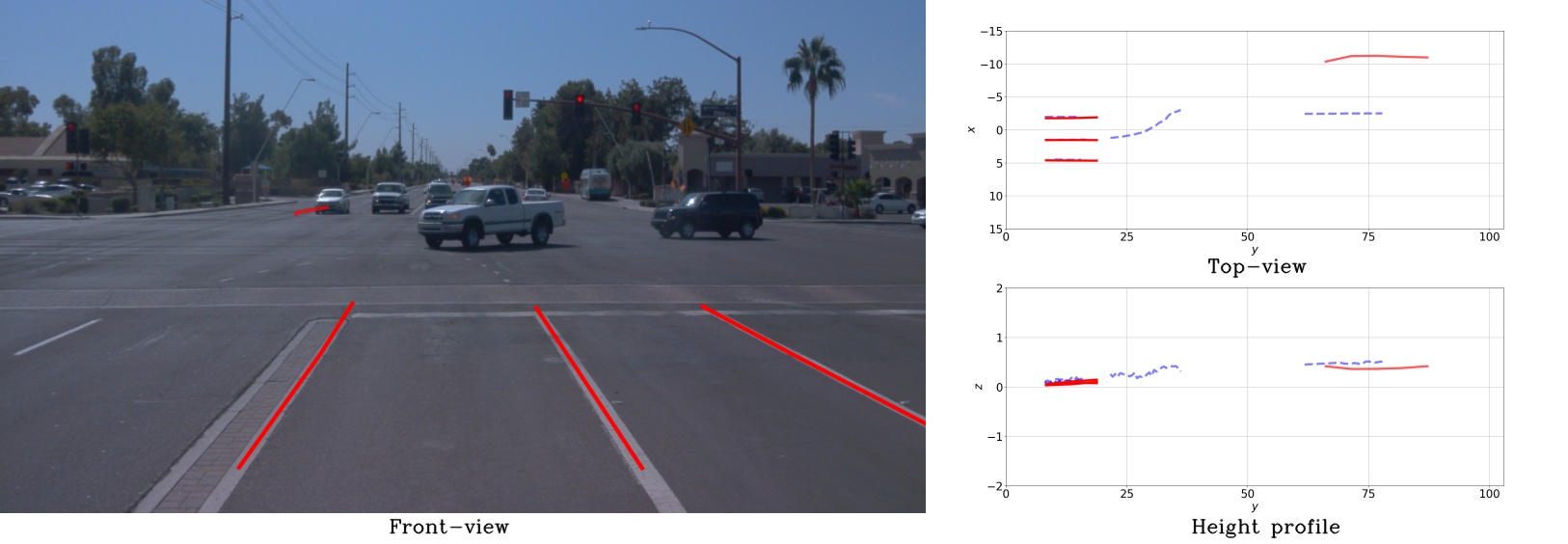}
		\includegraphics[width=1.\linewidth]{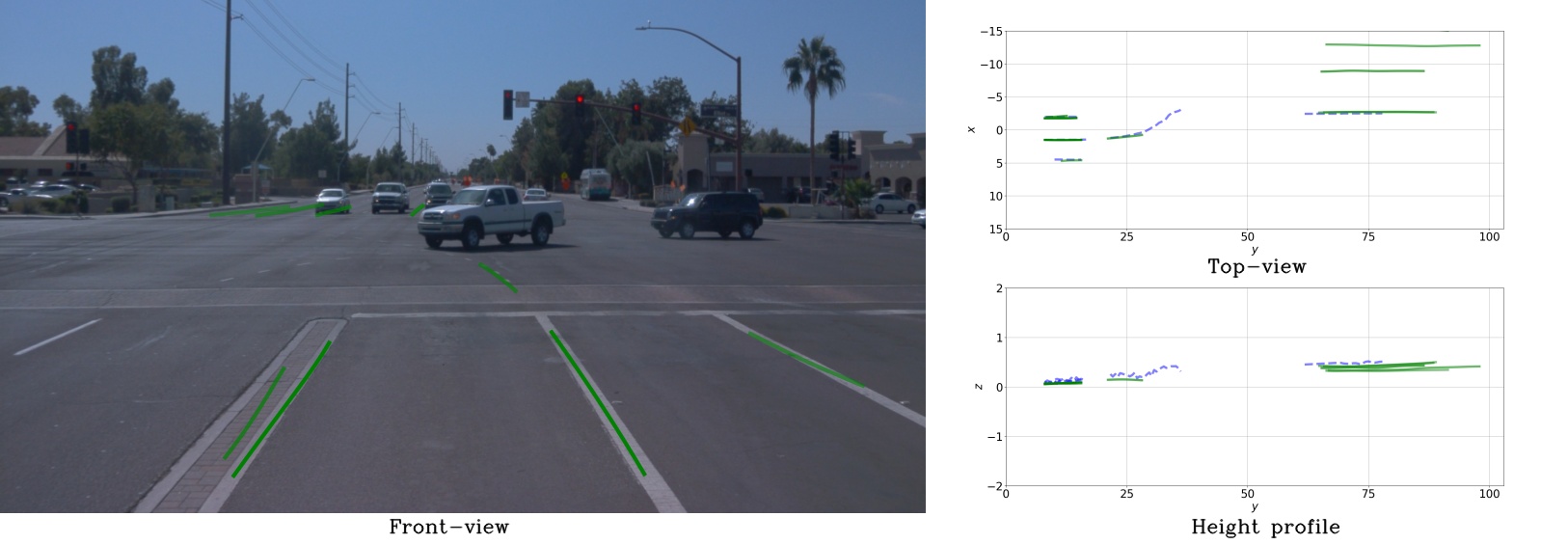}
		\caption{\label{fig:supp-comparison-ol-5}}
	\end{subfigure}
	\begin{subfigure}[b]{0.325\linewidth}
		\centering
		\includegraphics[width=1.\linewidth]{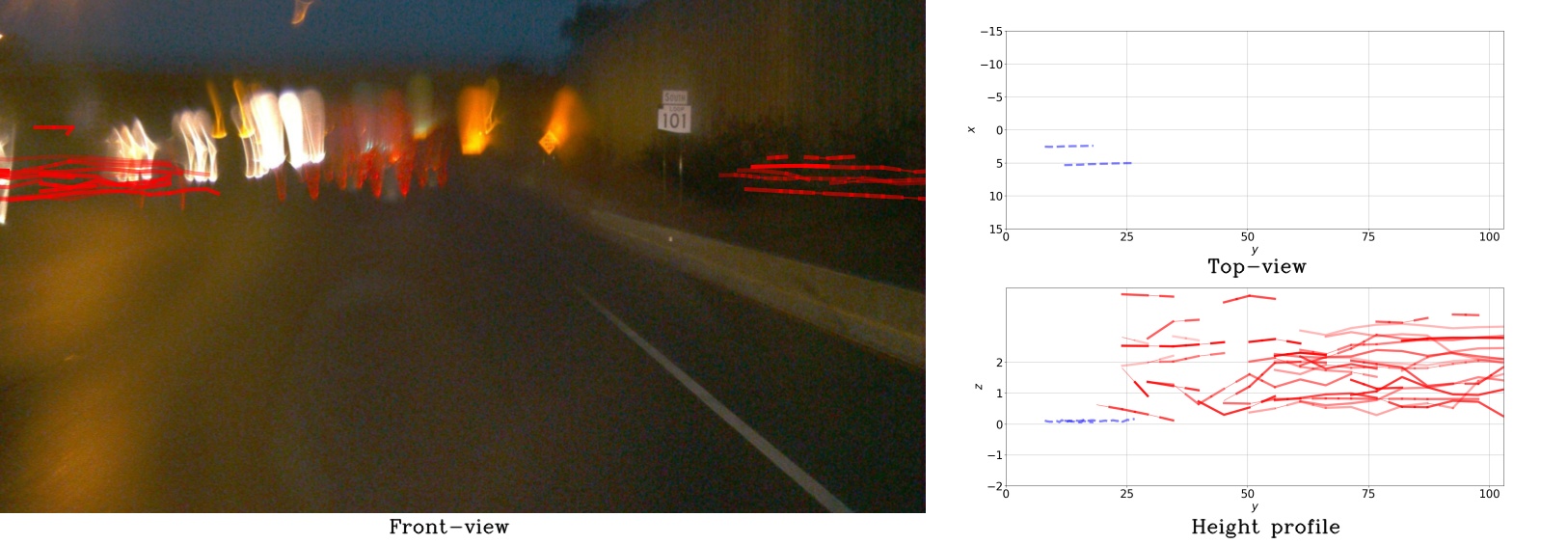}
		\includegraphics[width=1.\linewidth]{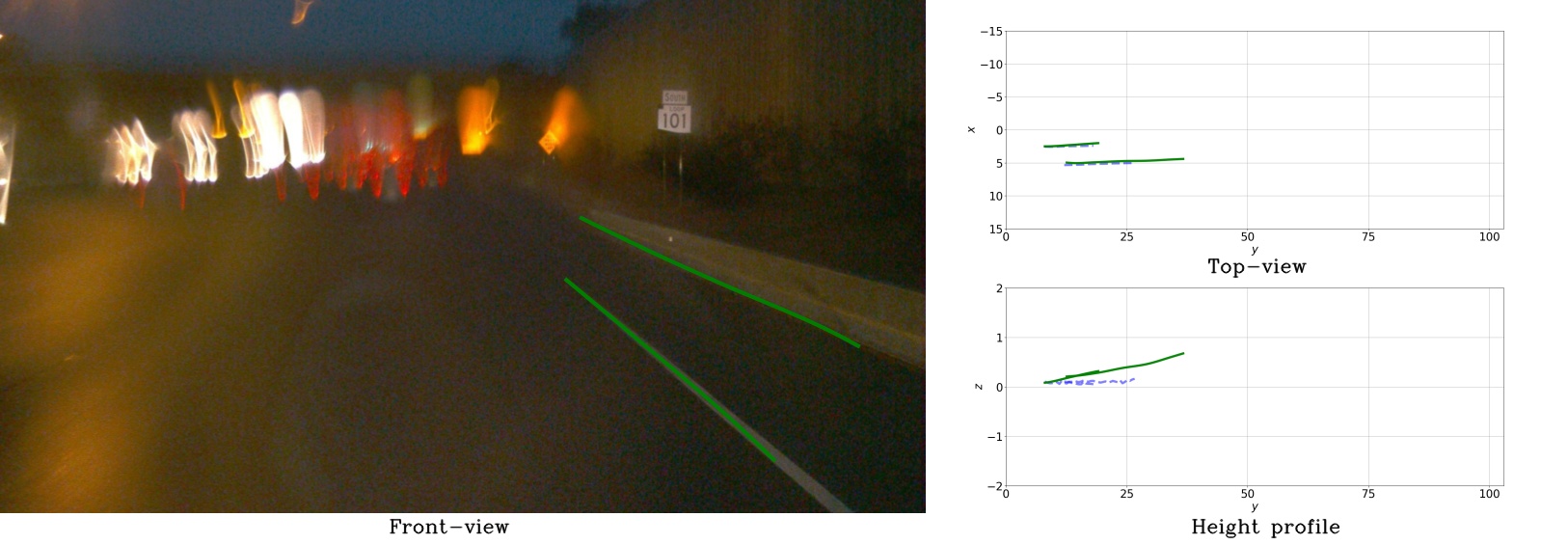}
		\caption{\label{fig:supp-comparison-ol-6}}
	\end{subfigure}
	\caption{Qualitative comparison of \textcolor{red}{LATR} and \textcolor{darkgreen}{SparseLaneSTP} on OpenLane with \textcolor{blue}{ground truth} for reference. 
}\label{fig:supp-comparison-ol}
\end{figure*}

\begin{figure*}
	\centering
	\begin{subfigure}[b]{0.325\linewidth}
		\centering
		\includegraphics[width=1.\linewidth]{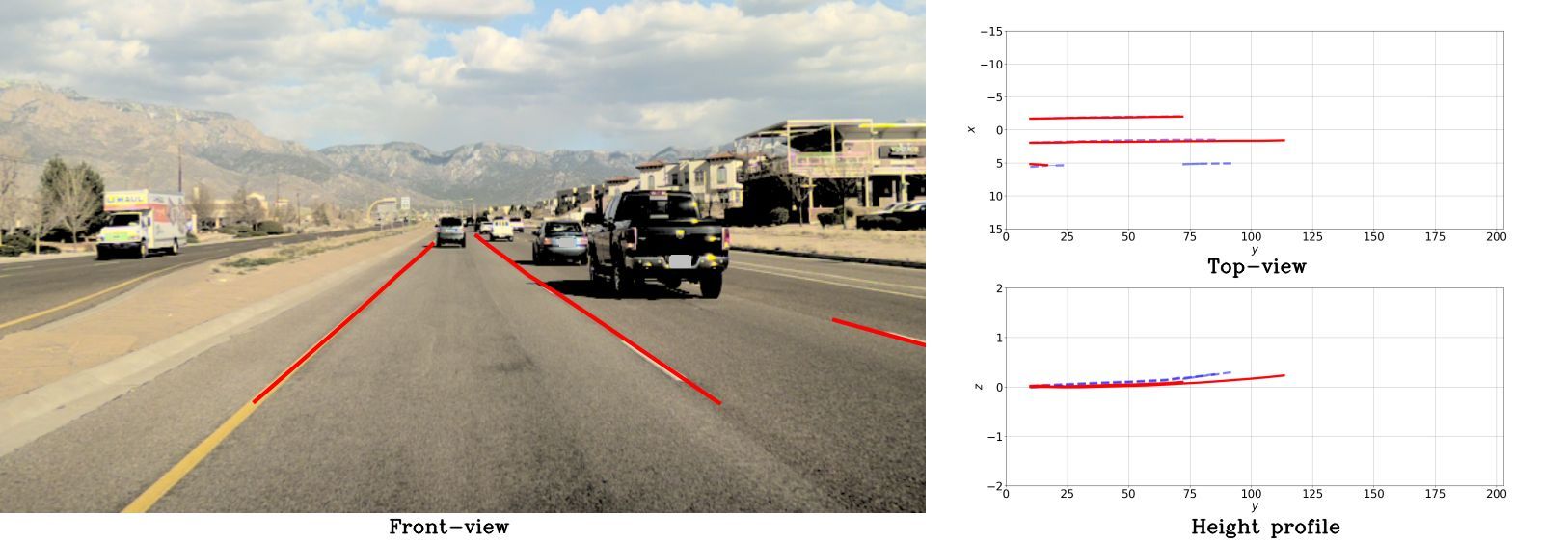}
		\includegraphics[width=1.\linewidth]{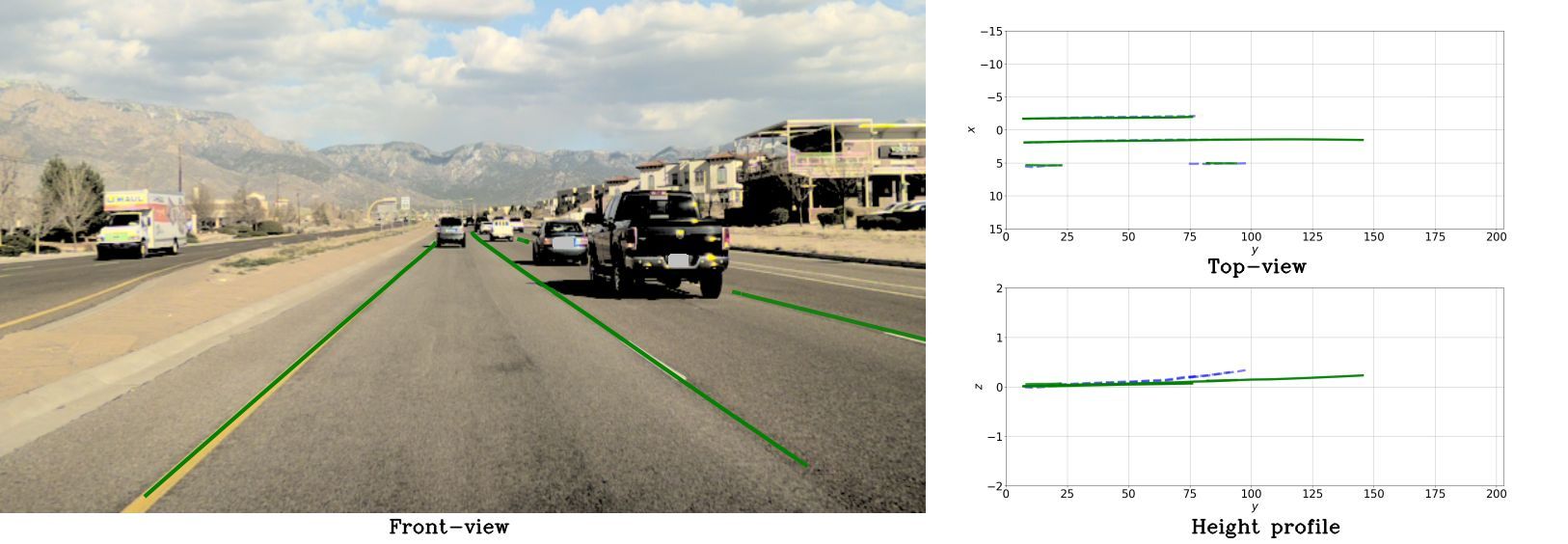}
		\caption{\label{fig:supp-comparison-mpc-1}}
	\end{subfigure}
	\begin{subfigure}[b]{0.325\linewidth}
		\centering
		\includegraphics[width=1.\linewidth]{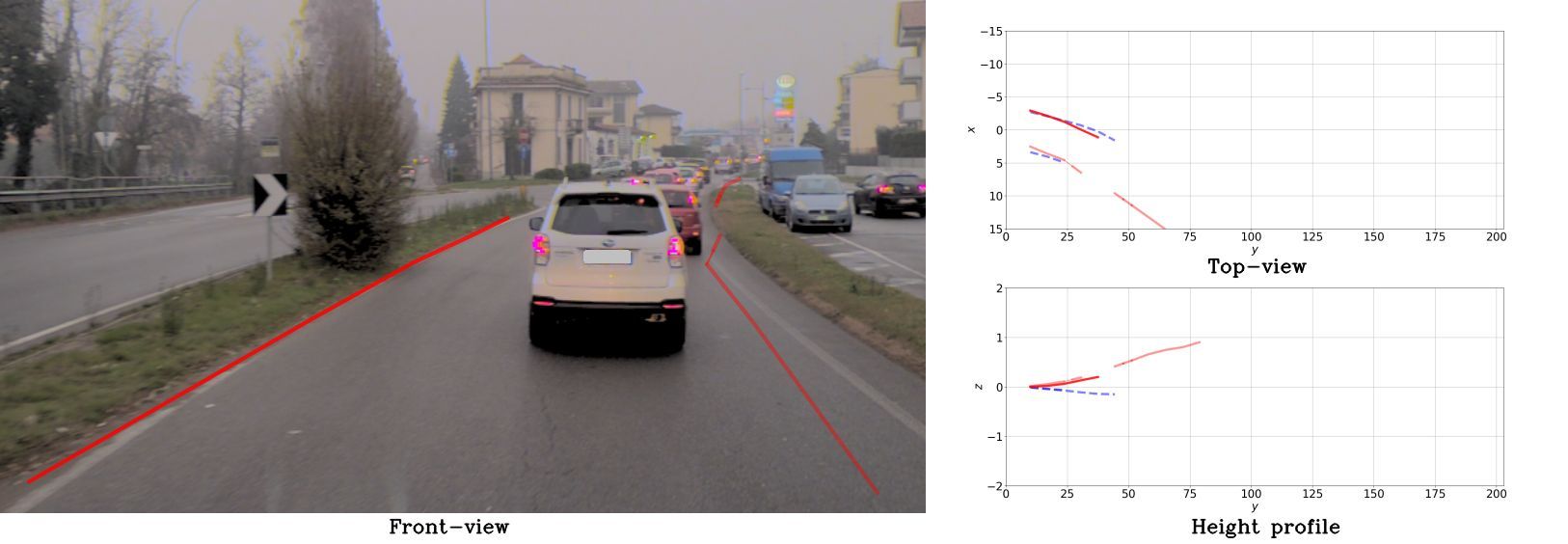}
		\includegraphics[width=1.\linewidth]{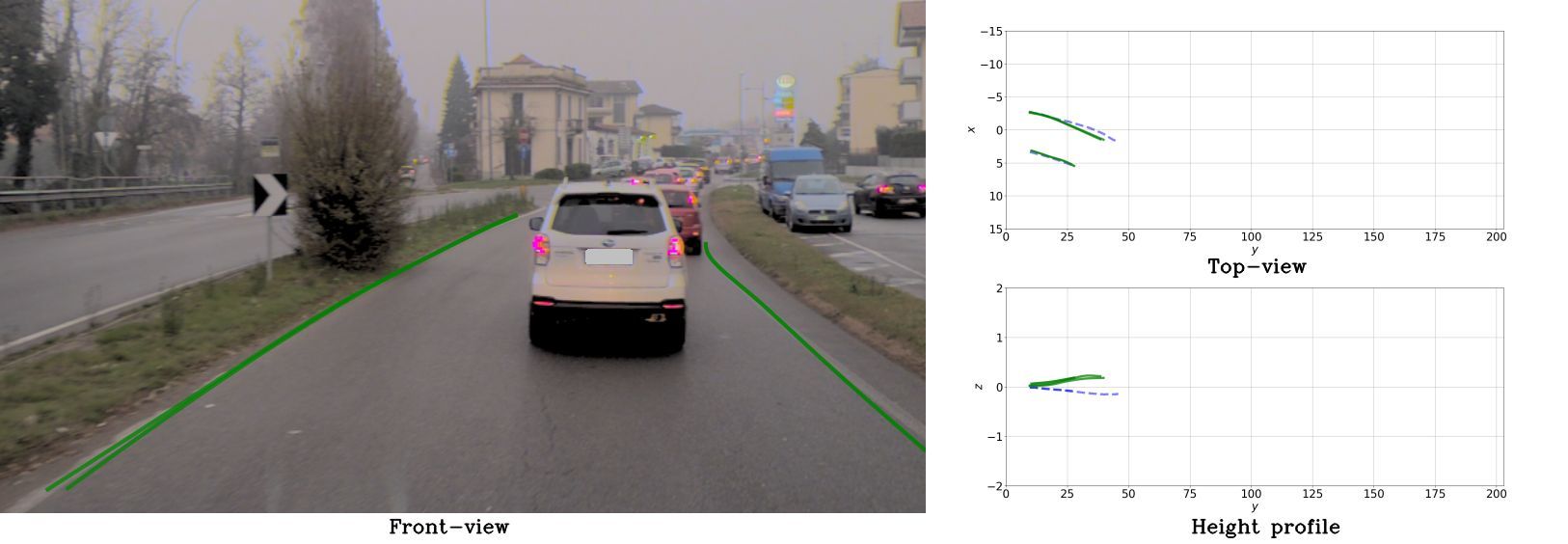}
		\caption{\label{fig:supp-comparison-mpc-2}}
	\end{subfigure}
	\begin{subfigure}[b]{0.325\linewidth}
		\centering
		\includegraphics[width=1.\linewidth]{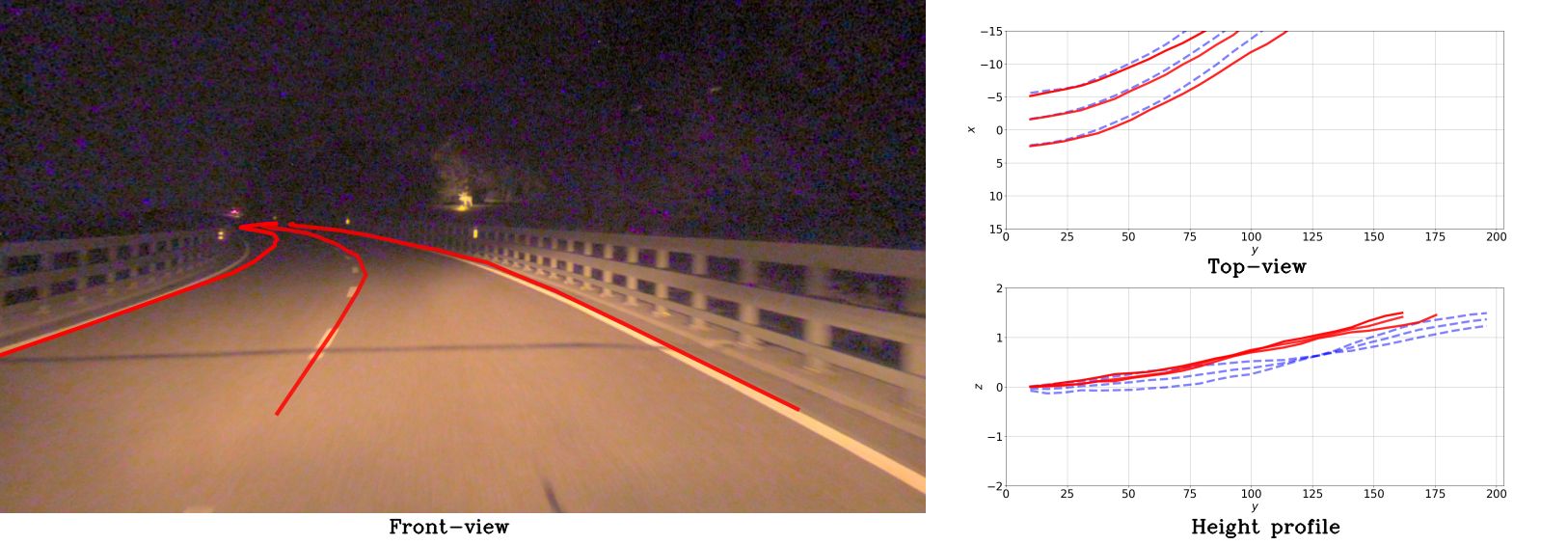}
		\includegraphics[width=1.\linewidth]{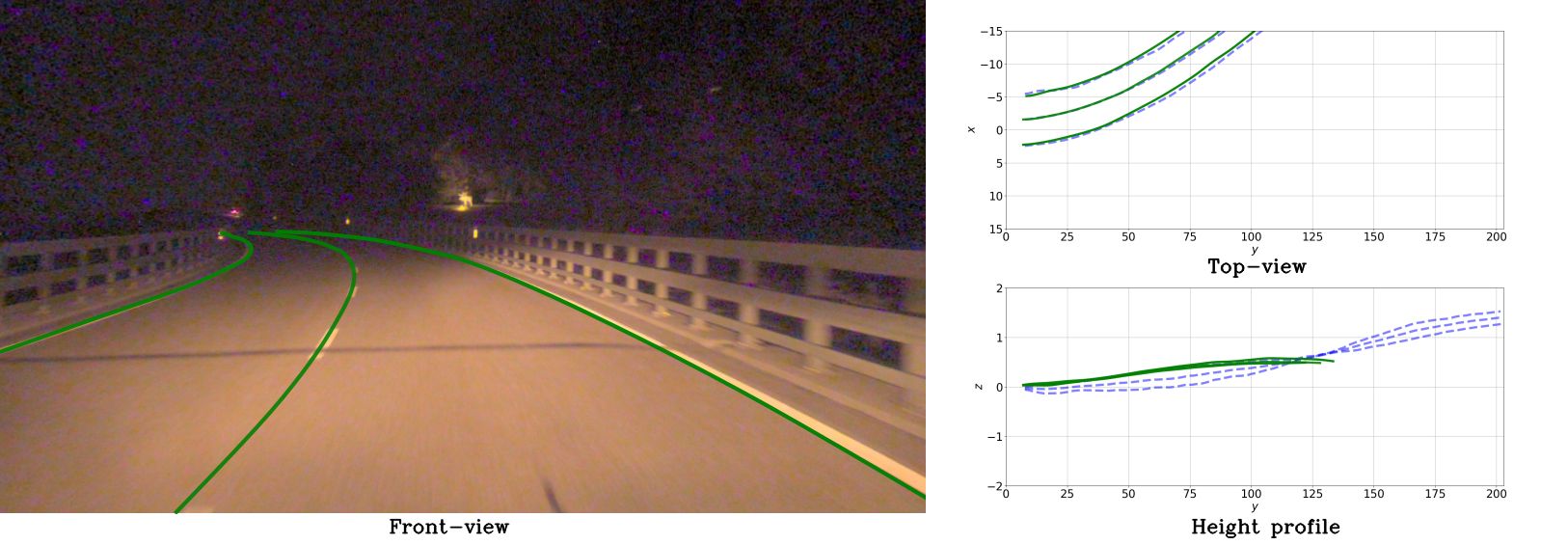}
		\caption{\label{fig:supp-comparison-mpc-3}}
	\end{subfigure}

\begin{subfigure}[b]{0.325\linewidth}
		\centering
		\includegraphics[width=1.\linewidth]{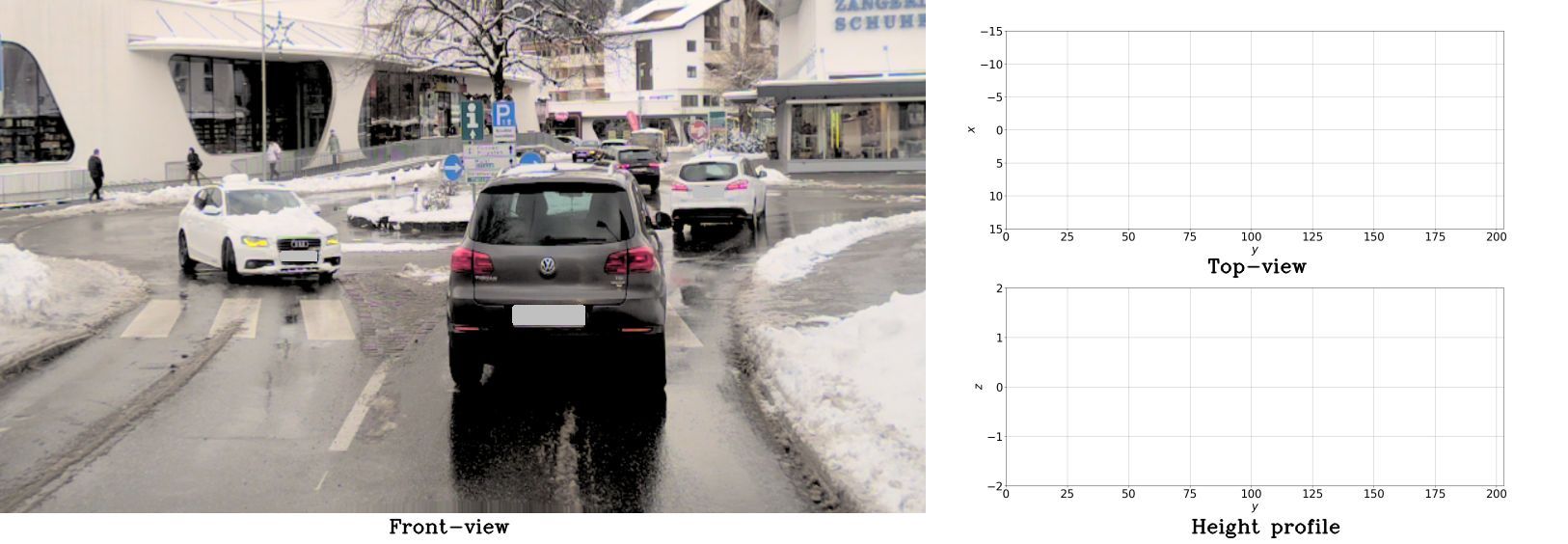}
		\includegraphics[width=1.\linewidth]{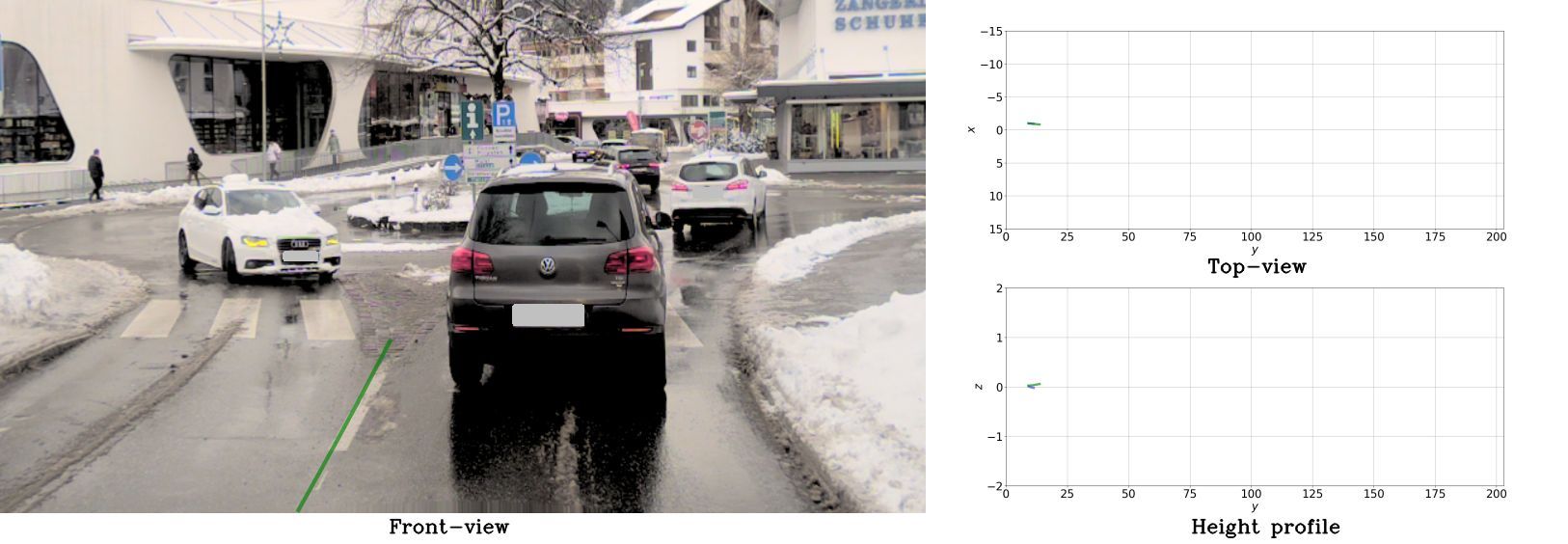}
		\caption{\label{fig:supp-comparison-mpc-4}}
	\end{subfigure}
	\begin{subfigure}[b]{0.325\linewidth}
		\centering
		\includegraphics[width=1.\linewidth]{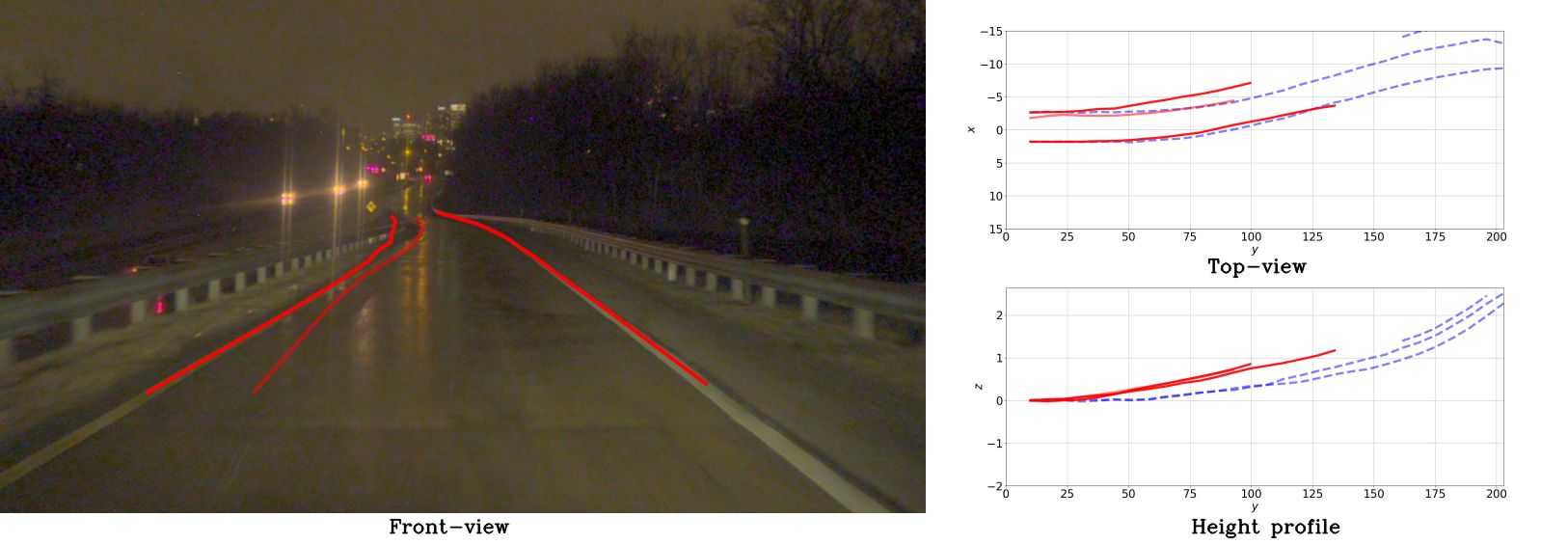}
		\includegraphics[width=1.\linewidth]{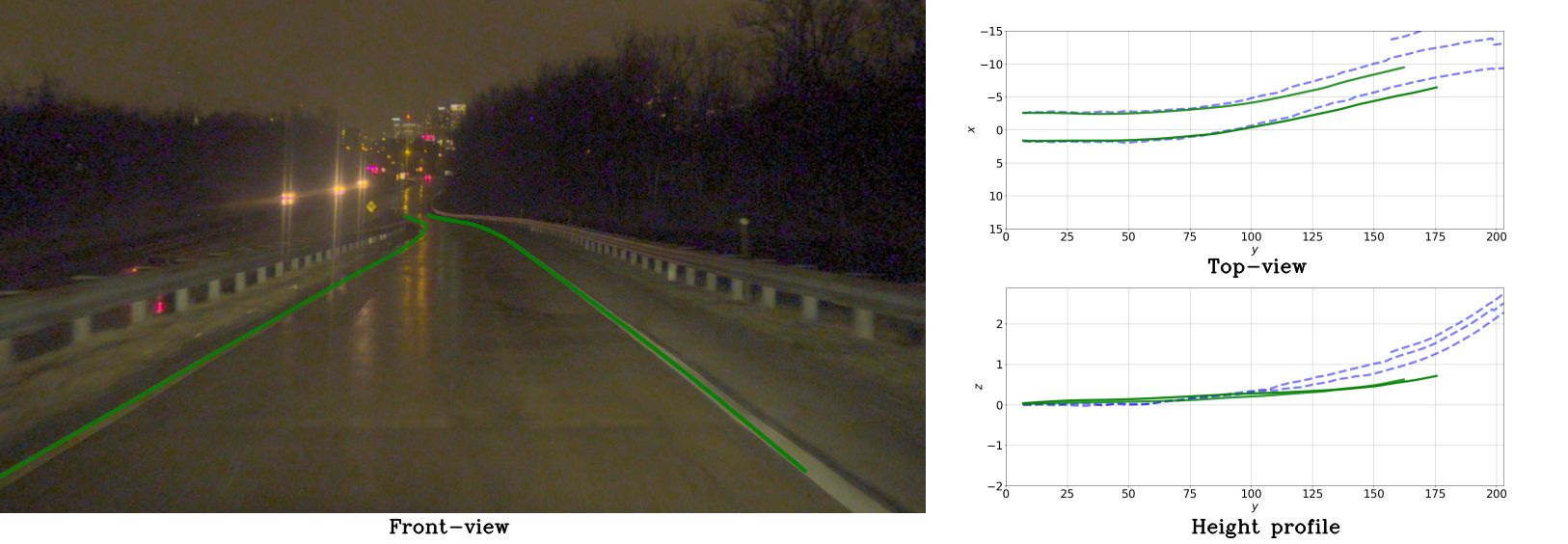}
		\caption{\label{fig:supp-comparison-mpc-5}}
	\end{subfigure}
	\begin{subfigure}[b]{0.325\linewidth}
		\centering
		\includegraphics[width=1.\linewidth]{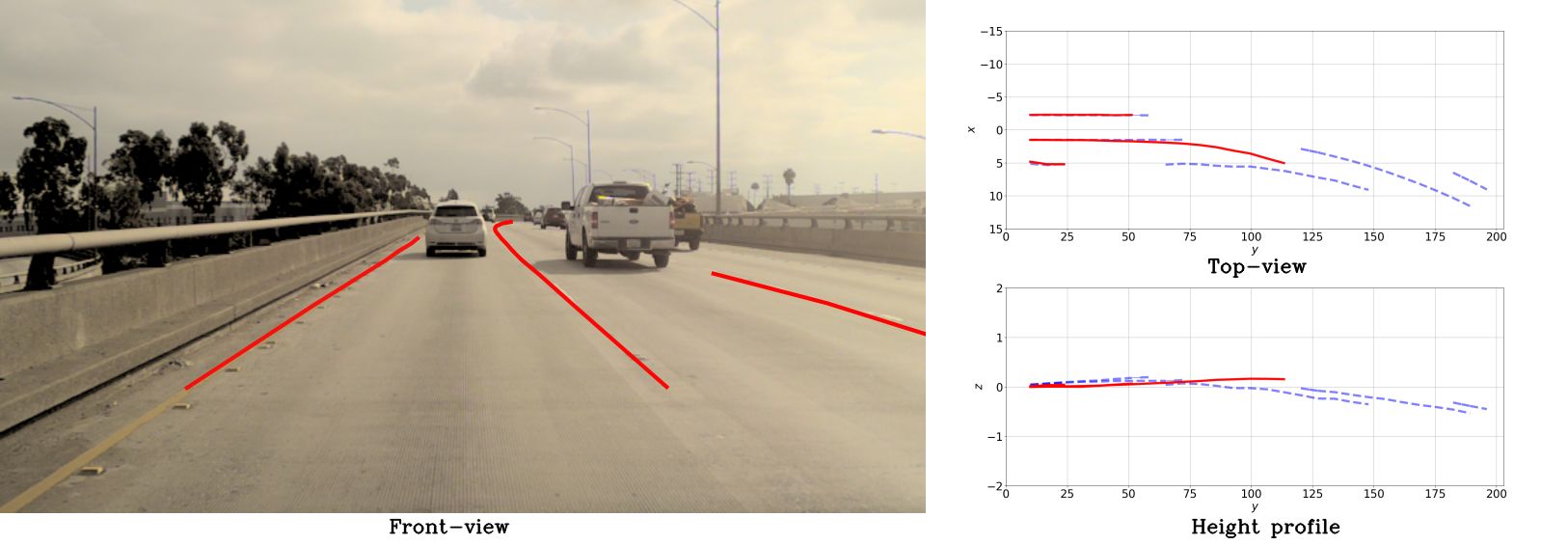}
		\includegraphics[width=1.\linewidth]{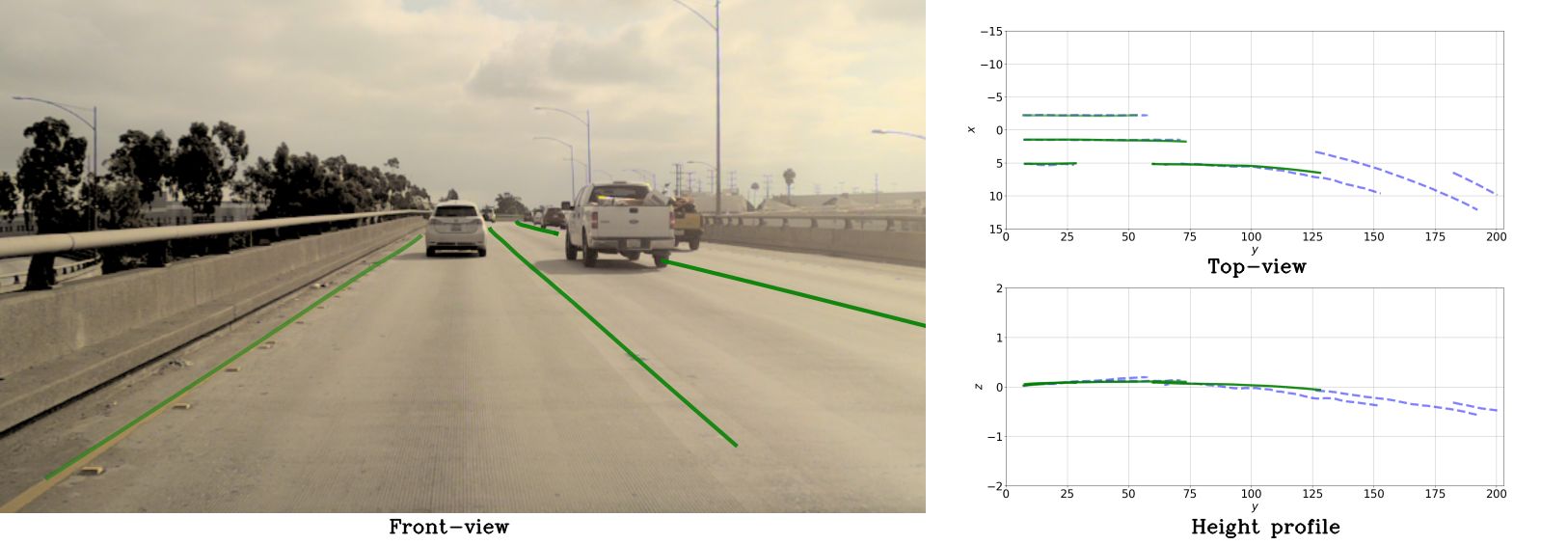}
		\caption{\label{fig:supp-comparison-mpc-6}}
	\end{subfigure}
	\caption{Qualitative comparison of \textcolor{red}{LATR} and \textcolor{darkgreen}{SparseLaneSTP} on our dataset with \textcolor{blue}{ground truth} for reference. 
}\label{fig:supp-comparison-mpc}
\end{figure*}

We also compare our method to the baseline on our 3D lane dataset.
Note that the top-view and height plots in \figref{fig:supp-comparison-mpc} depict twice the range of those in \figref{fig:supp-comparison-ol} while being scaled to the same width for consistency. 

\figref{fig:supp-comparison-mpc-1} demonstrates that our model accurately detects occlusions and even captures small visible segments, whereas the baseline using the discrete representation fails to represent such fine details. 

As shown in \figref{fig:supp-comparison-mpc-3}, our method consistently captures the long-range curve within the visible area, whereas the baseline exhibits inaccurate regression and visibility estimation due to simple extrapolation. 
Similarly, in \figref{fig:supp-comparison-mpc-4} and \figref{fig:supp-comparison-mpc-5}, our model maintains consistent detection throughout the sequence - despite reflections caused by poor lighting conditions. Notably, \figref{fig:supp-comparison-mpc-5} presents a highly challenging scenario, where our model still achieves reasonably accurate results. 

An interesting case is presented in \figref{fig:supp-comparison-mpc-4}. At first glance, the baseline appears to yield better results, however, the top-view and height profile reveal inaccuracies in the 3D geometry due to the absence of priors. In contrast, our method, which incorporates spatial priors, demonstrates a more accurate understanding of lane structure, likely attributed to the STA mechanism.


\subsection{Scenario-based quantitative comparison on our 3D lane dataset}
Given the variety of our dataset with respect to driving environment, daytime, weather and curvature as illustrated in \figref{fig:datasetstats}, we split the test set into eleven different scenario subsets and include a quantitative comparison of our model to the other methods for each subset in \tabref{tab:comparison-quant-scenarios}. 

From the comparison it is clear that our model outperforms the other methods for each scenario. Notably, the margin is even larger for curves and strong curves than for straights, highlighting the capability of SparseLaneSTP to accurately detect lanes in scenarios that show challenging road geometries. 

Besides, the benefit of spatio-temporal priors becomes evident in scenarios of poor visibility (rain, fog). Here the model apparently leverages prior knowledge about road structure and / or previous predictions and queries instead of suffering under poor signal due to the bad view of single frames, leading to a more robust detection behavior. Note that situations in the snow test subset, where the gap of F1-Score is less significant, do not necessarily imply poor visibility since ``Snow'' does not correspond to precipitation but only to snowy environments.
\end{document}